\documentclass[runningheads]{llncs}
\usepackage[T1]{fontenc}
\usepackage{graphicx}
\usepackage{booktabs}
\usepackage[misc]{ifsym}
\usepackage{bm}

\usepackage{mwe}
\usepackage{amsmath,amsfonts}
\usepackage{algorithm}
\usepackage{algorithmic}
\usepackage{appendix}
\usepackage{xcolor}
\begin{document}

\title{Neural Bandit Based Optimal LLM Selection for a Pipeline of Subtasks}

\titlerunning{Neural Bandit Based Optimal LLM Selection for a Pipeline of Subtasks}


\author{
Baran Atalar* \and
Eddie Zhang* \and
Carlee Joe-Wong
}

\institute{
Carnegie Mellon University, Pittsburgh, USA\\
\email{\{batalar, eyzhang, cjoewong\}@andrew.cmu.edu}\\
* Equal contribution.
}



\maketitle              

\begin{abstract}
As large language models (LLMs) become increasingly popular, there is a growing need to predict which out of a set of LLMs will yield a successful answer to a given query at low cost. This problem promises to become even more relevant as LLM agents are asked to solve an increasing variety of ``agentic'' AI tasks. Such tasks are often broken into smaller subtasks, each of which can then be executed by a LLM expected to perform well on that specific subtask. For example, to extract a diagnosis from medical records, one can first select an LLM to summarize the record, select another to validate the summary, and then select a possibly different LLM to extract the diagnosis from the summarized record. Unlike existing LLM selection or routing algorithms, this setting requires selecting a sequence of LLMs, with the output of each LLM feeding into the next and potentially influencing its success. Thus, unlike single LLM selection, the quality of each subtask's output directly affects the inputs, and hence the cost and success rate, of downstream LLMs, creating complex performance dependencies that must be learned during selection. We propose a neural contextual bandit-based algorithm that trains neural networks to guide LLM selections for the different subtasks, without requiring historical LLM performance data. We prove that our proposed Sequential Bandits algorithm achieves a sublinear regret in the number of tasks, and we experimentally validate its superior performance compared to other LLM selection algorithms on two real datasets.

\keywords{neural bandits  \and online LLM selection.}
\end{abstract}

\section{Introduction}
Large Language Models (LLMs) have transformed numerous applications with their ability to summarize, generate, and interpret text. 
Indeed, they are now considered the basis for LLM \textit{agents}~\cite{zhang2024aflow,qiu2024llm}, which in an agentic AI paradigm can solve tasks such as coding, analyzing text, and even acting as a personal assistant. Such agentic AI workflows often improve success rates on a given task by dividing it into a pipeline of multiple subtasks, each of which can then be executed by a specialized LLM agent. For example, in our experiments on predicting a medical diagnosis from medical reports (Section~\ref{exp}), we find that using a summarizer LLM first and then passing its output as the prompt to a diagnoser LLM, i.e., \textit{using a pipelined subtask approach, performs better than using a single LLM without division into summarization and analysis subtasks}. 

Due to different training configurations and model structures, LLMs' performance on different tasks and subtasks can vary widely, raising the need to identify the best-performing LLM agent for a given subtask. While intuitively one could simply select an agent specialized for each subtask, in practice multiple such LLMs may be available, e.g., one of multiple LLMs from OpenAI's marketplace or Azure's Assistants~\cite{assistants}, which may greatly expand the set of available agents. A natural question is then: \textbf{How should we select the best LLM agent to complete each subtask within a given task?}

\textbf{Research Challenges.} Selecting the most suitable LLM for a specific task or sequence of subtasks presents significant challenges, particularly in terms of computational efficiency and performance optimization, as has been explored in prior work on LLM selection for single tasks~\cite{zhang2024efficientcontextualllmcascades,zhao2023automaticmodelselectionlarge}. 
Due to computational, monetary, and latency concerns, simply running a subtask through all the available models and choosing the model that yields the best performance for that query is not feasible \cite{shazeer2017outrageouslylargeneuralnetworks}, while simply selecting the largest LLM for every subtask may be either prohibitively expensive (as these LLMs tend to be the most costly) or ignore the potential of highly specialized LLMs to perform well on specific types of tasks~\cite{shnitzer2023largelanguagemodelrouting}. These challenges are compounded within agentic AI workflows in which a LLM agent must be selected for \textit{each} subtask, resulting in an exponential number of possible LLM agent combinations. 
Even worse, the LLM agents chosen for each subtask cannot be chosen independently: the output of the agent executing the first subtask becomes the input for the next subtask, leading to coupled performance and compounding costs. In our medical diagnosis example, among 7 candidate LLMs, a specialized Med-III model has the highest accuracy when a summarizer is not used (\%42 vs \%38 with summarizer), but the Llama-3.3B model has the highest accuracy when a summarizer is used (\%43 vs \%41 with no summarizer) (Figure \ref{fig:avg_acc} in the Appendix shows the summarizer step improves performance of most models). 

Adaptively learning which LLM to choose for each subtask is made difficult by the fact that in practice, users may not have historical data on LLM performance for specific subtasks, especially if some candidate LLMs are customized ``assistants''. Thus, a natural solution is to take an \textit{online approach}, where the user both learns and optimizes the LLM selection as it sends queries/tasks. Online LLM selection for a given subtask can be viewed as an instance of the classical contextual multi-armed bandit (MAB) problem~\cite{linucb}. Each LLM is modeled as an ``arm,'' and the MAB algorithm sequentially pulls the arm (i.e., uses this LLM), monitoring the subtask success and adapting future LLM selections accordingly. MAB balances exploration, i.e., trying new LLMs in the hope that they outperform the identified current best one; with exploitation, i.e., selecting the best LLM found so far. Context, i.e., an embedding of the current query, is used to predict the success of each LLM.
However, conventional contextual MAB methods do not capture the \textit{sequential} nature of LLM selection in our agentic workflow scenario: we must select one LLM for \textit{each} subtask. Combinatorial variants of contextual MAB~\cite{pmlr-v28-chen13a}, in which a combination of arms (here, LLMs) are chosen in each round, is closer to our scenario, but it requires selecting LLMs for each subtask at once, instead of selecting the next LLM in the pipeline after observing the results from prior LLMs. Moreover, it is not clear how to modify these algorithms to account for both the success rate and the monetary cost of using each LLM. Thus, we propose a novel MAB variant designed specifically for \textit{sequential} LLM selection, which we show outperforms conventional MAB methods. To the best of our knowledge, \textbf{ours is the first work to examine cost-effective LLM selection within such a pipeline of tasks}.


The main \textbf{contributions} of this work are as follows:
\begin{itemize}
    \item We introduce a novel \textbf{problem formulation} of selecting a pipeline of LLMs to solve a task decomposed into interconnected, smaller subtasks.
    \item We \textbf{adapt contextual MAB algorithms} into the Sequential Bandit algorithm, which sequentially selects LLMs in the task pipeline. Sequential Bandits trains neural networks to effectively 
    optimize a combination of LLM success at each subtask and cost in an online manner.
    \item To evaluate SeqBandits, we \textbf{create a new diagnosis prediction dataset} from an existing medical dataset \cite{johnson2016mimic} of deidentified medical reports.
    \item We \textbf{experimentally show} that SeqBandits identifies better LLMs than existing MAB  algorithms on medical diagnosis and telecommunications tasks.
    \item We \textbf{prove} 
    that SeqBandits achieves $\tilde{O}(\sqrt{T})$ regret over  $T$ tasks.
\end{itemize}
We first outline related work (Sec.~\ref{sec:related}) and describe our problem formulation (Sec.~\ref{sec:formulation}). We present our SeqBandits algorithm in Sec.~\ref{sec:algorithm}, theoretical results in Sec. \ref{sec:theory} and experimental evaluation in Sec.~\ref{exp},  then conclude in Sec.~\ref{sec:conclusion}.

\section{Related Work}
\label{sec:related}
The proliferation of LLMs has led to growing interest in methods to predict the best-performing LLM for a single subtask. 
We divide these strategies into budget‑aware frameworks, LLM cascades, and LLM routing.

\textbf{Budget Constrained Online Algorithms and Bandits}. These algorithms maximize cumulative reward subject to a hard cap on total resource consumption. Primal–dual schemes embed standard regret minimizers as black‐box components to enforce long‑term resource constraints via dual variables \cite{castiglioni2022unifyingframeworkonlineoptimization}, which can extend to integrate budgeted expert‐query mechanisms that judiciously allocate a limited number of advice calls to sharpen decisions \cite{benomar2024addressingbiasonlineselection}. 
Non‑stationarity and adaptive primal–dual updates have been shown to ensure constraint satisfaction even as cost and reward distributions shift over time \cite{NEURIPS2022_69469da8}, including contextual MAB approaches with evolving contexts~\cite{poon2025onlinemultillmselectioncontextual}. Most recently, weakly adaptive regret minimizers have been woven into primal–dual frameworks to simultaneously honor budget and return‑on‑investment limits \cite{pmlr-v235-castiglioni24a}. 
Another work \cite{dong2024cost}, considers single-step knowledge-based question answering with contextual MAB. 
None of these works, however, use sequential decision frameworks for a pipeline of subtasks, as proposed in our setting.

\textbf{Cost-Efficient LLM Cascades}.
In a typical cascading framework, inputs are processed through a pre-determined sequence of LLMs, from the least to the most resource-intensive. At each stage, the system evaluates the output to determine whether to accept the result or continue to the next model in the sequence \cite{zhang2024efficientcontextualllmcascades}.
Recent advancements have focused on integrating cascading with routing strategies, i.e., routing a query to the ``best'' LLM~\cite{chen2023frugalgptuselargelanguage,dekoninck2025unifiedapproachroutingcascading}. 
%
Approaches like the Mixture of Thought representations combine chain-of-thought and program-of-thought prompts, in order to adaptively route simpler queries to smaller, less costly models, reserving more complex tasks for larger models \cite{cheng2023binding}. This strategy has demonstrated significant reductions in inference costs while maintaining accuracy comparable to using the most robust LLM alone \cite{yue2024largelanguagemodelcascades}.
However, cascading can be inefficient if the sequence is not optimally configured, as each input may need to pass through multiple models before reaching an adequate response \cite{dekoninck2025polyratingcosteffectivebiasawarerating}. 
Unlike cascades, Sequential Bandits considers a pipeline of LLMs, in which the different LLMs perform subtasks that feed into each other \cite{li2025llmbanditcostefficientllm}.


\textbf{Model Selection and Adaptive Routing}.
Dynamic routing mechanisms, which intelligently direct queries to the most appropriate LLMs, can significantly improve performance and computational resources \cite{varangotreille2025doingimplementingrouting}, as unlike cascades, they do not make multiple passes through different LLMs. Systems like Tryage propose context-aware routing mechanisms that optimally select expert models based on individual input prompts \cite{hari2023tryagerealtimeintelligentrouting}, allowing users to explore trade-offs between task accuracy and secondary goals like minimizing model size~\cite{sikeridis2024pickllmcontextawarerlassistedlarge}. 
%
Other approaches like Zooter train a routing function using reward models' scores as supervision signals \cite{lu2023routingexpertefficientrewardguided}, efficiently directing queries to specialized LLMs~\cite{chen2025harnessingmultiplelargelanguage}.

Within the popular mixture-of-experts framework, Routing Experts introduces a dynamic expert scheme for multimodal LLMs that also 
aims to learn more efficient inference pathways ~\cite{wu2025routingexpertslearningroute}. Other works focus on the challenge of dynamic routing, where new, previously unobserved LLMs become available at test time. These strategies generalize by representing each LLM as a feature vector \cite{jitkrittum2025universalmodelroutingefficient}. Frameworks like AutoMix~\cite{aggarwal2025automixautomaticallymixinglanguage} focus on predicting LLM success, e.g., with a few-shot self-verification mechanism to estimate output reliability from smaller LLMs \cite{ding2024hybrid}. 
Recent work like BEST-Route has expanded routing by introducing architectures that dynamically allocate queries across specialized LLMs, optimize test-time compute for accuracy–latency trade-offs, and adapt routing policies to real-time budget constraints \cite{ding2025bestrouteadaptivellmrouting}. ADaPT \cite{prasad2024adaptasneededdecompositionplanning} introduces an adaptive task-routing framework where an LLM recursively decomposes a task only when the current executor is predicted to fail, 
dynamically alternating between planning and execution to adjust task difficulty based on LLM capability. 
All of these methods, however, focus on selecting a single LLM per query, and cannot be directly applied to selecting a pipeline of LLMs for different pre-defined subtasks as in our setting.



\section{Problem Formulation}
\label{sec:formulation}

\begin{figure} 
  \vspace{-12pt} 
  \centering
  \includegraphics[width=\linewidth]{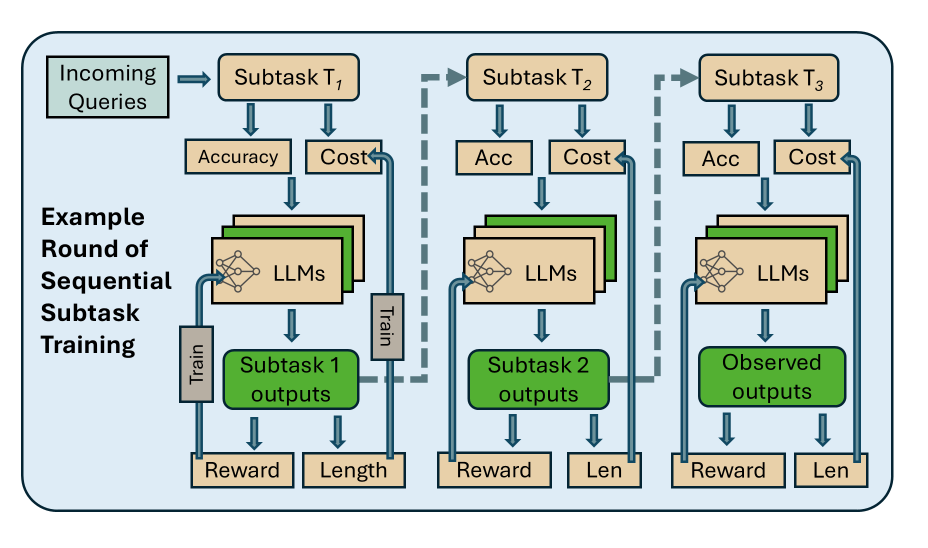}
  \vspace{-17pt}
  \caption{Our Sequential Bandits problem setting and approach, where an incoming query is split into simpler subtasks in a sequential pipeline. An LLM is chosen to complete each subtask, and its output is fed as the input to the next subtask's LLM.}
  \label{fig:formulation}
  \vspace{-6pt} 
\end{figure}

We consider multiple rounds $t = 1,2,\ldots$, where each round is defined by an incoming query $q_t$. 
In our formulation, we assume that the breakdown of a task into simpler subtasks $\{T_1,T_2,...,T_K\}$ is given. These subtasks form a directed acyclic graph (DAG), as the output of a subtask becomes the input of the next task in the pipeline. For example, a query could be a medical diagnosis prediction based on a provided medical report, broken into the subtasks of (i) summarizing the report, (ii) validating the summary, and (iii) predicting a diagnosis given the validated summary. 
In the remainder of the paper, 
our aim is to select the LLM returning the highest accuracy and lowest cost for each subtask. Note that a special case of this setting is a query with one subtask (itself).
Figure \ref{fig:formulation} illustrates our pipelined/sequential problem setting. This approach is in essence similar to the popular chain-of-thought (CoT) prompting \cite{wei2023chainofthoughtpromptingelicitsreasoning}, but we \textit{allow different LLMs to complete each subtask}, unlike CoT in which a single LLM decomposes the task and completes each subtask. We formalize this selection problem below. Throughout, we denote by $[N]$, $N \in \mathbb{Z}^+$, the set $\left\{1,2,\ldots,N\right\}$. 

\textbf{Bandit Formulation.} In this paper, we model the LLM selection problem as a neural contextual bandit. We consider $T$ rounds. The subtasks $\{T_1,T_2,...,T_K\}$ correspondingly have $\{N_1,N_2,...,N_K\}$ available LLMs, each of which we refer to as an ``arm'' for a particular subtask. We use $\mathcal{M}$ to denote the set of all LLMs. In every round $t$, the agent selects an arm (LLM) for each of the subtasks corresponding to query $q_t$, and we name the overall set of arms selected as the \textbf{super arm} $S_t$, following the combinatorial bandit literature \cite{pmlr-v28-chen13a}. Formally, for each subtask $T_i$ where $i \in \{1, \dots, K\}$, select an arm $b_{i,j}$ where $ j\in [N_i]$. Then $S_t = (b_{1,j}, b_{2,j}, \dots, b_{K,j}) \text{ where } b_{i,j}$ represents the arm $j$ selected from $[N_i]$ for subtask $i$. When a super arm $S_t$ is selected in round $t$, the agent observes the \textbf{base arm rewards} of the chosen super arm, namely $\{ r_{t,j}\}_{b_{i,j} \in S_t}$ and receives a \textbf{total (super arm) reward} of $R(S_t,\mathbf{r}_t)$ where $\mathbf{r}_t = [r_{t,j}]_{b_{i,j} \in S_t}$, i.e., the super arm reward is a function of the individual base arm rewards. In our setting, the base arm rewards correspond to the ``goodness'' of the result for each subtask, and the super arm reward is a measure of goodness of the LLMs' collective result on the query, which is a function of the selected base arm rewards. For example, the super-arm reward could be the accuracy of the output returned by the last LLM in the pipeline for a prediction task like medical diagnosis, which is the final output returned to the user. 
We also account for the \textbf{cost} associated with deploying the selected LLM for each subtask. While we take this to be the monetary cost, it could also be taken as the energy consumed or latency of the LLM's inference on the given subtask. We calculate the cost in terms of the number of input and output tokens of the chosen LLM for the given subtask. We give more detail about how cost estimation is done in the next section.

\textbf{Context, Reward and Cost Estimation.} In every round $t$, the agent observes the \textbf{context} $\mathbf{x}_{t}(b_{i,j})$ of arm $b_{i,j}\text{ for each } i \in \{1, \dots, K\}, j\in[N_i]$, which helps guide the prediction of LLM $j$'s success on subtask $i$. The context $\mathbf{x}_{t}(b_{i,j})$ in our application is a function of the given prompt for subtask $i \in \{1, \dots, K\}$, which is the output of the LLM chosen for the previous subtask $i-1$ as shown in Figure \ref{fig:formulation}, as well as features of the LLM corresponding to arm $(i,j)$. In our experiments, these features include
a description of the LLM's capabilities, as well as whether it was finetuned on a specific dataset for a certain task. We give more details on the context construction in the Appendix. We denote the description of an LLM $j$ as $d_j$ and the set of descriptions of all available LLMs as $\mathcal{D}$. Letting $a_{t,i-1}$ be the LLM selected for subtask $i-1$, we formalize the generation of the contexts in the pipeline via the transition function $o_{i-1}(\mathbf{x}_{t,i-1},a_{t,i-1})=\mathbf{x}_{t,i}$ where $\mathbf{x}_{t,i}$ represents the output of stage $i-1$ for the chosen LLM at round $t$ (we drop the $j$ index of the chosen LLM for simplicity). 

Following prior works on neural bandits, we use a neural network(s) to learn the underlying reward function(s). We assume that $ \forall t \in [T]$, the base arm reward $r_{t}(b_{i,j})$ of arm $b_{i,j}$ (i.e., LLM $j$ on subtask $i$) is generated as follows \cite{neuralucbzhou20a}:
\begin{align}
r_{t}(b_{i,j}) = h_{i,j}(\mathbf{x}_{t}(b_{i,j})) + \xi_{t}
\label{eq:reward}
\end{align}
 where $h_{i,j}$ is the underlying unknown reward function for arm $b_{i,j}$ and $\xi_{t}$ is $Q$-sub-Gaussian noise, modeling uncertainty in the ``goodness'' of the LLM output. We allow separate reward functions for the different (subtask, arm) combinations, since each combination corresponds to a different model structure and task. To learn the base arm reward function $h_{i,j}$, we use a fully connected neural network with depth $L+1\geq 3$ and width $m$ \cite{neuralucbzhou20a}:
\begin{align}
f_{i,j}(\mathbf{x};\boldsymbol{\theta}) = \sqrt{m}\mathbf{W}^{(L)}_{i,j}\kappa(\mathbf{W}^{(L-1)}_{i,j}\kappa(...\kappa(\mathbf{W}^{(0)}_{i,j}\mathbf{x}))
\end{align}
where $\boldsymbol{\theta}_{i,j} = [$vec$(\mathbf{W}_{i,j}^{(0)})^T,...,$vec$(\mathbf{W}^{(L)}_{i,j})^T]^T$ is the weight vector and $\kappa(x) = $ max$\{x,0\}$ is the ReLU activation. We denote the weight vector in round $t$ for arm $b_{i,j}$ as $\boldsymbol{\theta}^t_{i,j}$. We let the gradient of the neural network at time $t$ be $\mathbf{g}^t_{i,j}(\mathbf{x};\boldsymbol{\theta}^t_{i,j})$.


The monetary cost of passing prompt $q_t$ for LLM $j$ is modelled as the number of input and output tokens, multiplied by LLM $j$'s corresponding per input token and output token costs in Microsoft Azure (see the Appendix for more details). We define the maximum output token cost among all LLMs as $P_a$. We know the number of input tokens since $q_t$ is given. Since we do not know the number of output tokens LLM $j$ will output in advance for a given prompt, 
we train an output token length prediction model \cite{qiu2024efficient}, as detailed in Sec.~\ref{exp}. We denote the actual cost of passing a prompt $\mathbf{x}$ through LLM $a$ for subtask $i$ as $c_i(\mathbf{x},a)$. Let $C(\mathbf{x},j;\boldsymbol{\Theta}_t)$ be the BERT-based cost regression model's (with width $n$) predicted number of output tokens for input prompt $\mathbf{x}$ and LLM $j$ with parameters $\Theta_t$ at round $t$. Let $\mathbf{v}_j^t(\mathbf{x};\boldsymbol{\Theta}_t)=\nabla_{\boldsymbol{\Theta}}C(\mathbf{x},j;\boldsymbol{\Theta}_t)$ be the gradient of the model at time $t$. 

\textbf{Problem Objective.} Our main objective is to maximize the reward (accuracy) while also minimizing the cost of deploying LLMs for the tasks. To combine these two metrics into a single one, we define the \textbf{net reward} for subtask $i$, prompt $\mathbf{x}$ and LLM $a$ as $n_i(\mathbf{x},a) = h_i(\mathbf{x},a)-\alpha_i c_i(\mathbf{x},a)$ where $\alpha_i >0$ trades off the relative importance of reward and cost for each subtask. We denote the algorithm's total utility at round $t$ by $U_t:=\sum_{i=1}^Kn_i(\mathbf{x}_{t,i},a_{t,i})$ where $a_{t,i}$ denotes the chosen LLM of the algorithm for subtask $i$ at round $t$ and $\mathbf{x}_{t,i}$ is the input prompt for subtask $i$, generated sequentially through the selected LLMs at previous steps of the pipeline. We also define the \emph{dynamic pipeline oracle} as the policy that, given $\mathbf{x}_{t,1}$, selects
$a^*_{t,1:K} = (a^*_{t,1}, \dots, a^*_{t,K})$ maximizing $U_t^*:=\sum_{i=1}^K n_i(\mathbf{x}^*_{t,i}, a^*_{t,i})$, where $\mathbf{x}^*_{t,1} = \mathbf{x}_{t,1}$ and $\mathbf{x}^*_{t,i+1} = o_i(\mathbf{x}^*_{t,i}, a^*_{t,i})$.

Since we aim to maximize the net reward in our setting, we define the \textbf{regret} as $R_T:=\sum_{t=1}^T(U_t^*-U_t)$. The optimal cumulative \textbf{value function} from stage $i$ is defined recursively as $V_i(\mathbf{x})=\max_a[n_i(\mathbf{x},a)+V_{i+1}(o_i(\mathbf{x},a))]$ with $V_{K+1}(\mathbf{x})=0.$ 

\section{Sequential Bandits Algorithm}
\label{sec:algorithm}

In this section, we present our proposed algorithm, Sequential Bandits (Algorithm 1), which aims to maximize the net reward. Sequential Bandits is a neural network based contextual bandit algorithm that initializes a neural network for every (subtask, LLM) combination. Figure \ref{fig:formulation} illustrates how these networks are trained using reward feedback, which is formally described in Algorithm 1.

The task is first divided into simpler subtasks. 
For the first subtask, for every available LLM, we construct an upper confidence bound (UCB) on the net reward using the neural network estimate and token length prediction models output for exploitation and their gradient for exploration (line 6), similar to \cite{neuralucbzhou20a}. Here we denote the $\ell_2$ norm by a positive definite matrix $\mathbf{A}$ by $\|\mathbf{x}\|_\mathbf{A} := \sqrt{\mathbf{x}^T\mathbf{A}\mathbf{x}}$. 
However, unlike \cite{neuralucbzhou20a}, our algorithm includes a cost sensitivity parameter $\alpha_i$ for subtask $i$, which is multiplied by the cost term $C(\mathbf{x},j)$, the predicted cost of using the chosen LLM $j$ for subtask $i$, and then subtracted from the UCB term. Setting $\alpha_i=0$ reduces to the cost-agnostic setting. 
As the value of $\alpha_i$ increases, the algorithm will prefer cheaper models, prioritizing cost over accuracy for task $i$. We define $\sigma_{h,t-1}(\mathbf{x})=\| \mathbf{g}^{t-1}_{i,j}(\mathbf{x};\boldsymbol{\theta}^{t-1}_{{i,j}})/ \sqrt{m} \| _{\mathbf{Z}^{-1}_{h,t-1}(b_{i,j})}$,
$\sigma_{c,t-1}(\mathbf{x})=\| \mathbf{v}^{t-1}_j(\mathbf{x};\boldsymbol{\Theta}_{t-1})/ \sqrt{m} \| _{\mathbf{Z}^{-1}_{c,t-1}(j)}$. $\beta_{h,t}$ and $\beta_{c,t}$ are exploration parameters (for reward and cost) that get updated over time as in \cite{neuralucbzhou20a}.

As described in Sec.~\ref{sec:formulation}, the input to the neural network includes embeddings of the description $d_j$ of each LLM $j$, as well as the incoming query $q_t$. We then choose the LLM that has the highest reward estimate (line 7) and pass its output as the input prompt for the next subtask in the pipeline (lines 8-9). We then follow these steps for each subtask (lines 11-14).
After making LLM selections, we observe their rewards and the overall super arm reward $R(S_t, \mathbf{r}_t)$ as well as the token lengths of the outputs, which are used to train the reward, cost models (line 18). We then update the weights of the chosen LLMs' corresponding networks using the observed rewards, as well as the exploration parameters (lines 17-18). 

\textbf{Incorporating Costs.} An alternative to incorporating cost into the objective would impose a cost constraint (e.g., a monetary budget per query or task). However, \textit{LLM inference costs are inherently uncertain before query execution}, due to their dependence on the number of output tokens. Thus, a hard budget constraint could favor subtasks early in the pipeline, as underestimations of their costs could then lead to severe budget constraints for later subtasks. Incorporating cost into our objective also allows us to tune $\alpha_i$ for different subtasks $i$, which may have different inherent costs: for example a summarization task may be quite expensive as its input consists of a long text to be summarized. 

\textbf{Differences from Existing Neural MAB Algorithms.} Prior neural contextual bandit algorithms \cite{neuralucbzhou20a,xu2020neural} can be na\"ively adapted to our problem setting by simply using them for each subtask's LLM selection in sequence, which we use as baselines in Sec.~\ref{exp}. However, such algorithms train and use \textit{one} neural network for reward estimation and LLM selection for each subtask, which does not account for the fact that different models imply a different inherent reward function for the different subtasks.
Experimentally, we find that Sequantial Bandits' \textit{use of a separate neural network encourages more exploration}. Using the same neural network estimator for each (subtask, arm) combination leads to more similar success estimates, so that the (deterministic) cost estimates dominate the relative ranking of the different arms is then dominated by the (deterministic) cost estimates, limiting exploration. 
We note that \textit{using multiple neural networks does not increase our training overhead}: in each round we only train the neural networks of the LLMs that we have chosen for every subtask. Thus, our required compute is no greater than the other neural contextual MAB algorithms. 



\begin{algorithm}[t] 
\caption{Sequential Bandits}
\begin{algorithmic}[1]
    \STATE Input: LLMs $\in\mathcal{M}$, descriptions $\mathbf{d}\in D$, queries $[\mathbf{q}_1,\mathbf{q}_2,\ldots,\mathbf{q}_T],$ number of gradient descent steps $J$, learning rate $\eta$, cost weights $\alpha_i$, number of rounds $T$, regularization parameter $\lambda$, initialize $\mathbf{Z}_{h,0}(b_{i,j})=\mathbf{Z}_{c,0}(j)=\lambda \mathbf{I}$ for all arms $b_{i,j}$ and all LLMs $j$.
    \FOR{$t = 1, \ldots, T$}
        \STATE Observe descriptions $\mathbf{d}\in D$ and subtasks $i\in\{1,2,...,K\}$
        \FOR{subtask $i = 1, \ldots, K$}
            \IF{$i=1$:}                \STATE $\forall j\in [N_i]$, $ u_{i,j}=f_{i,j}(\mathbf{q}_t,\mathbf{d}_j)+ \beta_{h,t}\sigma_{h,t-1}(\mathbf{q}_t)-\alpha_i C(\mathbf{q}_t,j)+\alpha_i\beta_{c,t}\sigma_{c,t-1}(\mathbf{q}_t)$
            \STATE $s_i=\text{argmax}_j(\mathbf{u}_{i})$ (LLM chosen for subtask $i$)
            \STATE Pass query $\mathbf{q}_t$ through LLM $s_i$
            \STATE Observe output $\mathbf{p}_{i+1}$ of the chosen LLM
            \ELSE
            \STATE  $\forall j\in [N_i]$, $ u_{i,j}=f_{i,j}(\mathbf{p}_i,d_j)+ \beta_{h,t}\sigma_{h,t-1}(\mathbf{p}_i) -\alpha_i C(\mathbf{p}_i,j)+\alpha_i\beta_{c,t}\sigma_{c,t-1}(\mathbf{p}_i)$
            \STATE $s_i=\text{argmax}_j(\mathbf{u}_{i})$ (LLM chosen for subtask $i$)
            \STATE Pass prompt $\mathbf{p}_{i}$ through LLM $s_i$
            \STATE Observe output $\mathbf{p}_{i+1}$ of the chosen LLM
            \ENDIF
        \ENDFOR
        \STATE For selected arms ${b_{i,j} \in S_t}$ and selected LLMs $j$, update $\mathbf{Z}_{h,t}(b_{i,j}) \gets \mathbf{Z}_{h,t-1}(b_{i,j}) + \mathbf{g}_{i,j}^t(\mathbf{g}_{i,j}^t)^T/m$, $\mathbf{Z}_{c,t}(j) \gets \mathbf{Z}_{c,t-1}(j) + \mathbf{v}_{j}^t(\mathbf{v}_{j}^t)^T/n$
        \STATE Update weights of selected arms $\boldsymbol{\theta}^{t}_{{i,j}}$ and $\boldsymbol{\Theta}_t$ for the reward and cost models by minimizing the MSE loss using gradient descent with step size $\eta$ for $J$ iterations by using the reward and cost feedback.
        
    \ENDFOR
\end{algorithmic}
\end{algorithm}

\section{Theoretical Results}
\label{sec:theory}
We establish the theoretical foundation of Sequential Bandits by first deriving high-probability confidence bounds for the neural reward (Lemma 1) and BERT-based cost model (Lemma 2) within the NTK (neural tangent kernel) regime. We then characterize the Lipschitz continuity of the optimal value function to quantify semantic error propagation (Lemma 3) through the pipeline and bound the instantaneous net reward regret (Lemma 4). These components are combined in Theorem 1 to prove a sublinear regret bound in the number of rounds $T$. We include the proofs of the lemmas and the theorem in the Appendix.
\begin{lemma}[Neural Reward Concentration]
Let $f(\mathbf{x}; \bm{\theta})$ be a neural network of width $m$ with NTK parameterization, and $g(\mathbf{x}) = \nabla_{\bm{\theta}} f(\mathbf{x}; \bm{\theta}_0)$ its initial gradient. Let $h(\mathbf{x})$ be the true underlying reward function, which lies in the Reproducing Kernel Hilbert Space (RKHS) $\mathcal{H}_{K}$ induced by the NTK, with $\|h\|_{\mathcal{H}_K} \le S$. 
For any $\delta \in (0,1)$ and $Q>0$, if the width $m$ is sufficiently large, then with probability at least $1-\delta$, the prediction error for any $\mathbf{x}$ at round $t$ is bounded by:
\begin{equation}
    |f(\mathbf{x}; \bm{\theta}_{t-1}) - h(\mathbf{x})| \le \beta_{h,t} \underbrace{\sqrt{\mathbf{g}(\mathbf{x})^\top \mathbf{Z}_{t-1}^{-1} \mathbf{g}(\mathbf{x}) / m}}_{\sigma_{t-1}(\mathbf{x})} + \epsilon(m)
\end{equation}
where $\mathbf{Z}_{t-1} = \lambda \mathbf{I} + \sum_{\tau=1}^{t-1} \mathbf{g}(\mathbf{x}_\tau) \mathbf{g}(\mathbf{x}_\tau)^\top / m$, $\epsilon(m) = \mathcal{O}(m^{-1/6})$ is the linearization error, and $\beta_{h,t} = Q \sqrt{\log(\det \mathbf{Z}_t / \det \lambda \mathbf{I}) - 2 \log \delta} + \lambda^{1/2} S$.
\end{lemma}

\begin{lemma}[BERT Cost Estimation Confidence Bound]
This lemma is an analogous version of Lemma 1 to the BERT cost model instead of a neural network. We include the full Lemma statement and its proof in the Appendix.
\end{lemma}

\begin{lemma}[Lipschitz Continuity and Context Propagation]\label{lem:lipschitz_prop}
Suppose the following hold for all pipeline subtasks $i$:
\begin{enumerate}
    \item $h_i(\mathbf{x}, a)$ is $L_h$-Lipschitz with respect to $\mathbf{x}$: $|h_i(\mathbf{x}, a) - h_i(\mathbf{x}', a)| \le L_h \|\mathbf{x} - \mathbf{x}'\|_2$.
    \item $c_i(\mathbf{x},a)$ is $L_c$-Lipschitz with respect to $\mathbf{x}$: $|c_i(\mathbf{x}, a) - c_i(\mathbf{x}', a)| \le L_c \|\mathbf{x} - \mathbf{x}'\|_2$
    \item $o_i(\mathbf{x}, a)$ is $L_o$-Lipschitz with respect to $\mathbf{x}$: $\|o_i(\mathbf{x}, a) - o_i(\mathbf{x}', a)\|_2 \le L_o \|\mathbf{x}- \mathbf{x}'\|_2$.
\end{enumerate}
Then, the value function $V_i(\mathbf{x}_i)$ is Lipschitz continuous in the initial context $\mathbf{x}_i.$ Hence the maximum regret contributed by a context mismatch $\Delta \mathbf{x}_i =\| \mathbf{x}_i-\mathbf{x}_i^* \|_2$ at stage $i$ satisfies $|V_i(\mathbf{x}_i)-V_i(\mathbf{x}_i^*)| \leq \left(L_h+\max_i(\alpha_i) L_c\right) \sum_{j=0}^{K-i}L_o^j\|\mathbf{x}_i-\mathbf{x}_i^*\|_2$.


\end{lemma}
Assumptions 1-2 of this lemma assumes that the LLM performance (accuracy) and token consumption are stable with respect to the input prompt. 
Practically, this means a small change in the semantic content of a prompt should not lead to a greatly different reward or a large difference in output length. Assumption 3 states that the semantic mapping from one stage in the pipeline to the next is non-expansive. 
As this embedding model is specifically trained to map similar sentences to nearby points in a vector space (we use Sentence Transformer embeddings in our experiments), a small deviation in the context $\mathbf{x}$ (the current prompt) results in a bounded change in the context $o_i(\mathbf{x}, a)$ (the next prompt).

\begin{lemma}[Single-Step Net Reward Regret]
Let the selection rule be:
\begin{equation*}
    a_{t,i} = \arg \max_{a \in \mathcal{A}_i} \left\{ \underbrace{f_i(\mathbf{x}_{t,i}, a; \theta_{t-1}) + \beta_{h,t} \sigma_{h,t}(a)}_{UCB_{reward}} - \alpha_i \underbrace{(\hat{c}_i(\mathbf{x}_{t,i}, a) - \beta_{c,t} \sigma_{c,t}(a))}_{LCB_{cost}} \right\}
\end{equation*}
Then, conditioned on the high probability events in Lemmas 1, 2 and 3, the instantaneous net reward regret, where $a_{t,i}^* = \arg\max_a n_i(\mathbf{x}_{t,i}, a)$, is bounded by:
\begin{equation*}
    y_{t,i} = n_i(\mathbf{x}_{t,i}, a_{t,i}^*) - n_i(\mathbf{x}_{t,i}, a_{t,i}) \le 2 \beta_{h,t} \sigma_{h,t}(a_{t,i}) + 2 \alpha_i \beta_{c,t} \sigma_{c,t}(a_{t,i}) + 2(\epsilon_m + \alpha_i \epsilon_{n}).
\end{equation*}
\end{lemma}

\begin{theorem}[High-probability sublinear regret]
\label{thm:pipeline_regret_dynamic_oracle}
%
%
Assume the high probability events of Lemma 4 holds for all rounds and stages,
and that the conditions required in Lemma~5.4 in NeuralUCB \cite{neuralucbzhou20a} are satisfied. Define the global effective dimension as $d_{eff}=\max(\max_{i,j}\{d^h_{i,j}\},d^c)$ where $d_{i,j}^h$ and $d^c$ are respectively the effective dimensions of the NTK associated with the reward model for LLM $j$ at stage $i$ and the cost regression model. Then the total regret scales as follows:
\begin{align*}
    R_T= \tilde{O}(Kd_{\text{eff}}\sqrt{T}).
\end{align*}

\end{theorem}

\section{Experiments}\label{exp}
In this section, we present the results of our proposed Sequential Bandits algorithm on two use cases: medical diagnosis prediction and telecommunications question answering tasks. We compare our algorithm with the following baselines: (1) \textbf{Random}, which randomly selects an LLM for each subtask; (2) \textbf{Llama} or \textbf{Tele}, which always selects Llama or Tele for the given subtasks (Llama is chosen as it is the best performing model across tasks for the medical setting and Tele is the best for the telecom setting); (3) \textbf{Cost-Aware (CA) NeuralUCB} \cite{neuralucbzhou20a}, which is the cost sensitive version of NeuralUCB that uses a neural network for each subtask's reward prediction and adds the same weighted cost term in the objective as used in Sequential Bandits; and (4) \textbf{CA NeuralLinUCB} \cite{xu2020neural}, the cost-sensitive version of NeuralLinUCB, which makes use of neural networks for each subtask to learn representations of the contexts, and applies a linear model on these learned features to predict rewards, then adds the weighted cost term to the objective; (5) \textbf{CA NeuralUCB Joint}, which makes use of neural networks for each subtask as (3) but selects LLMs for subtasks all at once rather than sequentially, and (6) \textbf{FrugalGPT} \cite{chen2023frugalgptuselargelanguage}, which is a offline cascade-construction algorithm that we integrated with our own online evaluation pipeline (hence how we generate cumulative reward curves), and datasets to ensure fair comparison (more details in Appendix). Comparisons to these algorithms respectively demonstrate the value of (1) having a non-static algorithm that learns over time, (2) intelligently selecting a LLM for each subtask, (3,4) using a separate neural network to predict each LLM's performance on each subtask, (5) sequentially selecting LLMs, and (6) having an online learning algorithm.

\subsection{Experiment Settings}
We evaluate Sequential Bandits on two \textbf{datasets}:
one created from MIMIC-III (see the next subsection), a comprehensive clinical database containing de-identified health-related data from over 40,000 critical care patients \cite{johnson2016mimic}; and TeleQnA, a dataset comprising 10,000 multiple-choice questions designed to assess LLMs' knowledge in telecommunications \cite{maatouk2023teleqna}. We thus assess our framework's performance across diverse, domain-specific tasks. 
We use pipelines with 2 and 3 subtasks respectively for the medical and telecommunications datasets. 

\textbf{Subtasks.} Our two subtasks on the medical dataset are (1) a summarizer that summarizes the long medical report, whose summary is then fed to (2) a diagnoser that gives a diagnosis based on the summary. 
The \textbf{rewards} for the summarization subtask are obtained using an evaluation LLM that is fed the prompt, context, and benchmark. The reward for the diagnosis subtask given the summarized report compares the diagnoses that the LLM outputs to the actual diagnoses of the patient, which also serves as the super arm reward. 
For example, if a patient has two diagnoses, and the second subtask's LLM correctly predicts one of the two diagnoses of the patient, we assign a reward of 0.5.

For TeleQnA, we have a 3 subtask structure. We start with a summary subtask with a similar reward metric as the medical dataset's summary task. The second subtask is to answer the question, for which we compute the LLM rewards by comparing their output choice (among the 4-5 options) to the correct one. The third subtask was to explain the answer obtained from the previous subtask, whose reward is obtained by comparison to TeleQnA's explanation benchmark. 

\textbf{Cost Prediction.} 
For the output token length prediction model, which is used to construct the expected cost ($C(\mathbf{x},j)$ in Algorithm 1) of LLM query execution, we first train an offline Bert regression model on the LMSYS-Chat-1M dataset \cite{zheng2023lmsys} using L1 loss, following \cite{qiu2024efficient}. This model is also updated online as shown in Algorithm 1. We provide more information about the offline training of this model and the loss curves for offline and online training in the Appendix.

All of the \textbf{models} we used in our experiments were deployed on Microsoft Azure. The models include a combination of base models (GPT-3.5-turbo, GPT-4.1 mini, GPT-4o, Llama-3.3-70B-instruct, Mistral-3B, Phi-4), finetuned models that were finetuned using Azure on general medical and telecom knowledge, and GPT-3.5-turbo assistants that used file search prompted with relevant field knowledge. The models were selected in a way such that there are low performing/SLMs, base models, and finetuned/custom domain knowledge possessing LLMs. However, as the LMSYS-Chat-1M dataset \cite{zheng2023lmsys}, which we use to train our token length prediction model, does not contain the Mistral and Phi models, we omit them in the experiments presented in here but include them in the cost-agnostic ($\boldsymbol{\alpha}=\mathbf{0}$) results presented in the Appendix. 
The models ordered from cheapest to most expensive are: GPT-4.1 mini, Llama 3.3, Med, Tele, Med III.

The n2c2 smoking dataset \cite{uzuner2007evaluating} was used to fine-tune the Med model while Med III, Tele were respectively fine-tuned with the MIMIC-III, TeleQnA datasets. \textbf{Evaluations} were performed using the Azure evaluator. The fine-tuned models generally performed better than base models on the task they were tuned on. 

\begin{figure*}[t!]
    \centering
    \includegraphics[width=\textwidth]{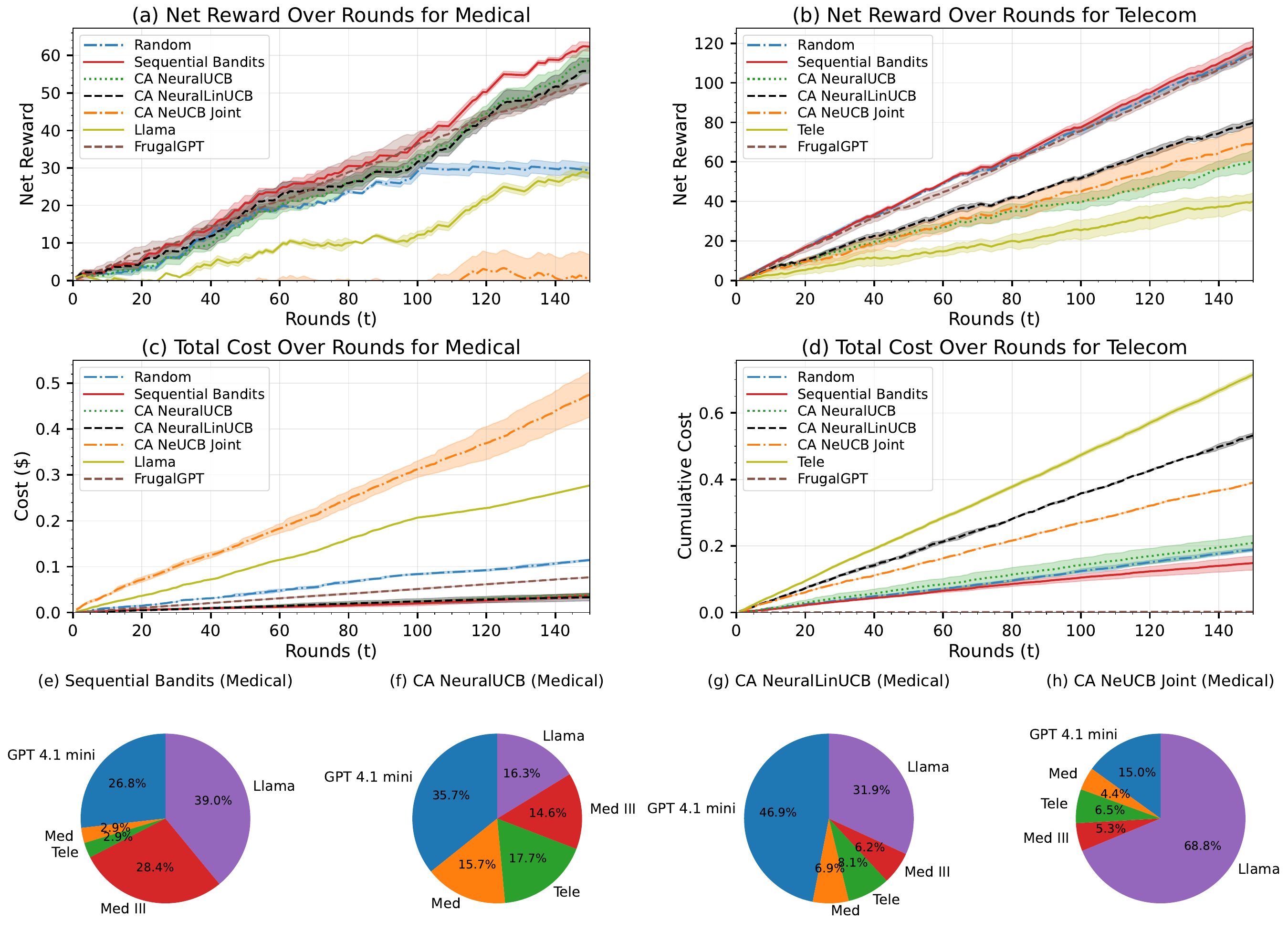}
    \caption{(a) Net reward of our algorithm compared with other baselines for the medical pipeline (b) Net reward of our algorithm compared with other baselines for the telecom pipeline (c) Total cost incurred by the algorithms for the LLM selections they make for the medical pipeline (d) Total cost incurred by the algorithms for the LLM selections they make for the telecom pipeline (e)-(h) Model selections for the diagnosis subtask in medical experiments} 
    \label{fig:net_utility_cost}
\end{figure*}
\subsection{Diagnosis Prediction Dataset}
There are some widely available medical datasets that include medical reports for de-identified patients. However, to our knowledge, there is no available dataset that is specifically tailored towards diagnosis prediction from medical reports. Hence, we developed our own dataset from MIMIC III \cite{johnson2016mimic} to be able to assess Sequential Bandits' performance. We include an explanation of how this dataset was constructed as well as a comprehensive list of all diagnoses in the Appendix along with an example report. The full dataset is available in our released code.

\subsection{Experimental Results}
We now present the experimental results for our medical diagnosis dataset and the TeleQnA dataset. All algorithms were run ten times and five times respectively for the medical and telecommunication settings, with tuned hyperparameters. The shaded regions in Figure~\ref{fig:net_utility_cost} indicate standard deviations. 


\textbf{Net Reward and Cost}. We first show that \textit{Sequential Bandits obtains a higher net reward than baseline algorithms on both pipelined settings}. For the medical setting, we consider the reward to be the reward of the final subtask in the pipeline (i.e., diagnosis accuracy) while for the telecom setting, the reward is the sum of the reward for the explanation and multiple choice answer. Figure \ref{fig:net_utility_cost} shows that our algorithm (solid red lines) achieves the highest net reward and also among the lowest costs, in both settings. When comparing the final net reward, Sequential Bandits has a \%7.20 improvement over the most competitive baseline (CA NeuralUCB) in the medical setting while displaying a \%7.27 improvement over the most competitive baseline (Random) in the telecom setting. FrugalGPT obtains the lowest cost among all baselines for the telecom setting, mainly because it doesn't use a pipeline, but it still gets a lower net reward than SeqBandits. For the telecom dataset, the Tele baseline performs the \textit{worst} among all in terms of net reward, despite the fact that it is fine-tuned on this dataset, since it incurs a much higher cost $(\approx\$0.6)$. Random performs poorly in the medical setting while it surprisingly performs well in the telecom setting. 
\textbf{Analyzing LLM Selections.} We finally take a closer look at which LLMs are selected by each algorithm for the medical diagnosis subtask and telecom explanation subtask respectively. Figure \ref{fig:net_utility_cost}(e) shows that Sequential Bandits selects Llama, Med III and GPT 4.1 mini the most often (\%39, \%28.4, \%26.8 respectively), which enables it to have high net reward and low cost as these are cheap models, and Llama and GPT 4.1 mini also have the highest and second highest accuracy for this subtask respectively. Even though Med III is also selected often, it gives less verbose outputs and just its predicted diagnosis, contributing to low cost. The other baselines, as shown in Figure \ref{fig:net_utility_cost}(f)-(h), select these models less frequently and make more suboptimal choices such as selecting the Med model more often, which has the lowest accuracy. CA NeUCB Joint selects similar models to our algorithm for diagnosis, but it selects different summarizers from Sequential Bandits. Additional pie charts for the telecommunications and medical settings, indicating the model choices, can be found in the Appendix.


\textbf{Additional Experiments.} We include more experimental results in the Appendix, including (i) the regret achieved by each algorithm and model accuracies on subtasks. We also present results for (ii) a 3 subtask version of the 2 subtask medical pipeline, and (iii) cost-agnostic $(\boldsymbol{\alpha} = 0)$ versions of the Medical Diagnosis task, as well as a single-subtask TeleQnA task. Since the cost can also be interpreted as response latency, we also characterize model response latencies. 


\section{Conclusion}\label{sec:conclusion}
In this paper, we introduced a novel approach to selecting a pipeline of LLMs for executing subtasks, employing a contextual MAB algorithm that sequentially chooses the best LLM for each subtask in an online manner. Our approach, Sequential Bandits, leverages neural networks and a BERT regression model to predict the expected success and cost of each LLM, thereby enabling more effective task completion across a sequence of dependent subtasks.  We demonstrated the effectiveness of our method through experiments involving medical diagnosis prediction and multiple choice telecommunications question answering. We also established a formal theoretical approach for this sequential decision framework by proving that our algorithm achieves a sublinear regret of $\tilde{O}(Kd_{\text{eff}}\sqrt{T})$. Future work includes handling a broader range of tasks with varying degrees of difficulty and interdependencies. 

\begin{credits}

\subsubsection{\discintname}
The authors have no competing interests to declare that are relevant to the content of this article.
\end{credits}
%
%
%
\bibliographystyle{splncs04}
\bibliography{references}
\pagebreak
\section{Appendix}

\subsection{Proofs}

\begin{proof}(Lemma 1)\\
Following the analysis of over-parameterized networks in \cite{caogu2019}, for a sufficiently wide network, the weights $\bm{\theta}_{t-1}$ remain in a small neighborhood of the initialization $\bm{\theta}_0$. We can express the network output via a first-order Taylor expansion:
\begin{equation}
    f(\mathbf{x}; \bm{\theta}_{t-1}) = f(\mathbf{x}; \bm{\theta}_0) + \langle \mathbf{g}(\mathbf{x}), \bm{\theta}_{t-1} - \bm{\theta}_0 \rangle + \epsilon_{lin}
\end{equation}
where $|\epsilon_{lin}| \le \mathcal{O}(m^{-1/6})$. In the NTK regime, the network behaves as a linear model $f^{lin}(\mathbf{x}) = \langle \bm{\phi}(\mathbf{x}), \mathbf{w}_{t-1} \rangle$ where $\bm{\phi}(\mathbf{x}) = \mathbf{g}(\mathbf{x})/\sqrt{m}$ and $\mathbf{w}_{t-1} = \sqrt{m}(\bm{\theta}_{t-1} - \bm{\theta}_0)$.

By the RKHS assumption, the true function $h(\mathbf{x})$ can be approximated by a linear form in the feature space $\bm{\phi}(\mathbf{x})$. Let $\mathbf{w}^*$ be the vector such that $h(\mathbf{x}) = \langle \bm{\phi}(\mathbf{x}), \mathbf{w}^* \rangle + \epsilon_{approx}$. 
Since the algorithm updates $\bm{\theta}$ using a squared loss with $\ell_2$ regularization (Algorithm 1, Lines 17-18), the dynamics are equivalent to performing online ridge regression on the features $\bm{\phi}(\mathbf{x})$ \cite{neuralucbzhou20a}. 
Applying Theorem 2 from \cite{improved_yadkori}, the parameter error is bounded within a confidence ellipsoid:
\begin{equation}
    \|\mathbf{w}_{t-1} - \mathbf{w}^*\|_{\mathbf{Z}_{t-1}} \le \beta_t
\end{equation}
where $\|\mathbf{u}\|_{\mathbf{Z}} = \sqrt{\mathbf{u}^\top \mathbf{Z} \mathbf{u}}$ is the weighted norm.

To bound the prediction error $| \langle \bm{\phi}(\mathbf{x}), \mathbf{w}_{t-1} - \mathbf{w}^* \rangle |$, we apply the generalized Cauchy-Schwarz inequality for the positive definite matrix $\mathbf{Z}_{t-1}$:
\begin{equation}
    | \langle \bm{\phi}(\mathbf{x}), \mathbf{w}_{t-1} - \mathbf{w}^* \rangle | \le \|\bm{\phi}(\mathbf{x})\|_{\mathbf{Z}_{t-1}^{-1}} \cdot \|\mathbf{w}_{t-1} - \mathbf{w}^*\|_{\mathbf{Z}_{t-1}}
\end{equation}
Substituting the bound from Step 2:
\begin{equation}
    | \langle \bm{\phi}(\mathbf{x}), \mathbf{w}_{t-1} - \mathbf{w}^* \rangle | \le \beta_t \sqrt{\bm{\phi}(\mathbf{x})^\top \mathbf{Z}_{t-1}^{-1} \bm{\phi}(\mathbf{x})}
\end{equation}
Since $\bm{\phi}(\mathbf{x}) = \mathbf{g}(\mathbf{x})/\sqrt{m}$, this term is exactly $\beta_t \sigma_{t-1}(\mathbf{x})$.

By the triangle inequality:
\begin{align*}
    |f(\mathbf{x}; \bm{\theta}_{t-1}) - h(\mathbf{x})| &\le |f^{lin}(\mathbf{x}) - h(\mathbf{x})| + |f(\mathbf{x}; \bm{\theta}_{t-1}) - f^{lin}(\mathbf{x})| \\
    &\le \beta_t \sigma_{t-1}(\mathbf{x}) + \epsilon_{lin} + \epsilon_{approx}
\end{align*}
Combining the linearization and approximation residuals into $\epsilon(m) = \mathcal{O}(m^{-1/6})$, we obtain the desired bound.
\end{proof}

\textbf{Lemma 2 (BERT Cost Estimation Confidence Bound).}
\textit{Let $C(\mathbf{x}; \bm{\Theta}_t)$ be the BERT-based cost regression model with width $n$ and parameters $\bm{\Theta}_t$ at round $t$. Let $c(\mathbf{x})$ and $c^*(\mathbf{x})$ be the true underlying cost function in terms of the monetary cost and number of output tokens respectively $($i.e. $c(\mathbf{x})\leq P_ac^*(\mathbf{x}))$ and $\mathbf{v}(\mathbf{x}) = \nabla_{\bm{\Theta}} C(\mathbf{x}; \bm{\Theta}_0)$. 
We assume the following standard conditions:
\begin{enumerate}
    \item The architecture of $C(\mathbf{x}; \bm{\Theta})$, which includes self-attention and layer normalization layers, admits a deterministic NTK limit $\Theta_{\text{BERT}}$ as $n \to \infty$ \cite{yang2020tensorprogramsiineural}.
    \item $c(\mathbf{x})$ has a bounded norm in the RKHS induced by $\Theta_{\text{BERT}}$: $\|c\|_{\mathcal{H}_{\text{BERT}}} \le S_c$.
    \item The model parameters remain close to initialization, $\|\bm{\Theta}_t - \bm{\Theta}_0\|_2 \le \omega$, where $\omega$ is sufficiently small.
\end{enumerate}
Then, for any $\delta \in (0, 1)$ and $Q_c>0$, if the width $n$ is sufficiently large, then with probability at least $1-\delta$, for any context $\mathbf{x}$ at round $t$: 
\begin{equation}
    |C(\mathbf{x}; \bm{\Theta}_{t-1}) - c(\mathbf{x})| \le P_a \left( \beta_{c,t} \sqrt{\frac{\mathbf{v}(\mathbf{x})^\top \mathbf{V}_{t-1}^{-1} \mathbf{v}(\mathbf{x})}{n}} + \epsilon(n) \right)
\end{equation}
where $\mathbf{V}_{t-1} = \lambda \mathbf{I} + \sum_{\tau=1}^{t-1} \mathbf{v}(\mathbf{x}_\tau) \mathbf{v}(\mathbf{x}_\tau)^\top / n$ is the covariance matrix of the normalized gradients, $\epsilon(n) = \mathcal{O}(n^{-1/6})$ is the linearization error, and $\beta_{c,t}$ is the confidence radius $\beta_{c,t} = Q_c \sqrt{\log(\det \mathbf{V}_{t}/\det \lambda \mathbf{I}) - 2 \log \delta} + \lambda^{1/2} S_c$.}

\begin{proof} (Lemma 2)
The proof extends on the \textit{Tensor Programs} framework \cite{yang2020tensorprogramsiineural} which rigorously extends NTK results to complex architectures like Transformers.

The BERT model consists of Multi-Head Self-Attention, Layer Normalization, and MLP blocks. Yang \cite{yang2020tensorprogramsiineural} proves that randomly initialized neural networks of any "standard architecture" (Definition 7.1) have an NTK that converges almost surely to a deterministic limit as the width $n \to \infty$. Specifically, Corollary 7.3 and Appendix E.5 of \cite{yang2020tensorprogramsiineural} derive the specific Tensor Program operations for the Attention mechanism (Softmax, Query/Key/Value projections), showing they yield deterministic limits. This justifies treating the infinite-width BERT as a kernel method governed by $\Theta_{\text{BERT}}$.

In the over-parameterized regime (large $n$), the training dynamics of the network under gradient descent can be approximated by its linearization around initialization $\bm{\theta}_0$:
\begin{equation}
    C(\mathbf{x}; \bm{\Theta}_t) \approx C(\mathbf{x}; \bm{\Theta}_0) + \langle \nabla_{\bm{\Theta}} C(\mathbf{x}; \bm{\Theta}_0), \bm{\Theta}_t - \bm{\Theta}_0 \rangle
\end{equation}
The approximation error $\epsilon_{lin} = |C(\mathbf{x}; \bm{\Theta}_t) - C^{lin}(\mathbf{x}; \bm{\Theta}_t)|$ is bounded by $\mathcal{O}(n^{-1/6})$ \cite{caogu2019}. This allows us to analyze the learning process as online ridge regression using the fixed feature map $\bm{\phi}(\mathbf{x}) = \mathbf{v}(\mathbf{x})/\sqrt{n}$.

By the RKHS assumption, there exists an ideal weight vector $\bm{\theta}^*$ such that $c^*(\mathbf{x}) \approx \langle \bm{\phi}(\mathbf{x}), \bm{\Theta}^* \rangle$. Applying the self-normalized concentration inequality for vector-valued martingales (Theorem 1 in \cite{improved_yadkori}), we have with high probability:
\begin{equation}
    \|\bm{\Theta}_{t-1} - \bm{\Theta}^*\|_{\mathbf{V}_{t-1}} \le \beta_{c,t}
\end{equation}

We bound the prediction error using the Cauchy-Schwarz inequality for the positive definite matrix $\mathbf{V}_{t-1}$:
\begin{align*}
    | \langle \bm{\phi}(\mathbf{x}), \bm{\Theta}_{t-1} - \bm{\Theta}^* \rangle | &\le \|\bm{\phi}(\mathbf{x})\|_{\mathbf{V}_{t-1}^{-1}} \cdot \|\bm{\Theta}_{t-1} - \bm{\Theta}^*\|_{\mathbf{V}_{t-1}} \\
    &\le \sqrt{\bm{\phi}(\mathbf{x})^\top \mathbf{V}_{t-1}^{-1} \bm{\phi}(\mathbf{x})} \cdot \beta_{c,t}
\end{align*}
Substituting $\bm{\phi}(\mathbf{x}) = \mathbf{v}(\mathbf{x})/\sqrt{n}$, the statistical error term becomes $\beta_{c,t} \sqrt{\mathbf{v}(\mathbf{x})^\top \mathbf{V}_{t-1}^{-1} \mathbf{v}(\mathbf{x}) / n}$.

Combining the linearization error $\epsilon_{lin}$ and the statistical estimation error, and scaling by the price factor $P_a$, we obtain the final bound:
\begin{equation}
    |C(\mathbf{x}; \bm{\Theta}_{t-1}) - c(\mathbf{x})| \le P_a \left( \beta_{c,t} \sqrt{\frac{\mathbf{v}(\mathbf{x})^\top \mathbf{V}_{t-1}^{-1} \mathbf{v}(\mathbf{x})}{n}} + \epsilon(n) \right)
\end{equation}
\end{proof}

\begin{proof} (Lemma 3)\\
We prove the propagation bound via induction on the remaining subtasks $k = K - i$.

\paragraph{Base Case ($i=K$):} 
For the final subtask in the pipeline, there are no future transitions hence $V_K(\mathbf{x}_K)=n_K(\mathbf{x}_K, a_K^*)$. By the assumption of $L_h$-Lipschitzness:
\begin{align*}
    |V_K(\mathbf{x}_K) - V_K(\mathbf{x}^*_K)| =|n_K(\mathbf{x}_K,a_K^*)-n_K(\mathbf{x}_K^*,a_K^*) | \le L_n \|\mathbf{x}_K - \mathbf{x}^*_K\|_2
\end{align*}
This matches the lemma's formula for $i=K$ since $\sum_{j=0}^0L_o^j=1$.

\paragraph{Inductive Step:}
Assume the bound holds for subtask $i+1$. That is, there is $L_{\text{total},i+1}=L_n\sum_{j=0}^{K-i-1}L_o^j$ such that 
\begin{align*}
    |V_{i+1}(\mathbf{x}_{i+1})-V_{i+1}(\mathbf{x}_{i+1}^*)|\leq L_{\text{total},i+1} \|\mathbf{x}_{i+1}-\mathbf{x}_{i+1}^*\|_2    
\end{align*}

\paragraph{Analyzing Stage $i$:}
The value function $V_i(\mathbf{x}_i)$ can be decomposed as $V_i(\mathbf{x}_i)=n_i(\mathbf{x}_i,a_i^*)+V_{i+1}(\mathbf{x}_{i+1})$. Then it follows that:
\begin{align*}
    |V_i(\mathbf{x}_i)-V_i(\mathbf{x}_i^*)| = |n_i(\mathbf{x}_i,a_i^*)+V_{i+1}(\mathbf{x}_{i+1})-n_i(\mathbf{x}_i^*,a_i^*)-V_{i+1}(\mathbf{x}_{i+1}^*)|
\end{align*}
Given the Lipschitz assumptions for $h_i$ and $c_i$, since $n_i(\mathbf{x},a)=h_i(\mathbf{x},a)-\alpha_i c_i(\mathbf{x},a)$, it follows that $n_i$ is Lipschitz with constant $L_n=L_h+\max_i(\alpha_i) L_c$. Now we apply triangle inequality to get
\begin{align*}
    |V_i(\mathbf{x}_i)-V_i(\mathbf{x}_i^*)| \le |n_i(\mathbf{x}_i,a_i^*)-n_i(\mathbf{x}_i^*,a_i^*)|+|V_{i+1}(\mathbf{x}_{i+1})-V_{i+1}(\mathbf{x}_{i+1}^*)| \\
    |V_i(\mathbf{x}_i)-V_i(\mathbf{x}_i^*)| \leq L_n\|\mathbf{x}_i-\mathbf{x}_i^*\|_2+L_{\text{total},i+1}\|\mathbf{x}_{i+1}-\mathbf{x}_{i+1}^*\|_2
\end{align*}
where we use the Lipschitzness of $n_i$ and the inductive hypothesis to bound the value function deviation. Because the next context is generated by the LLM transition function we have that 
\begin{align*}
    \|\mathbf{x}_{i+1}-\mathbf{x}_{i+1}^*\|_2 = \|o_i(\mathbf{x}_i,a_i^*)-o_i(\mathbf{x}_i^*,a_i^*)\|_2\leq L_o\|\mathbf{x}_i-\mathbf{x}_i^*\|_2
\end{align*}
hence it follows that 
\begin{align*}
    |V_i(\mathbf{x}_i)-V_i(\mathbf{x}_i^*)|\leq (L_n+L_{\text{total},i+1}L_o)\|\mathbf{x}_i-\mathbf{x}_i^*\|_2
\end{align*}
Now analyzing the constant gives 
\begin{align*}
    L_n+L_o(L_n\sum_{j=0}^{K-i-1}L_o^j)=L_n(1+\sum_{j=0}^{K-i-1}L_o^{j+1})=L_n\sum_{j=0}^{K-i}L_o^j
\end{align*}
which is exactly what we were trying to show. This completes the proof, showing that the error propagation remains bounded by a geometric series wighted by the inital context perturbation.

\end{proof}

\begin{proof} (Lemma 4)\\
Let $U_t(\mathbf{x}, a) = (f_i + \beta_{h,t} \sigma_{h,t}) - \alpha_i (\hat{c}_i - \beta_{c,t} \sigma_{c,t})$ be the index used for selection. 

By the results of Lemma 1 and Lemma 2, we have with high probability:
\begin{align*}
    h_i(\mathbf{x}, a) &\le f_i(\mathbf{x}, a; \theta_{t-1}) + \beta_{h,t} \sigma_{h,t}(a) + \epsilon_m \\
    c_i(\mathbf{x}, a) &\ge \hat{c}_i(\mathbf{x}, a) - \beta_{c,t} \sigma_{c,t}(a) - \epsilon_{n}
\end{align*}
Combining these, the true net reward is upper-bounded by the selection index plus approximation errors:
\begin{equation}
    n_i(\mathbf{x}_{t,i}, a) \le U_t(\mathbf{x}_{t,i}, a) + (\epsilon_m + \alpha \epsilon_{n})
\end{equation}
This implies that for the optimal action $a_{t,i}^*$:
\begin{equation}
    n_i(\mathbf{x}_{t,i}, a_{t,i}^*) \le U_t(\mathbf{x}_{t,i}, a_{t,i}^*) + (\epsilon_m + \alpha \epsilon_{n})
\end{equation}
By the definition of the selection rule $a_{t,i} = \arg\max_a U_t(\mathbf{x}_{t,i}, a)$, we know:
\begin{equation}
    U_t(\mathbf{x}_{t,i}, a_{t,i}^*) \le U_t(\mathbf{x}_{t,i}, a_{t,i})
\end{equation}
Thus, the true optimal net reward is bounded by the index of the \textit{chosen} action:
\begin{equation}
    n_i(\mathbf{x}_{t,i}, a_{t,i}^*) \le U_t(\mathbf{x}_{t,i}, a_{t,i}) + (\epsilon_m + \alpha \epsilon_{n})
\end{equation}
Conversely, the true net reward of the chosen action $a_{t,i}$ can be lower-bounded using the same concentration lemmas:
\begin{equation}
    n_i(\mathbf{x}_{t,i}, a_{t,i}) \ge (f_i - \beta_{h,t} \sigma_{h,t}) - \alpha_i (\hat{c}_i + \beta_{c,t} \sigma_{c,t}) - (\epsilon_m + \alpha_i \epsilon_{n})
\end{equation}
This can be rewritten in terms of the selection index $U_t$:
\begin{equation}
    n_i(\mathbf{x}_{t,i}, a_{t,i}) \ge U_t(\mathbf{x}_{t,i}, a_{t,i}) - 2\beta_{h,t} \sigma_{h,t}(a_{t,i}) - 2\alpha_i \beta_{c,t} \sigma_{c,t}(a_{t,i}) - (\epsilon_m + \alpha_i \epsilon_{n})
\end{equation}
The instantaneous regret is $y_{t,i} = n_i(\mathbf{x}_{t,i},a_{t,i}^*) - n_i(\mathbf{x}_{t,i},a_{t,i})$. Substituting the upper bound and the lower bound gives:
\begin{align*}
    y_{t,i} &\le [U_t(\mathbf{x}_{t,i},a_{t,i}) + \epsilon_{total}] - [U_t(\mathbf{x}_{t,i},a_{t,i}) - 2\beta_{h,t} \sigma_{h,t} - 2\alpha_i \beta_{c,t} \sigma_{c,t} - \epsilon_{total}] \\
    y_{t,i} &\le 2\beta_{h,t} \sigma_{h,t}(a_{t,i}) + 2\alpha_i \beta_{c,t} \sigma_{c,t}(a_{t,i}) + 2(\epsilon_m + \alpha_i \epsilon_{n})
\end{align*}
where $\epsilon_{total} = \epsilon_m + \alpha_i \epsilon_{n}$.
\end{proof}

\begin{proof} (Theorem 1)\\
Given the optimal value function definition $V_i(\mathbf{x})$, define the instantaneous regret at time $t$ as $y_t=V_1(\mathbf{x}_{t,1})-\sum_{i=1}^Kn_i(\mathbf{x}_{t,i},a_{t,i})$. Since $\mathbf{x}_{t,i+1}=o_i(\mathbf{x}_{t,i},a_{t,i})$ and $V_{K+1}(\mathbf{x})=0$ this telescopes:
\begin{align*}
    y_t = \sum_{i=1}^K[\underbrace{V_i(\mathbf{x}_{t,i})-n_i(\mathbf{x}_{t,i},a_{t,i})-V_{i+1}(\mathbf{x}_{t,i+1})}_{\Delta_{t,i}}]
\end{align*}
We can expand the first term to get 
\begin{align*}
    \Delta_{t,i}=\underbrace{n_i(\mathbf{x}_{t,i},a_{t,i}^*)-n_{i}(\mathbf{x}_{t,i},a_{t,i})}_{\text{immediate selection regret}}+\underbrace{V_{i+1}(o_i(\mathbf{x}_{t,i},a_{t,i}^*))-V_{i+1}(\mathbf{x}_{t,i+1})}_{\text{context propagation regret}}
\end{align*}
From Lemmas 1, 2, and 4 we can bound the immediate selection regret as follows:
\begin{align*}
    n_i(\mathbf{x}_{t,i},a_{t,i}^*)-n(\mathbf{x}_{t,i},a_{t,i})\leq 2\beta_{h,t}\sigma_{h,t}(a_{t,i})+2\alpha_i\beta_{c,t}\sigma_{c,t}(a_{t,i})+2\epsilon_{\text{total}}
\end{align*}
The context propagation regret term can be bounded using Lemma \ref{lem:lipschitz_prop}, hence since $\mathbf{x}_{t,i+1}=o_i(\mathbf{x}_{t,i},a_{t,i})$, $L_n = L_h+\max_i(\alpha_i)L_c$
\begin{align*}
   \text{context prop. regret}\leq  \left(L_n\sum_{j=0}^{K-i
   -1}L_o^j\right)\|o_i(\mathbf{x}_{t,i},a_{t,i}^*)-o_i(\mathbf{x}_{t,i},a_{t,i})\|_2
\end{align*}
Now to bound the distance between the two output embeddings $\|o_i(\mathbf{x}_{t,i},a_{t,i}^*)-o_i(\mathbf{x}_{t,i},a_{t,i})\|_2$ we make the assumption that the semantic distance between two generated outputs for task $i$ is bounded proportionally by the difference in their task utility (net reward). Formally, defining an alignment constant $C_{\text{align}}$ we have $\|o_i(\mathbf{x}_{t,i},a_{t,i}^*)-o_i(\mathbf{x}_{t,i},a_{t,i})\|_2\leq C_{\text{align}}|n_i(\mathbf{x}_{t,i},a_i^*)-n_i(\mathbf{x}_{t,i},a_{t,i})|$ which means that 
\begin{align*}
    \text{context prop. regret}\leq  \left(C_{\text{align}}L_n\sum_{j=0}^{K-i-1}L_o^j\right)\left(2\beta_{h,t}\sigma_{h,t}+2\alpha_i\beta_{c,t}\sigma_{c,t}+2\epsilon_{\text{total}}\right)\\
    \Delta_{t,i} \leq \left(1+C_{\text{align}}L_n\sum_{j=0}^{K-i-1}L_o^j\right)\left(2\beta_{h,t}\sigma_{h,t}+2\alpha_i\beta_{c,t}\sigma_{c,t}+2\epsilon_{\text{total}}\right)\\
    \Delta_{t,i}\leq 2C_{\text{Lip}}\left(\beta_{h,t}\sigma_{h,t}+\alpha_i\beta_{c,t}\sigma_{c,t}\right)+2C_{\text{Lip}}\epsilon_{\text{total}}
\end{align*}
where $C_{\text{Lip}}=1+C_{\text{align}}L_n\sum_{j=0}^{K-i-1}L_o^j$. To find the total cumulative regret $R_T$, we sum $\Delta_{t,i}$ over all $K$ subtasks and all $T$ rounds:
\begin{align*}
    R_T\leq 2C_{\text{Lip}}\sum_{t=1}^T\sum_{i=1}^K(\beta_{h,t}\sigma_{h,t}+\alpha_i\beta_{c,t}\sigma_{c,t})+O(TK\epsilon_{total})
\end{align*}
To bound the sum of the standard deviations $(\sigma_{h,t}$ and $\sigma_{c,t})$, we use the Cauchy-Schwarz inequality to obtain:
\begin{align*}
    \sum_{t=1}^T\sum_{i=1}^K\beta_{h,t}\sigma_{h,t}\leq \sqrt{TK}\sqrt{\sum_{t=1}^T\sum_{i=1}^K\beta_{h,t}^2\sigma_{h,t}^2}
\end{align*}
Let $N=\max\{N_1,N_2,...,N_k\}$, the maximum number of available arms for any subtask, then by Lemma 5.4 in \cite{neuralucbzhou20a},
\begin{align*}
    \sqrt{TK}\sqrt{\sum_{t=1}^T\sum_{i=1}^K\beta_{h,t}^2\sigma_{h,t}^2}\leq K\sqrt{T}d_{\text{eff}}\log(1+TN/\lambda)
\end{align*}
where $d_{\text{eff}}$ is the effective dimension of the neural feature space. Hence we have that 
\begin{align*}
    R_T= \tilde{O}(Kd_{\text{eff}}\sqrt{T})+O(TK\epsilon_{\text{total}})
\end{align*}
As the linearization error $\epsilon_{\text{total}}=O(m^{-1/6}+n^{-1/6})$ and under the overparameterized neural network setting $(m,n\rightarrow \infty)$ the residual error $O(TK\epsilon_{\text{total}})$ becomes a negligible constant and the total regret scales as $\tilde{O}(Kd_{\text{eff}}\sqrt{T})$ which is sublinear in $T$.
\end{proof}

\begin{figure}[t]
\centering
\includegraphics[width=\linewidth,height=5cm,keepaspectratio]{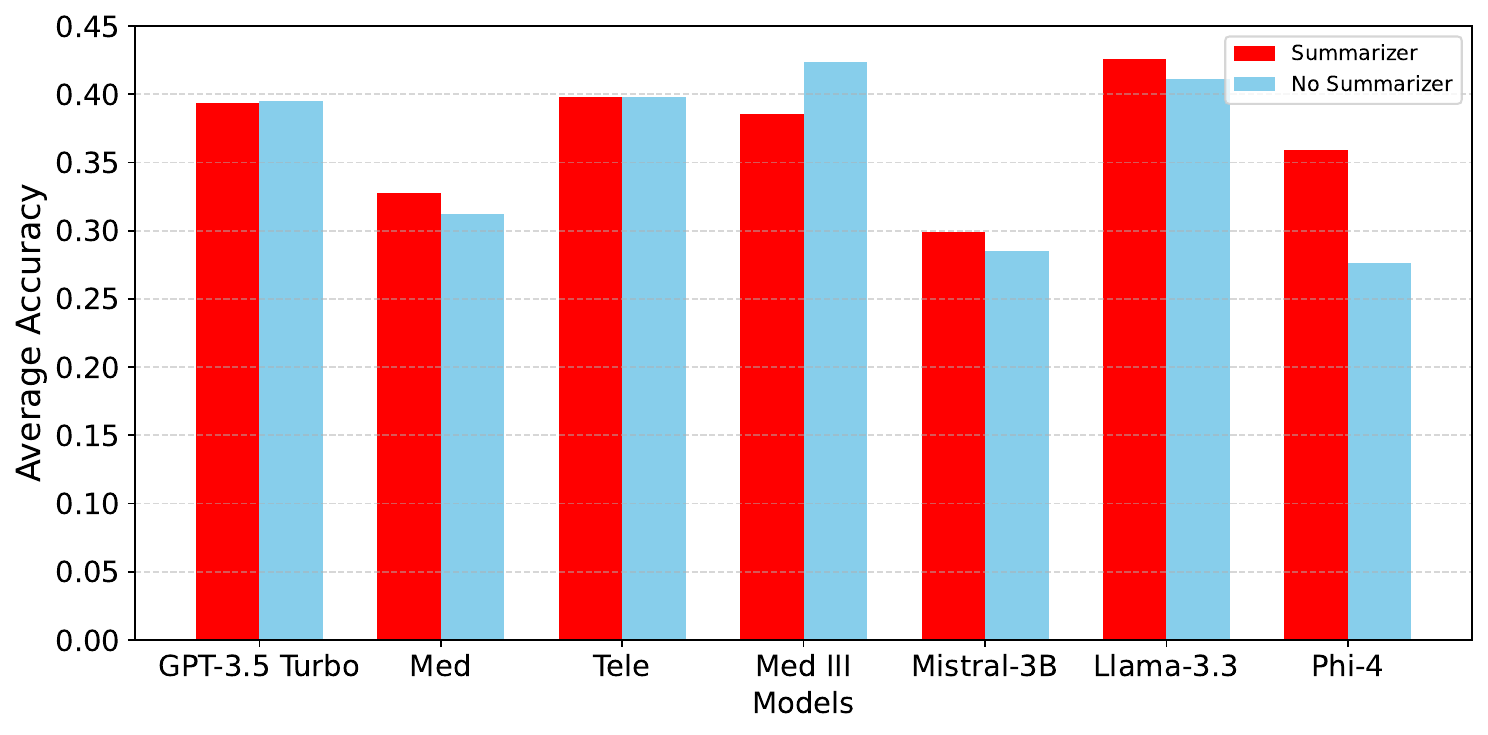}
\caption{Average accuracies of LLMs (Med: GPT 4o finetuned on medical data, Tele: GPT 4o finetuned on telecommunications questions, Med III: GPT 4o finetuned on MIMIC III dataset) on the medical diagnosis prediction task when the summarizer is chosen uniformly at random. Including a summarizer improves the average performance of most models, showing the value of breaking down a larger task into a pipeline of smaller ones. } 
\label{fig:avg_acc}
\end{figure}

\subsection{Experimental Details, Setup and Additional Results}
Now, we present some additional experimental results as well as experimental setup details. Figure \ref{fig:avg_acc} shows that using a summarizer improves the performance of the majority of the available LLMs, with only Med III model showing a significant dropoff in accuracy, which is due to the fact that this model is specifically finetuned on doing medical diagnoses on the unsummarized versions of the reports. Additionally, when we consider the results of Figure \ref{fig:net_utility_cost} together with Figure \ref{fig:avg_acc}, which displays the diagnosis accuracy for the different models in the medical setting, we would expect Llama to outperform Sequential Bandits, as Llama has the highest accuracy among the available models in this setting. However, the fact that our algorithm outperforms Llama shows that it can learn the complex dependencies between the subtasks, as there may be combinations other than Llama for both summarizer and diagnosis that obtain a higher net reward.

We now give more detailed explanation about the FrugalGPT baseline we implemented. We adapted FrugalGPT's cascade-construction module by replacing its internal response completion and  scoring with our own Azure-based inference wrapper and dataset specific reward function. We also implemented a round-based evaluator that repeatedly applies the static cascade to the query sequences, allowing us to compute the cumulative reward, cost, and regret curves even though FrugalGPT itself performs no online learning. Additionally, since difference cascade strategies are trained based on the allowed budget, for fairness we constrained the budget to be 3 times that our of diagnosis subtask cost.

Next, we detail how the Med, Med III and Tele models were finetuned on. The n2c2 smoking dataset \cite{uzuner2007evaluating} was used to fine-tune the Med model, with individual reports constructed into input/label pairs by extracting the diagnoses embedded within the reports. Given the format of the medical reports, they were initially processed through a summarizer to extract general medical knowledge before being utilized to fine-tune the GPT-4o models. The other two fine-tuned models, Med III and Tele, were respectively fine-tuned with the MIMIC-III and TeleQnA datasets.

\textbf{Scalability and Runtime.} 
A possible concern that might arise is whether learning separate neural networks for each subtask (or (subtask, LLM) pair in Sequential Bandits) would increase the runtime and required compute compared to other baselines. We want to note here that as Sequential Bandits is only training the corresponding network of the chosen model for each subtask, it's doing the same amount of training as the other neural bandit baselines. We have also compared the runtime of our algorithm with the Random baseline, which does not train any networks during training. Random achieved an average runtime of 5657 seconds per run on the Telecom dataset over 150 rounds whereas Sequential Bandits finished in 5666 seconds per run over 150 rounds averaged over 5 runs. This result shows that the main bottleneck of the runtime are the API calls rather than the online training as both achieve similar runtime.

\subsection{Cost Agnostic Setting Results}
First, we present our experimental results for the cost-agnostic versions of our algorithm for a non-pipelined single task telecommunications setting for the TeleQnA dataset \cite{maatouk2023teleqna} and for the cost-agnostic version of the 2 subtask medical pipeline we presented in the main paper. Here all the algorithms considered (NeuralUCB and NeuralLinUCB) are also cost-agnostic versions. As we are in the cost agnostic setting, $\boldsymbol{\alpha}=0$ hence net reward reduces to reward in this setting.

\subsubsection{Results for the Telecommunications setting:}
\begin{figure}[ht]
\centering
\includegraphics[width=\linewidth,keepaspectratio]{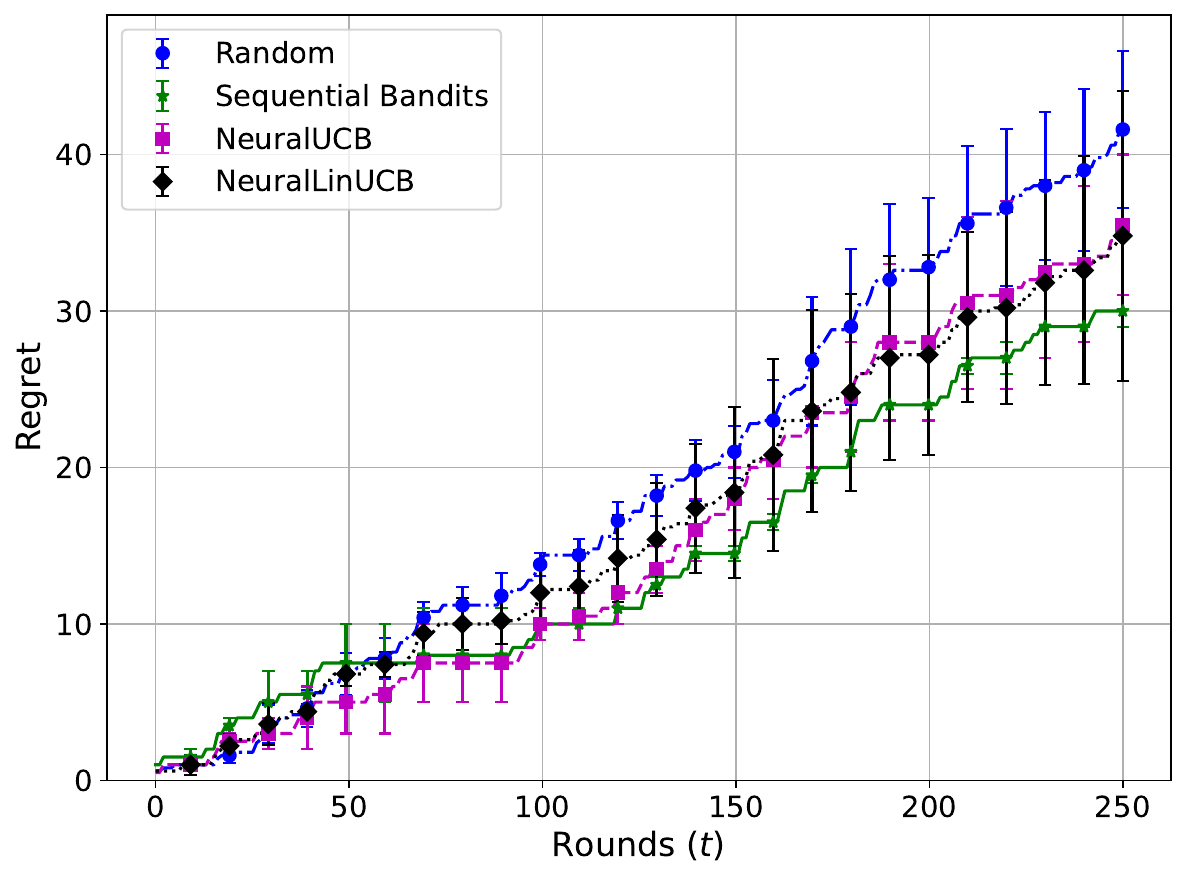}
\caption{Cumulative regret for TeleQnA in the single task cost agnostic setting} 
\label{fig:tele_reg_1}
\end{figure}

\begin{figure}[ht]
\centering
\includegraphics[width=\linewidth,keepaspectratio]{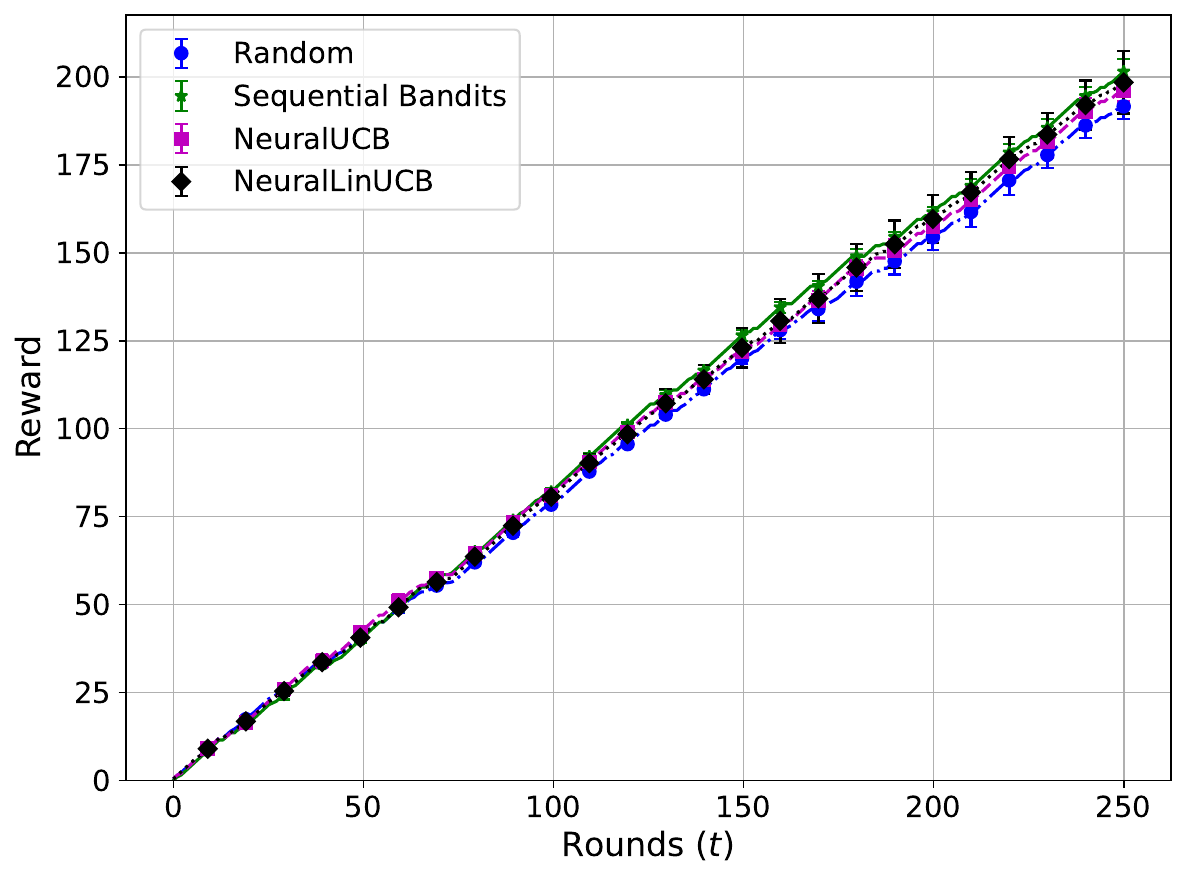}
\caption{Cumulative reward for TeleQnA in the single task cost agnostic setting} 
\label{fig:tele_rew_1}
\end{figure}

In this section, we present results on the non-pipelined multiple choice telecommunications question answering task. In this setting, a single LLM is chosen which is given as input the telecommunications question and outputs one of the available choices. Now, we show the reward and regret plots for this setting. 
Figures \ref{fig:tele_reg_1} and \ref{fig:tele_rew_1} show the cumulative regret and reward and shows that our algorithm outperforms the baselines. The error bars indicate the standard deviation which was obtained by averaging over 5 runs.

\begin{figure}[ht]
\centering
\includegraphics[width=\linewidth,height = 8cm,keepaspectratio]{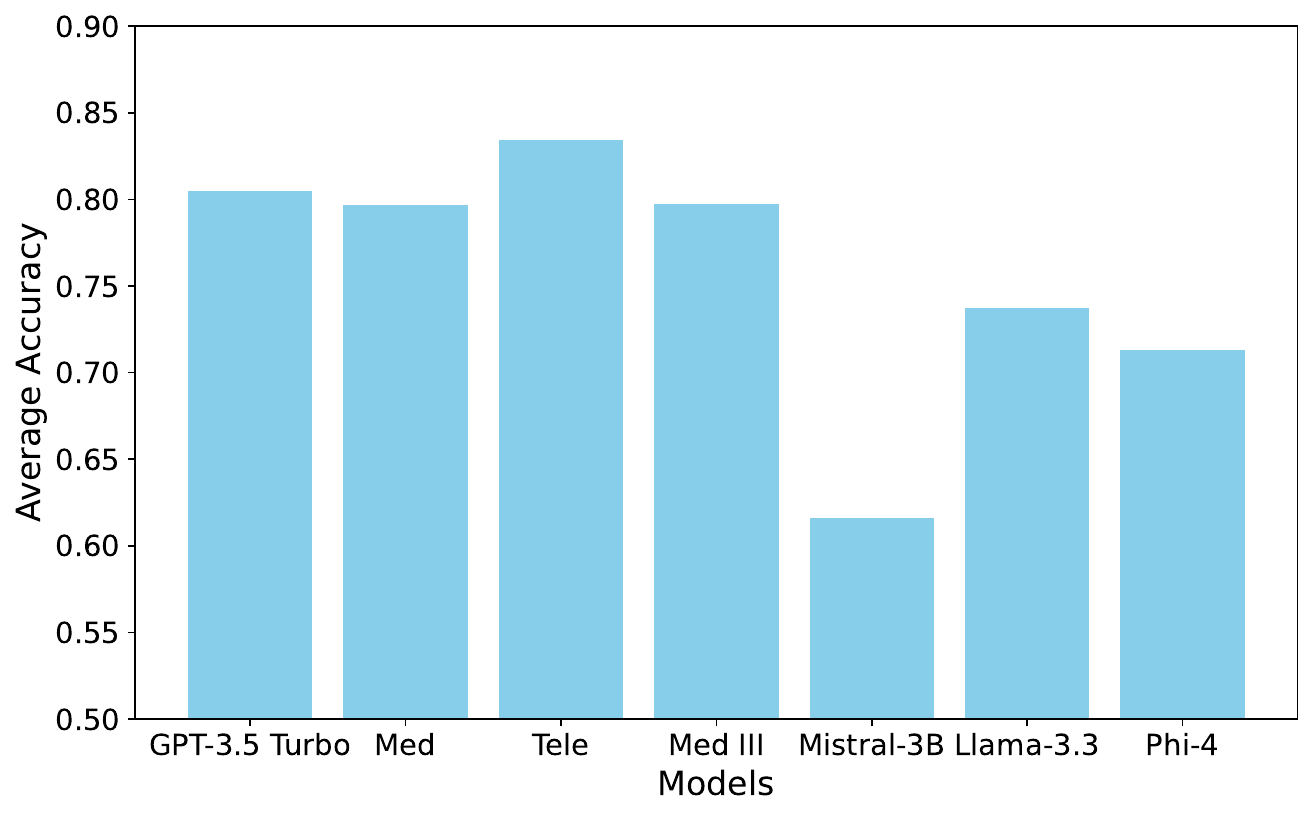}
\caption{Average accuracies of LLMs (Med: GPT 4o finetuned on medical data, Tele: GPT 4o finetuned on telecommunications questions, Med III: GPT 4o finetuned on MIMIC III dataset) of different LLMs on the multiple choice telecommunications questions answering task.} 
\label{fig:avg_tele}
\end{figure}

Next, we present the average accuracies of the LLMs for this task. The averages were taken over 15 runs. The accuracies in this setting is binary at every round, since in each round there is an incoming multiple choice telecommunications question which has only one correct choice. Figure \ref{fig:avg_tele} shows the accuracies for the different LLM models, indicating that the fine-tuned Telecom model from GPT-4o is the best performing one, followed by GPT-3.5 Turbo, Med III, and Med models. As in the diagnosis prediction task, Mistral-3B achieves the lowest accuracy for this task.



\begin{figure*}[ht]
    \centering
    \includegraphics[width=\linewidth]{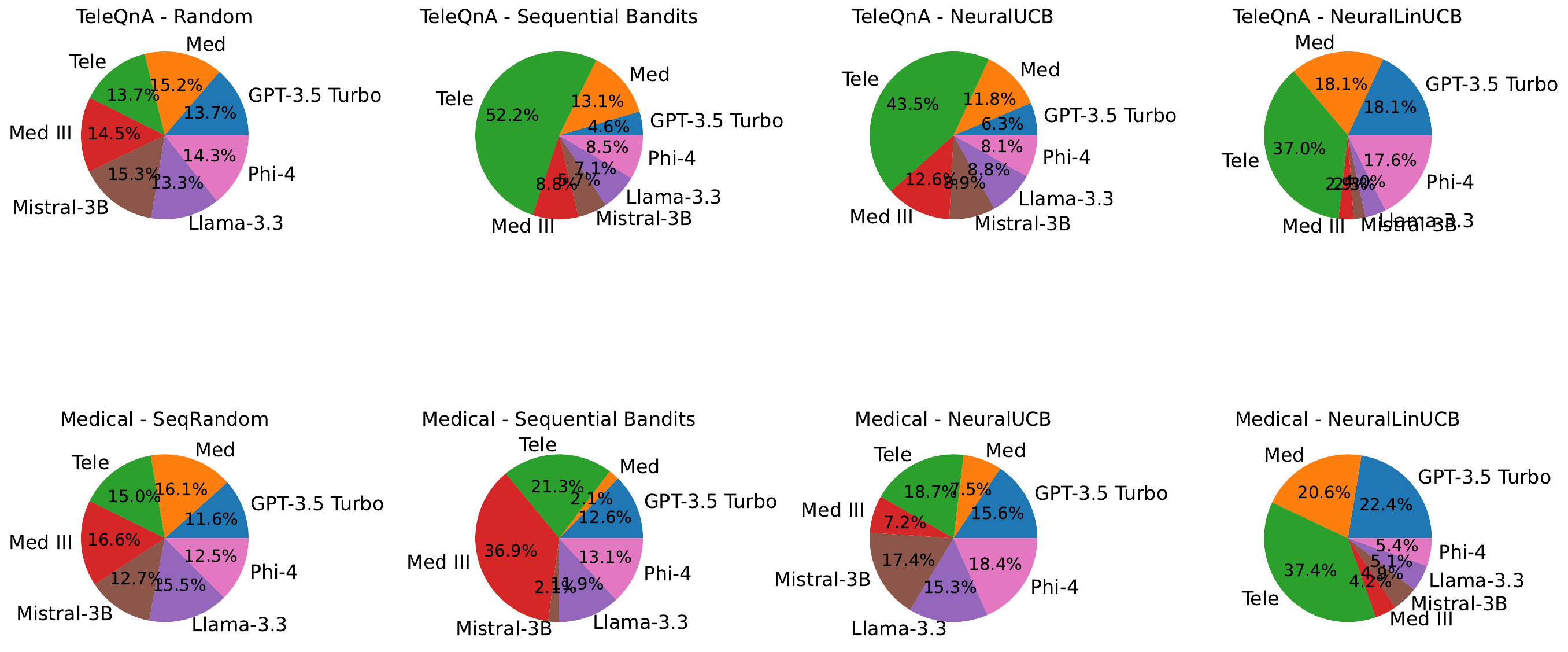}
    \caption{Model selection distribution for TeleQnA (1 task) and Medical (2 subtask) datasets (for diagnosis subtask) across different algorithms. Each pie chart shows the proportion of model usage.}
    \label{fig:pie_tele_med_all}
\end{figure*}

The pie charts that can be seen in Figure \ref{fig:pie_tele_med_all} show the percentage of model selections for the telecommunications task averaged over 5 runs. Random exhibits a relatively even selection of the models as expected while the remaining algorithms select the Telecom model the most. Our algorithm, Sequential Bandits plays the Telecom model the majority of the time (52.2\%) which is the highest among all algorithms. Compared to the application of diagnoses prediction from medical reports, NeuralUCB and NeuralLinUCB perform better, as indicated by their model selections as they are able to identify the models with higher accuracy and play them more.

\subsubsection{Results for the Medical setting:}
Now, we will present the results for the 2 subtask medical cost-agnostic setting. This is the same medical setting as we presented in the main paper where we have a summarizer LLM that summarizes the input medical report which is then passed to a diagnoser LLM that makes a diagnosis based on this summarized report. The main difference is that now we are in the cost-agnostic setting, meaning that we only look at the predicted accuracy of the models and not their cost. We consider two different types of random selection algorithms in these experiments. Random, selects an LLM at random to complete the diagnosis prediction task given the unsummarized medical report while SeqRandom operates under random selection for the 2 subtasks (equivalent to Random in the main paper).

\begin{figure}[ht]
\centering
\includegraphics[width=\linewidth,keepaspectratio]{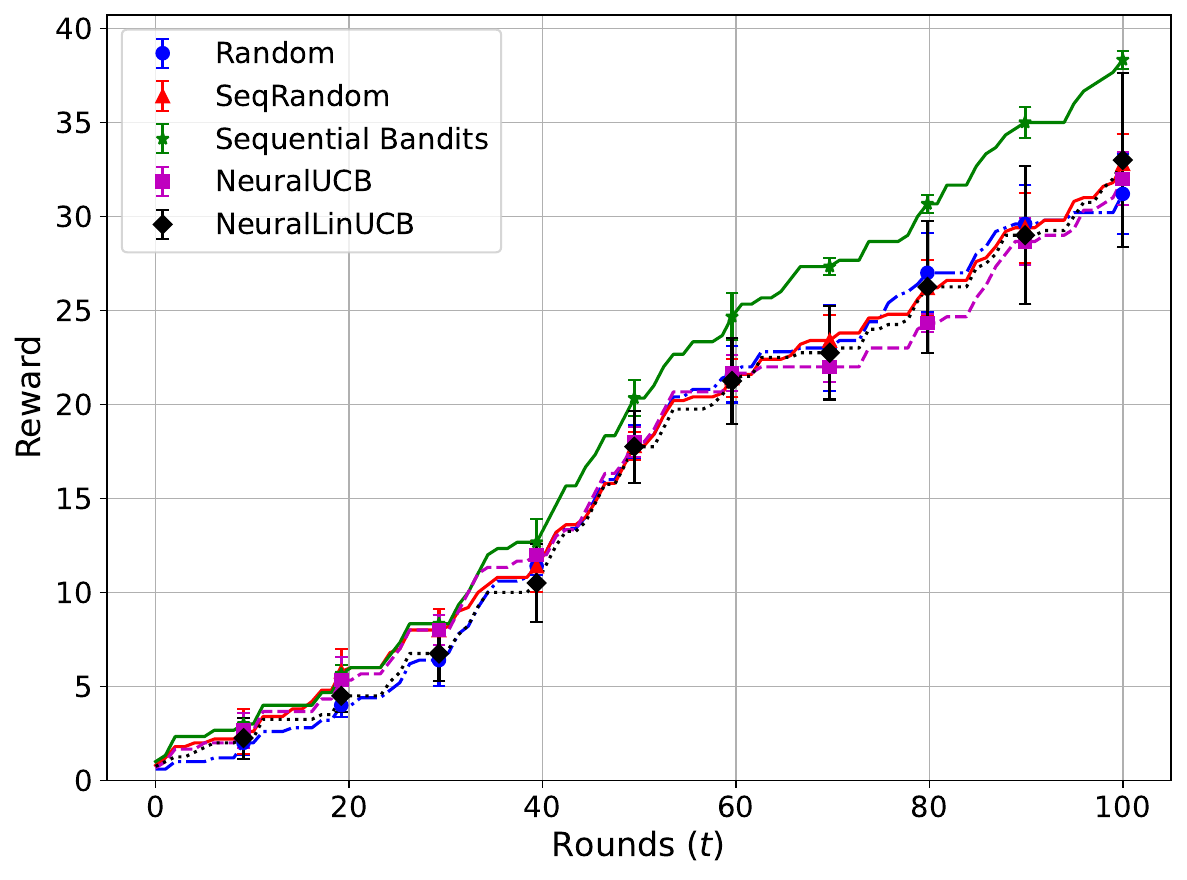}
\caption{Cumulative reward for 2 subtask medical cost agnostic setting} 
\label{fig:cost_agnostic_med_rew}
\end{figure}

\begin{figure}[ht]
\centering
\includegraphics[width=\linewidth,keepaspectratio]{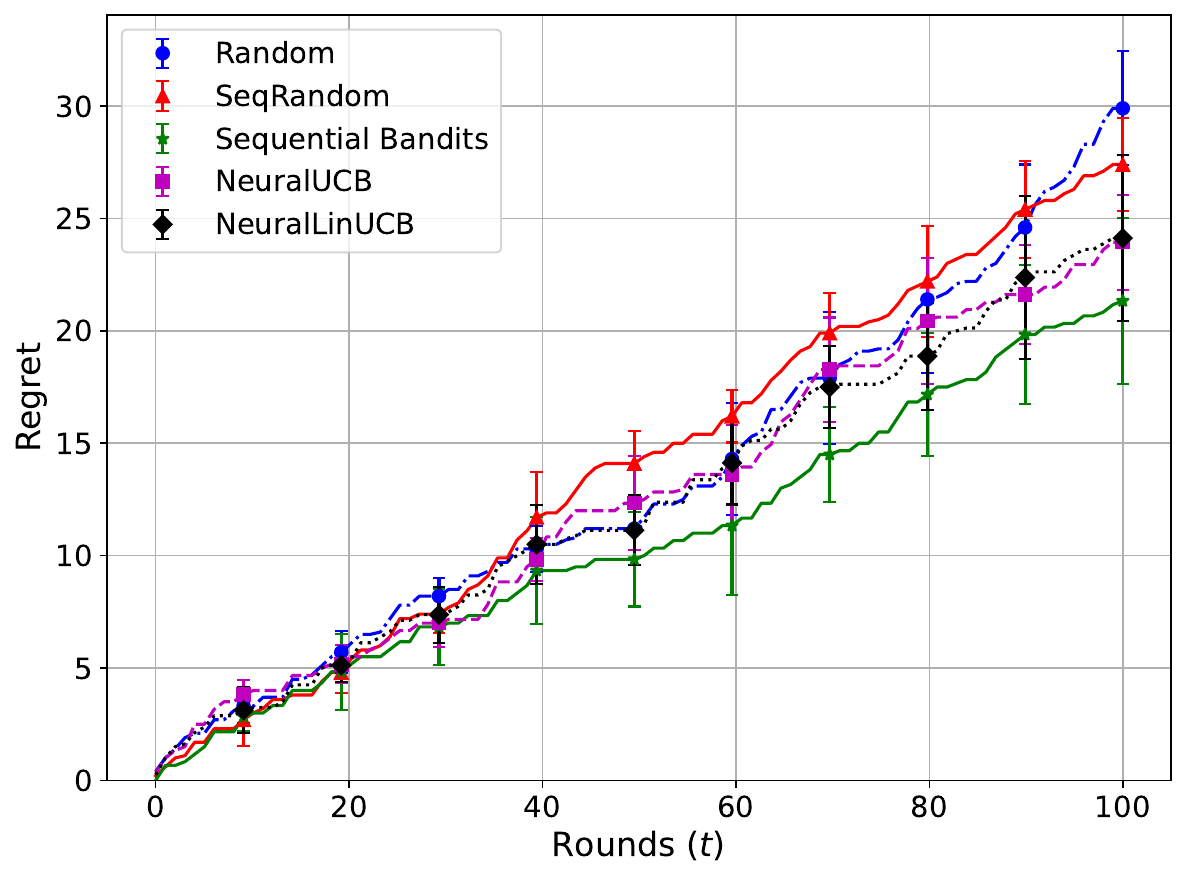}
\caption{Cumulative regret for 2 subtask medical cost agnostic setting} 
\label{fig:cost_agnostic_med_reg}
\end{figure}

As can be seen from Figures \ref{fig:cost_agnostic_med_rew} and \ref{fig:cost_agnostic_med_reg}, Sequential Bandits achieves the highest reward while also achieving the lowest regret among the baselines. The pie charts in Figure \ref{fig:pie_tele_med_all} show the percentage of models each algorithm selects for the diagnosis prediction subtask. As expected, Sequential Random shows a roughly uniform selection among the models. NeuralUCB, surprisingly, seems to exhibit a similar trend despite using a neural network to predict LLM success; however, this is because these pie charts show the average model selections among five individual experiments. In each experiment, NeuralUCB fixated on a different, usually suboptimal, LLM, causing the average to appear near-uniform across the LLMs. Our algorithm, Sequential Bandits, shows a preference toward the Med III and Telecom models, while selecting Llama-3.3, GPT-3.5 Turbo and Phi-4 relatively equally. Notably, our algorithm selects Mistral-3B and Medical very sparingly (both 2.1\%), which are the two worst performing models for this subtask, as indicated by the red bars in Figure \ref{fig:avg_acc}. NeuralLinUCB also selects Mistral-3B rarely (4.9\%);  however, it selects the Medical model quite often (20.6\%).

\subsection{Experimental Setup, Hyperparameter Tuning and Diagnosis Prediction Dataset}
\subsubsection{Experimental Setup and Hyperparameter Tuning:}
We now detail the hyperparameters we tuned for all the algorithms and how we formed the contexts for the different tasks from the input prompts and the descriptions of the LLMs. To get embeddings for the prompts and the descriptions we use HuggingFace's "Paraphrase-MiniLM-L6-v2" Sentence Transformer model which embeds into 384 dimensional vectors. We take the elementwise multiplication between the description of an LLM and its input prompt to form the context for the (LLM, subtask) combination. Elementwise multiplication is a commonly used similarity metric which is why we also utilize it in our experiments. This context is fed as input to the neural networks, which are 2-layer fully connected networks with a width of $50$ resulting in a total parameter count of $\approx19300$ for each of the networks. We set the regularization parameter $\lambda=1$ for all experiments and tune the learning rate $\eta$ between $10^{-2}$ and $10^{-4}$. We set the number of gradient descent steps as $J=5$. We also tune the ratio between the exploitation term (mu) and the exploration term (sigma) from 0.1-2 for the algorithms to balance exploration and exploitation as optimally as possible. Finally, we also tune the cost-sensitivity parameter $\boldsymbol{\alpha}$ for all our experiments. We tuned this parameter to ensure that the magnitude of the cost term and the accuracy term in the objective in Algorithm 1 is roughly equal to each other so that we give similar importance to cost and accuracy. The $\alpha$ values range from 50-150 in our experiments.

\begin{figure}[ht]
\centering
\includegraphics[width=\linewidth,keepaspectratio]{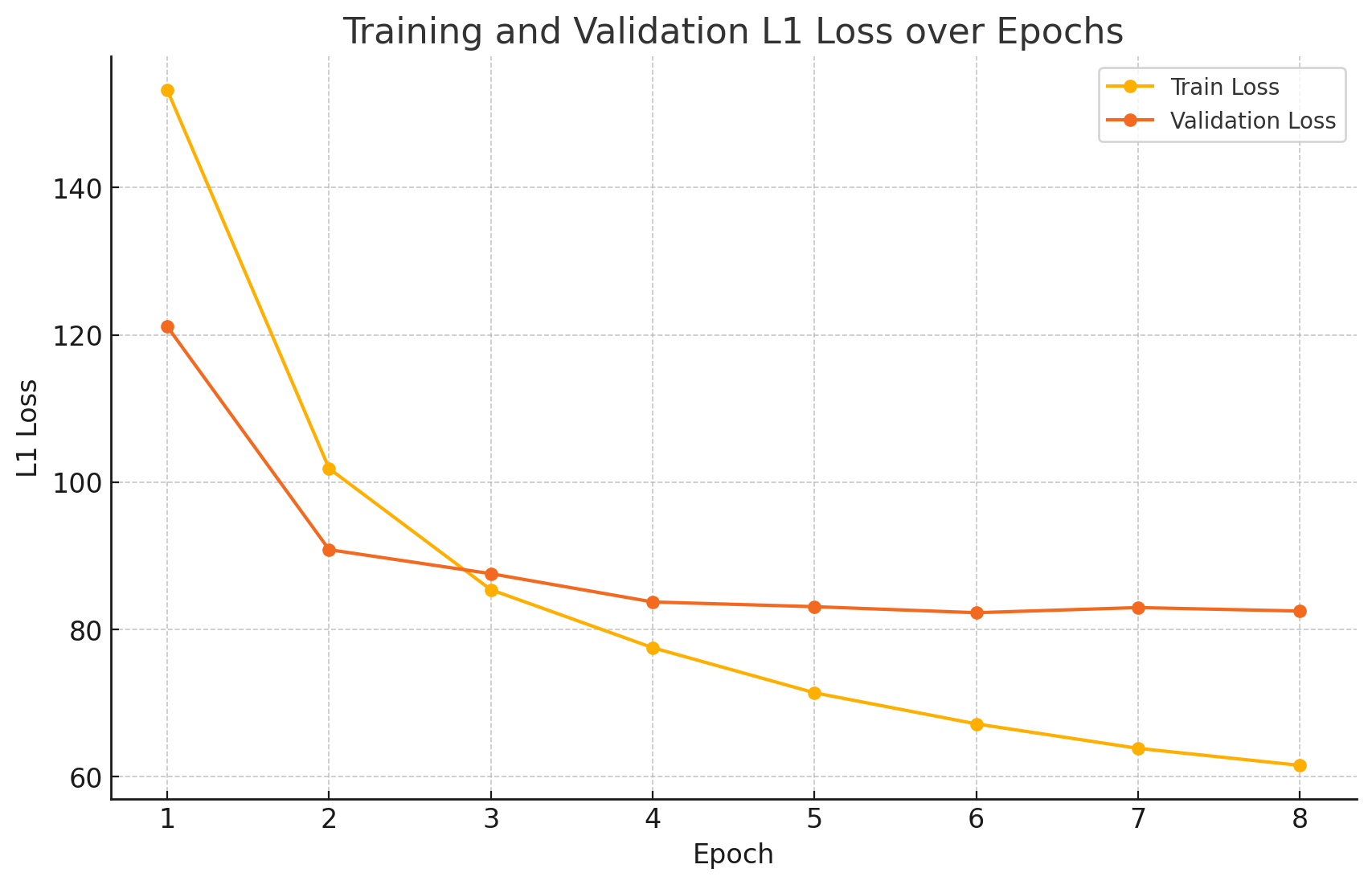}
\caption{Training and validation loss for offline training over epochs for Bert regression model for output token length prediction} 
\label{fig:bert_loss}
\end{figure}

\subsubsection{Details about our Diagnosis Prediction Dataset:}
We now give more details about the diagnostic prediction dataset that we created, listing the diagnoses of the 150 patients, and giving an example medical report. \cite{johnson2016mimic}'s MIMIC III dataset contains the admission diagnoses as well as diagnoses identified later, along with details of patient stay in their reports. Each patient has a number of reports based on their duration in the hospital, which we combine into a single report for each patient. We remove all explicit mentions of the patient's identified diagnoses, as well as diagnosis-related comments. We include the observations made and the test results of the patients in the reports. Patients we included in the dataset were mainly diagnosed with diseases related to the heart, kidneys, liver, and brain. Our dataset includes 150 medical reports with corresponding diagnoses. The diagnoses of the patients include the following: congestive heart failure, coronary artery disease, sepsis, acute coronary syndrome, v-tach, unstable angina, acute myocardial infarction, aortic stenosis, pulmonary embolism, pneumonia, opiate intoxication, carotid stenosis, complete heart block, respiratory failure, liver failure, encephalopathy, tamponade, neurosarcoidosis, aortic dissection, abdominal thoracic aneurysm, svc syndrome, urosepsis, hyponatremia, hypoxia, aortic insufficiency, cirrhosis of liver, vertebral/basilar stenosis, renal failure, chronic obstructive pulmonary disease, urinary tract infection, pulmonary edema. An example medical report for a patient is as follows:

FINAL REPORT TECHNIQUE: Portable semi-upright chest x-ray. There are no prior studies for comparison. There is an endotracheal tube with the tip approximately 6 cm above the carina.  There is cardiomegaly.  The pulmonary vessels are indistinct.  The mediastinum is unremarkable.  There are multiple bilateral pulmonary infiltrates slightly greater in the perihilar region.  There is no pneumothorax, there are no large effusions.  There are degenerative changes throughout the spine. IMPRESSION: CardiomegalyFINAL REPORT HISTORY: Line placement. A right IJ line has been inserted terminating in the upper SVC. There is no other change in the chest since the last chest xray at 11:27 PM on [**2138-8-2**]. Extensive bilateral fairly symmetric air space disease is noted. The heart is enlarged. An opaque catheter overlies the course of the proximal esophagus which may be an NGT. It cannot be seen distally below the level of the thoracic inlet. IMPRESSION: Right IJ line insertion.FINAL REPORT INDICATIONS:  SOB. CT SCAN OF THE CHEST USING A CT PA PROTOCOL: TECHNIQUE:   Helically acquired contiguous axial images of the chest with IV contrast using a CT PA protocol. CONTRAST:   100 cc Optiray IV due to fast bolus and allergies. FINDINGS:  There is no pulmonary embolism. There is bibasilar consolidation. There is a small left pleural effusion. There is no pericardial effusion. The tracheal bronchial tree is patent. There are no pathologically enlarged, hilar, mediastinal or axillary lymph nodes. There is diffuse regions of ground glass opacity involving both lungs. There are small regions of geographic sparring, predominantly in the periphery of the lungs. No pulmonary nodules or masses are identified.  Incidental note is made of a nasogastric tube and endotracheal tubes. IMPRESSION: 1. There is no pulmonary embolism. 2. Bibasilar dense consolidation. Diffuse ground glass opacity involving the lungs as described above.  The differential diagnosis would include edema and infectious/ inflammatory etiologies. 3. There is no pericardial effusions and there is no large pleural effusion.FINAL REPORT CHEST SINGLE AP FILM:  To reevaluate after diuresis. The endotracheal tube is 7 cm above carina.  Right jugular CV line is in right brachiocephalic vein.  Chest is rotated to the right.  Probable cardiomegaly. Since the prior study there has been persistent ill-defined hazy opacity at the lung bases.  No pneumothorax.FINAL REPORT AP SUPINE ABDOMINAL RADIOGRAPH INDICATION:  Tympanic abdomen on exam, respiratory failure of uncertain etiology. There are multiple dilated loops of small bowel overlying the abdomen which measure up to 5.7 cm.  There is a small amount of air in the recturm. There are no abnormal intra abdominal or intra pelvic calcifications.  The osseous structures are unremarkable. IMPRESSION:  Multiple dilated loops of small bowel are concerning for mechanical obstruction probably of the distal small bowel.FINAL REPORT HISTORY: Assess change in lines and tubes. Single view of the chest is compared to a previous study dated [**2138-8-5**]. The ETT is well above the carina. The right IJ line is in the right brachiocephalic vein. The cardiomediastinal contours are stable. There are persistent hazy opacities at both bases. There is no pneumothorax. IMPRESSION: Hazy illdefined opacities at both bases which may represent atelectasis although infection cannot be excluded.FINAL REPORT INDICATIONS:  Decreased breath sounds on the left. PORTABLE CHEST:  Comparison is made to previous films from  [**2138-8-8**]. The ET tube and right IJ line are unchanged in position. The tip of the NG tube is not identified.  The cardiomediastinal contours are stable. There is bilateral basilar hazy opacity. There is no pneumothorax. IMPRESSION:  Increasing bilateral pleural effusion and associated atelectasis.FINAL REPORT IMPRESSION:  Resolution of lower lobe opacities. FINDINGS:  Portable AP chest radiograph is reviewed and compared to previous study of [**2138-8-7**]. The previously identified opacities in both lower lobes have markedly improved.  These findings most likely represent improving aspiration pneumonia.   The heart is mildly enlarged.  The tip of the endotracheal tube is identified at the thoracic inlet.  A nasogastric tube courses towards the stomach.  No pneumothorax is seen.FINAL REPORT  Fifty year old female with recent episode requiring intubation.  Now post extubation.  Shortness of breath.  Assess for ischemia. SUMMARY OF EXERCISE DATA FROM THE REPORT OF THE EXERCISE LAB: Persantine was infused intravenously for approximately 4 minutes at a dose of approximately 0.142 mg/kg/min with no reported anginal symptoms or ST segment changes. INTERPRETATION:  One to three minutes after the cessation of infusion, MIBI was administered IV. Image Protocol: Gated SPECT. Resting perfusion images were obtained with thallium. Tracer was injected 15 minutes prior to obtaining the resting images. Stress and resting perfusion images show no evidence of myocardial perfusion defects.  Noted is a prominent right ventricle. Ejection fraction calculated from gated wall motion images obtained after Persantine administration shows a left ventricular ejection fraction of approximately 55\%.  There are no wall motion abnormalities. IMPRESSION:  No evidence of myocardial perfusion defects.  Prominent right ventricle.

We created the medical report above by combining multiple reports for the patient after different entries were made (after new observations, test results, imaging etc.) and removing any explicit mentions of the diagnoses. The diagnoses for the patient with the medical report above is pulmonary edema. Below we attach the summarized version of this report after being passed through an Assistant summarizer:

This extensive medical report describes multiple imaging studies performed on a fifty-year-old female patient with a history of recent intubation. The report includes findings from several chest X-rays, a CT scan of the chest, an abdominal radiograph, and results from a cardiac exercise lab.1. **Chest Imaging Findings:**   - Multiple portable chest X-rays reveal cardiomegaly (enlarged heart) and bilateral pulmonary infiltrates predominantly in the perihilar region. Throughout the studies, there is persistent hazy opacity at the lung bases, which may represent atelectasis or infection. Cardiomegaly and stable cardiomediastinal contours are consistently noted.   - The CT scan of the chest shows bibasilar consolidation, diffuse ground glass opacity potentially due to edema, and no pulmonary embolism. There is a small left pleural effusion without pericardial effusion.   - The most recent chest radiograph shows resolution of lower lobe opacities, suggesting improvement in previously suspected aspiration pneumonia.2. **Abdominal Imaging Findings:**   - The abdominal radiograph indicates multiple dilated loops of small bowel suggesting possible mechanical obstruction in the distal small bowel. There is no abnormal calcification.3. **Cardiac Exercise Lab Findings:**   - Perfusion images show no evidence of myocardial perfusion defects, and ejection fraction is calculated at approximately 55\%, with no wall motion abnormalities. Notably, there is a prominent right ventricle.4. **Line Placement and Tube Position:**   - There are placements of an endotracheal tube, right internal jugular line terminating in the upper SVC or right brachiocephalic vein, and an opaque catheter possibly representing an NG (nasogastric) tube whose tip isn't identified. The positions are stable in subsequent images.Overall, the patient presents with cardiomegaly, resolving pulmonary opacities, possible small bowel obstruction, and a prominent right ventricle noted on cardiac imaging. There is no evidence of myocardial perfusion defects or pulmonary embolism, and the pleural effusion is small.Provide a medical diagnosis based on the provided medical report summary above. This is for experimentation purposes only.

As seen above, the summarization of the medical report above highlights the most important information that may be helpful in making a diagnosis and hence simplifies the task for the next LLM in the pipeline.

\subsubsection{Output Token Length Prediction Model:}
Next, we give more details on the output token length prediction model that we use to predict the number of output tokens given the input prompt length. As we briefly explained in the main paper, we train a Bert regression model with L1 loss to train this model using the LMSYS-Chat 1M dataset \cite{zheng2023lmsys}.

\begin{figure}[ht]
\centering
\includegraphics[width=\linewidth,keepaspectratio]{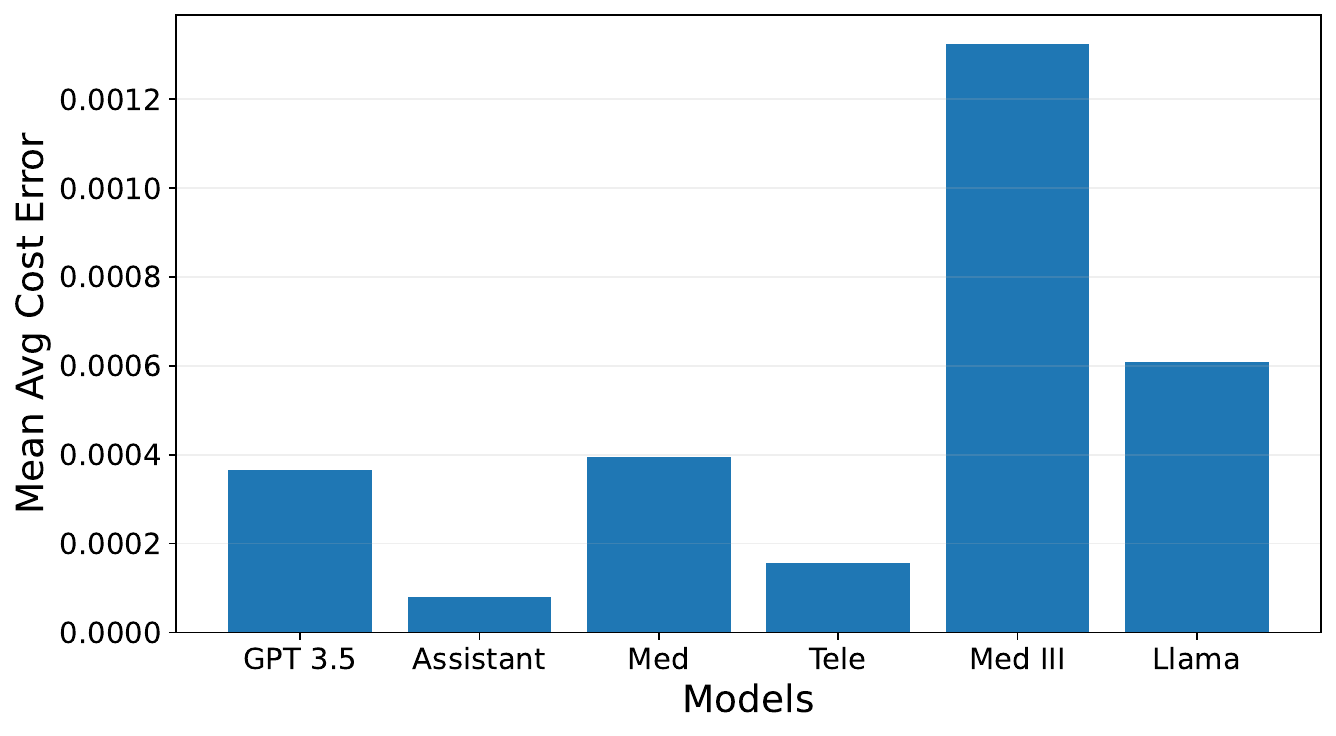}
\caption{Average cost estimation error for the different models with the offline token length prediction model} 
\label{fig:cost_err}
\end{figure}

\begin{figure}[ht]
\centering
\includegraphics[width=\linewidth,keepaspectratio]{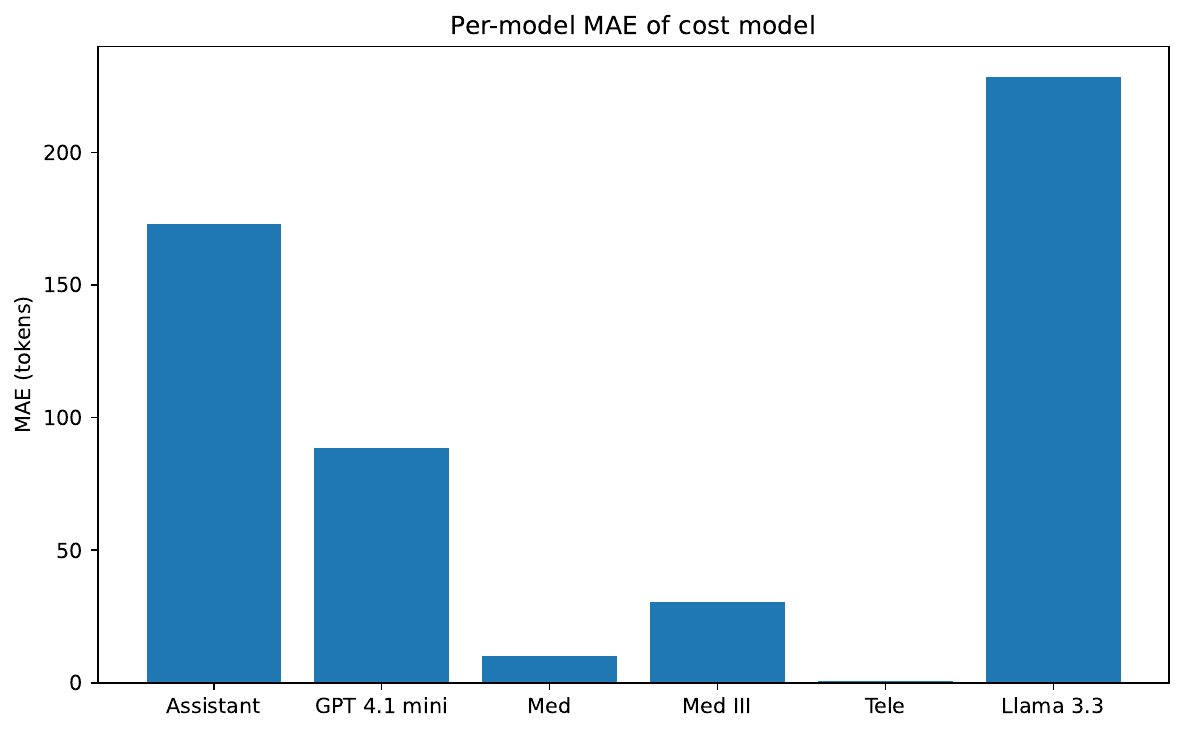}
\caption{Mean absolute token length estimation error for the different models with the online token length prediction model} 
\label{fig:mae_err}
\end{figure}

\begin{figure}[ht]
\centering
\includegraphics[width=\linewidth,keepaspectratio]{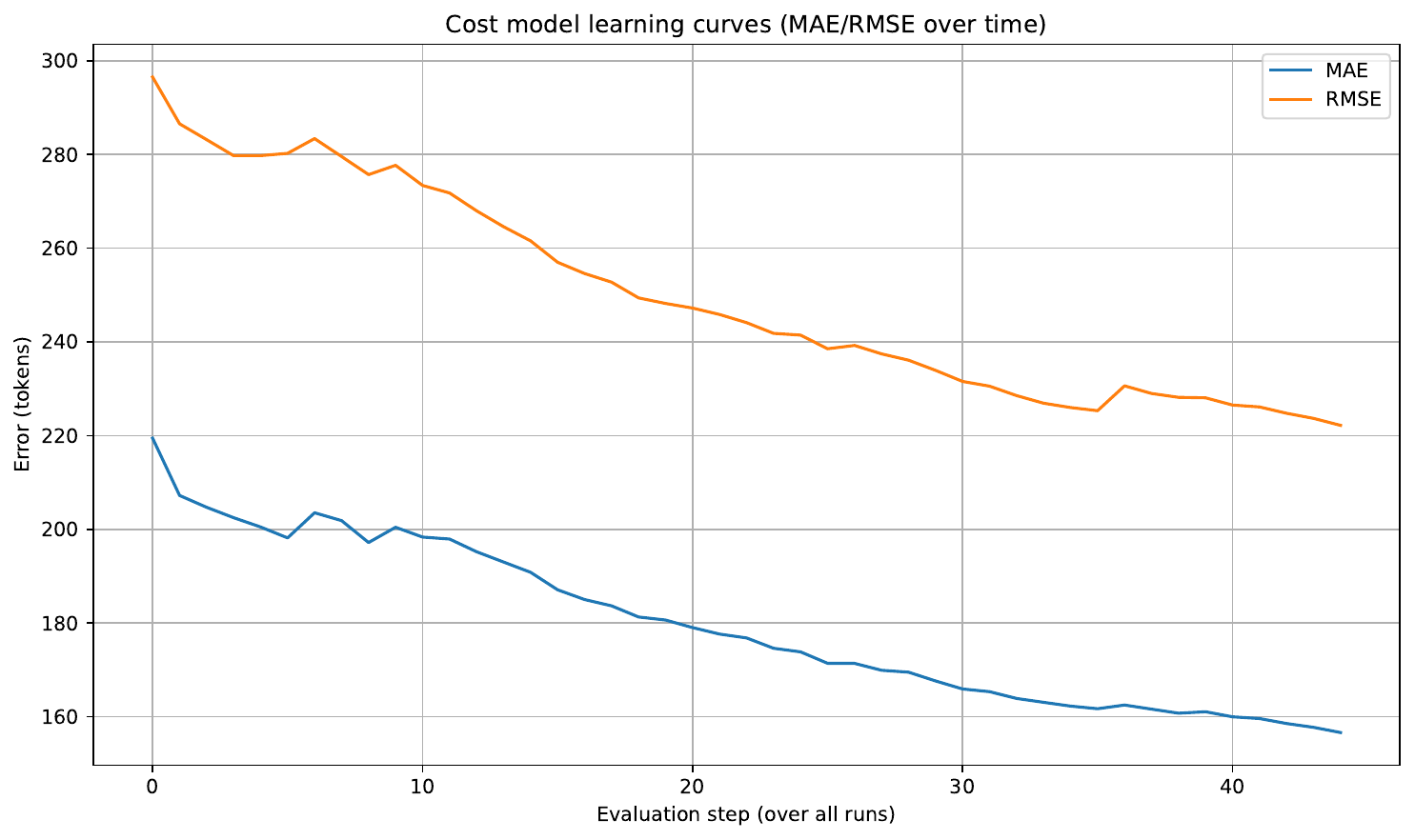}
\caption{MAE and RMSE variation with the evaluation steps (model updates) for the token length prediction model (online training)} 
\label{fig:mae_rmse_vs_t}
\end{figure}

We present the training and validation loss curves over the number of epochs for the Bert regression model with L1 loss using 50,000 examples with a 90\% to 10\% training and validation split in Figure \ref{fig:bert_loss}. The validation loss starts to increase after 6 epochs, hence we used the model trained up to 6 epochs for output token length prediction in our experiments as the model we use before making online updates to this model in our algorithm. This model is only trained once before starting the experiments for cost aware settings and is iteratively updated by taking an AdamW optimizer step every 5 token length observations with a learning rate of $10^{-6}$ during the training process shown in Algorithm 1. We will now provide the absolute average cost estimation error over all rounds for the different models that we use in our experiments. Figure \ref{fig:cost_err} shows the cost estimation error for the different models for the offline token length prediction model. It can be seen that Med III has considerably higher cost estimation error compared to the other models, which might be due to the fact that it was finetuned on a highly specialized dataset, making it difficult to predict. Figure \ref{fig:mae_err} shows the mean absolute error for the token length estimation for the online cost prediction model across all subtasks for the telecom experimental setting. Comparing Figure \ref{fig:cost_err} to Figure \ref{fig:mae_err} shows that the online updates especially benefits the finetuned models Med, Tele and Med III. This is expected, as finetuned models tend to give less verbose, predictable output lengths compared to other models like GPT or Llama. Llama seems to obtain relatively high estimation error according to both plots, a possible explanation for this is that the offline training for the cost model before starting the online training was done with the Llama 13B model as the LMSYS Chat 1M includes examples from this Llama model while the API calls we are making are for the Llama 70B model. This may explain why even after online training the error for Llama is higher compared to other models. Similarly for the Assistant, the experiments in Figure \ref{fig:cost_err} were done when the Assistant model was hosted on GPT 3.5-Turbo, but Azure was no longer hosting that model when we conducted the experiments in Figure \ref{fig:mae_err} and GPT 3.5-Turbo was replaced with GPT 4.1 mini. Also, LMSYS Chat 1M dataset had datapoints for GPT 3.5-Turbo which might have again caused a mismatch like in Llama during the offline training phase, contributing to higher error. Finally, we look at the variation of MAE and RMSE over time when we train the token length predictor online as shown in Figure \ref{fig:mae_rmse_vs_t}. Both MAE and RMSE show significant decreases in error, highlighting the benefit of online training. Every evaluation step corresponds to taking an optimizer step after observing 5 input token length-output token length pairs from the chosen models. In our experimental settings, we initialized the cost model at the beginning of the first run for each of the baselines with the offline trained model and train the cost model throughout all the runs so that the next run uses the improved model from the previous run. However, to ensure fairness across the different baselines, we did not use the online trained cost model across different baselines as this would create an advantage for the baselines that are run later on since their cost prediction model would be more accurate.

\subsubsection{Calculation of LLM Deployment Costs:}
To construct the total cost plots that can be seen in Figure \ref{fig:net_utility_cost} we use the per token pricing information that is available on Microsoft Azure, which is where we run all our experiments. The cost per input token for GPT 3.5 Turbo is \$0.0000005, for GPT 4.1-mini is \$0.0000001, \$0.00000025 for finetuned GPT 4o models (Med, Tele, Med III) and \$0.00000071 for Llama. The cost per output token for GPT 3.5 Turbo is \$0.0000015, for GPT 4.1-mini is \$0.000004, \$0.00001 for finetuned GPT 4o models (Med, Tele, Med III) and \$0.00000071 for Llama. Hence the finetuned models are the most expensive models, then Llama and the cheapest is GPT 3.5 Turbo. We use these per token costs for input and output and multiply them with the actual number of input tokens and the predicted number of output tokens to get the corresponding estimated cost values for using these LLMs. To calculate the actual costs, we use the actual number of output tokens instead, after observing the outputs of the LLMs.



\subsection{Cost Aware Setting Results}

In this section we present extra results for the 3 subtask telecommunication setting we detailed in the main paper such as the model selections with pie charts, accuracy of models for this setting as well as results for the 3 subtask version of the medical pipeline we considered before and extra results for the 2 subtask medical pipeline we detail in the main paper. We want to note that the cumulative regret results we present here are in terms of the regret for the reward rather than net reward, hence the presented regret plots do not take cost into consideration when computing regret.

\subsubsection{3 Subtask Medical Setting:}
We consider a task decomposition into three sequential subtasks: summarization, debate, and diagnosis. The rewards for the debate subtask is obtained by getting a pro-hallucination and a con-hallucination bias on the summary from before and getting the difference. The results presented here are averaged over 5 independent runs.

\begin{figure}[ht]
\centering
\includegraphics[width=\linewidth,keepaspectratio]{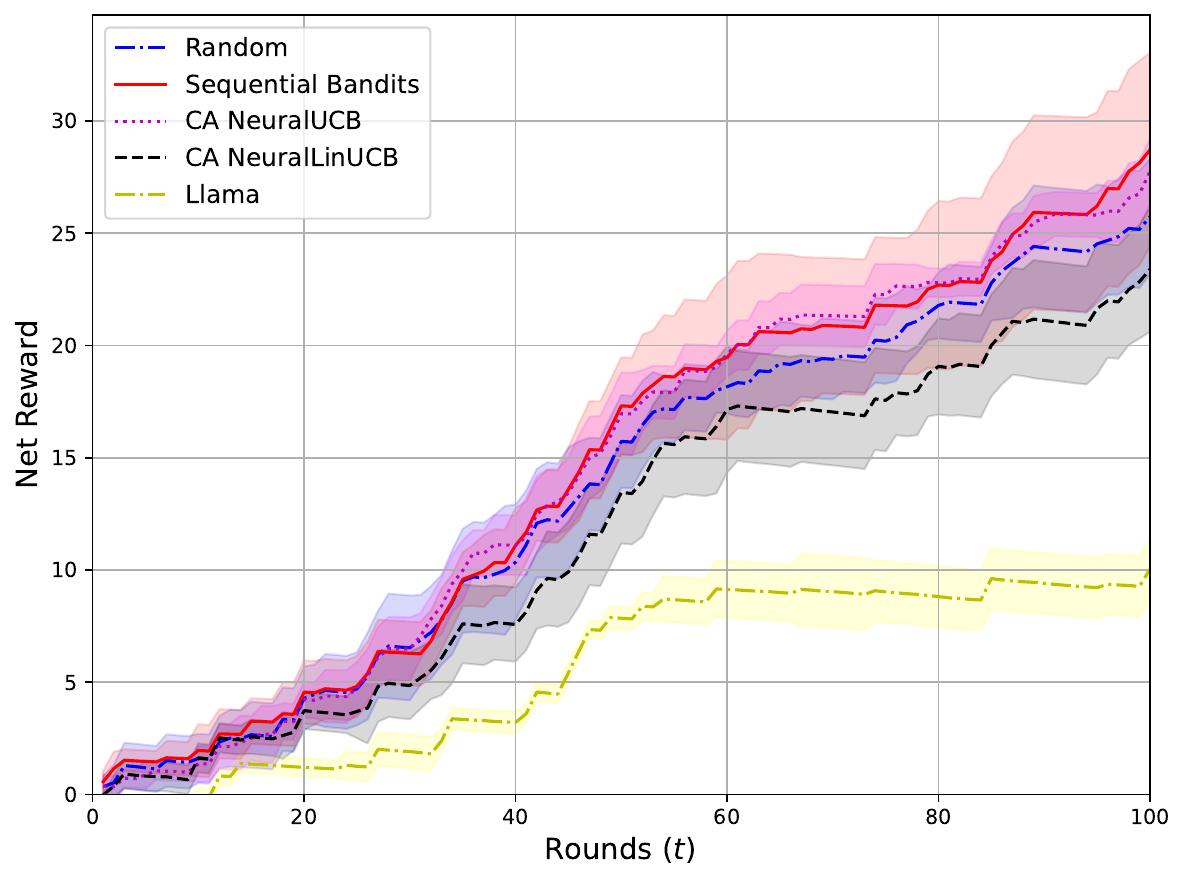}
\caption{Net reward for 3 subtask medical setting} 
\label{fig:net_3}
\end{figure}

As can be seen in Figure \ref{fig:net_3}, our algorithm indicated by the solid red line achieves the highest net reward in this cost aware setting. Cost Aware NeuralUCB also performs comparably to our algorithm, indicating its success. Random surprisingly outperforms both Cost Aware NeuralLinUCB and Llama and Llama performs considerably worse then all the other algorithms which indicates that static LLM selection strategies in these more complex pipelines might not be very reliable and effective.

\begin{figure}[ht]
\centering
\includegraphics[width=\linewidth,keepaspectratio]{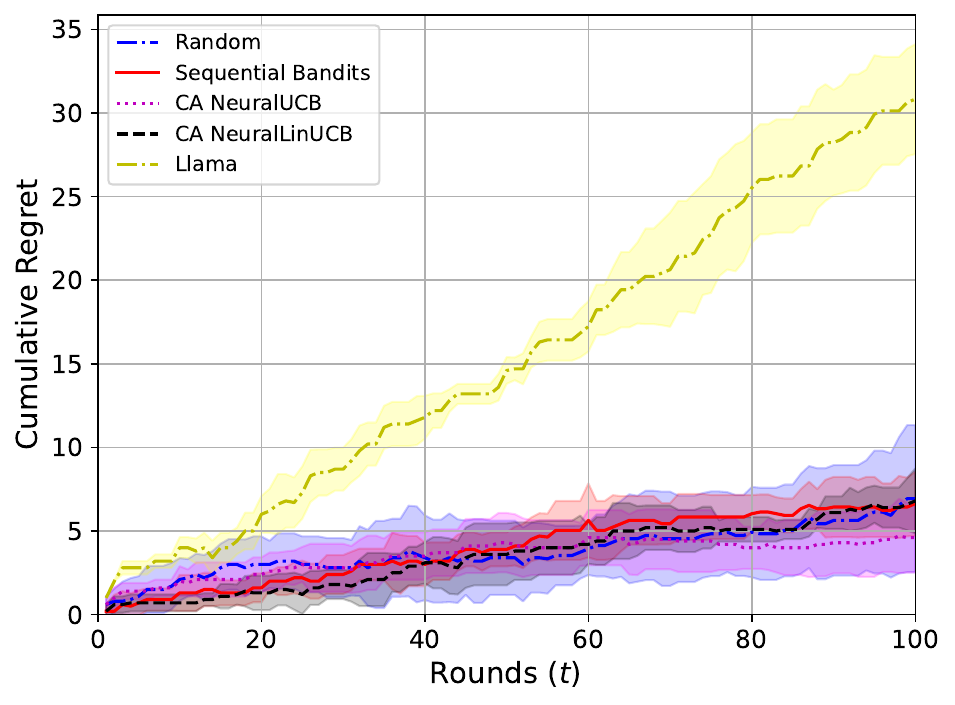}
\caption{Cumulative regret for 3 subtask medical setting} 
\label{fig:reg_3_med}
\end{figure}

Now, we consider the regrets of the algorithms as shown in Figure \ref{fig:reg_3_med}. It can be seen that the regret of the algorithms are mostly similar, except for Llama which performs considerably worse. There is some randomness involved with the regret as we measure it as the difference in the performance of the best model for that round compared to the performance of the selected model. This means that even if we are able to identify the best model for this subtask in hindsight, we may still incur regret as that model may perform poorly for a specific type of input whereas another gets an accuracy of 1, making us incur regret. Hence, this can be considered as a more competitive regret than is usually measured.

\begin{figure}[ht]
\centering
\includegraphics[width=\linewidth,keepaspectratio]{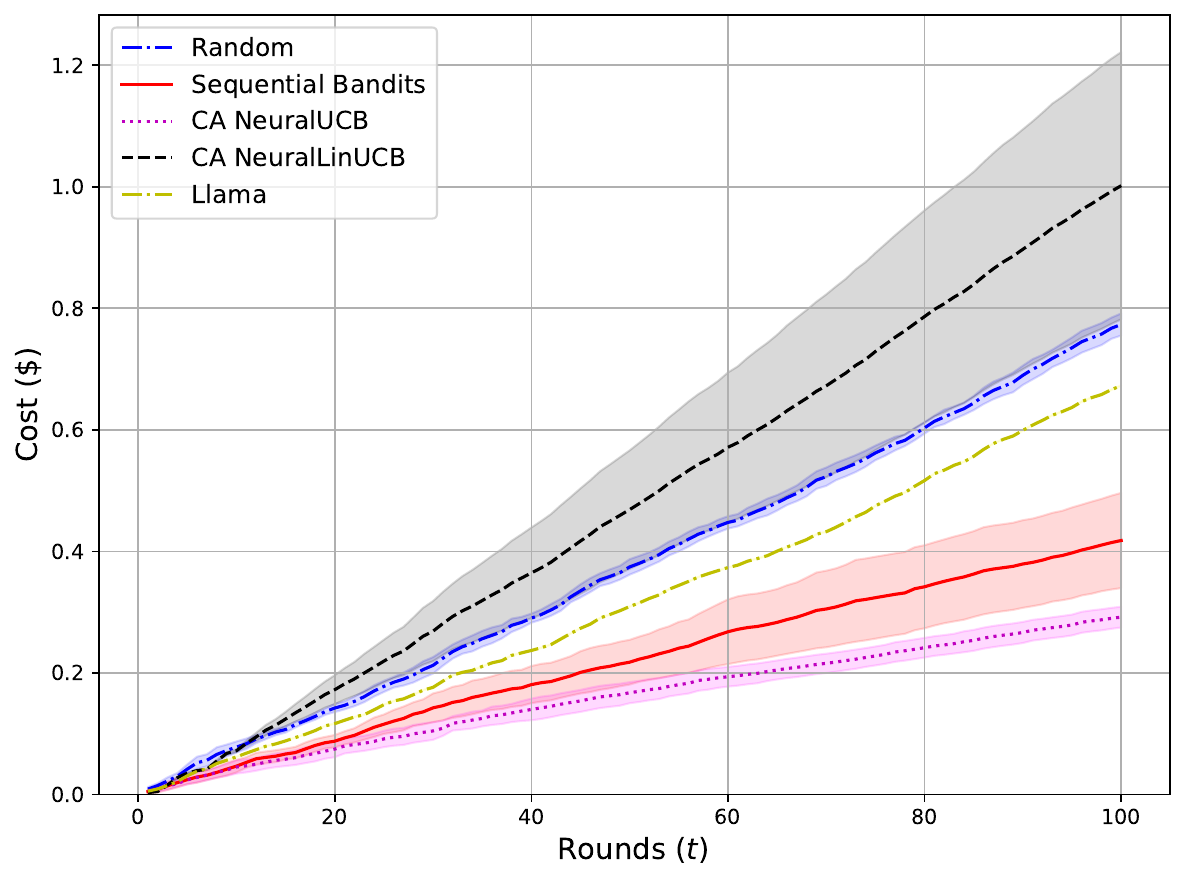}
\caption{Total cost for 3 subtask medical setting} 
\label{fig:cost_3}
\end{figure}

Next, we show the total cost incurred by the algorithms for the 3 subtask medical setting. As can be seen from Figure \ref{fig:cost_3}, Cost Aware NeuralLinUCB gets the highest cost, followed by Random, Llama, Sequential Bandits and Cost Aware NeuralLinUCB. It can be observed that the Llama baseline gets the lowest standard deviation when compared to the other algorithms for total cost as its shaded regions are barely visible. This is an advantage of the static LLM selection baselines as the total cost that they get show a low variation among different runs.

\begin{figure}[ht]
\centering
\includegraphics[width=\linewidth,keepaspectratio]{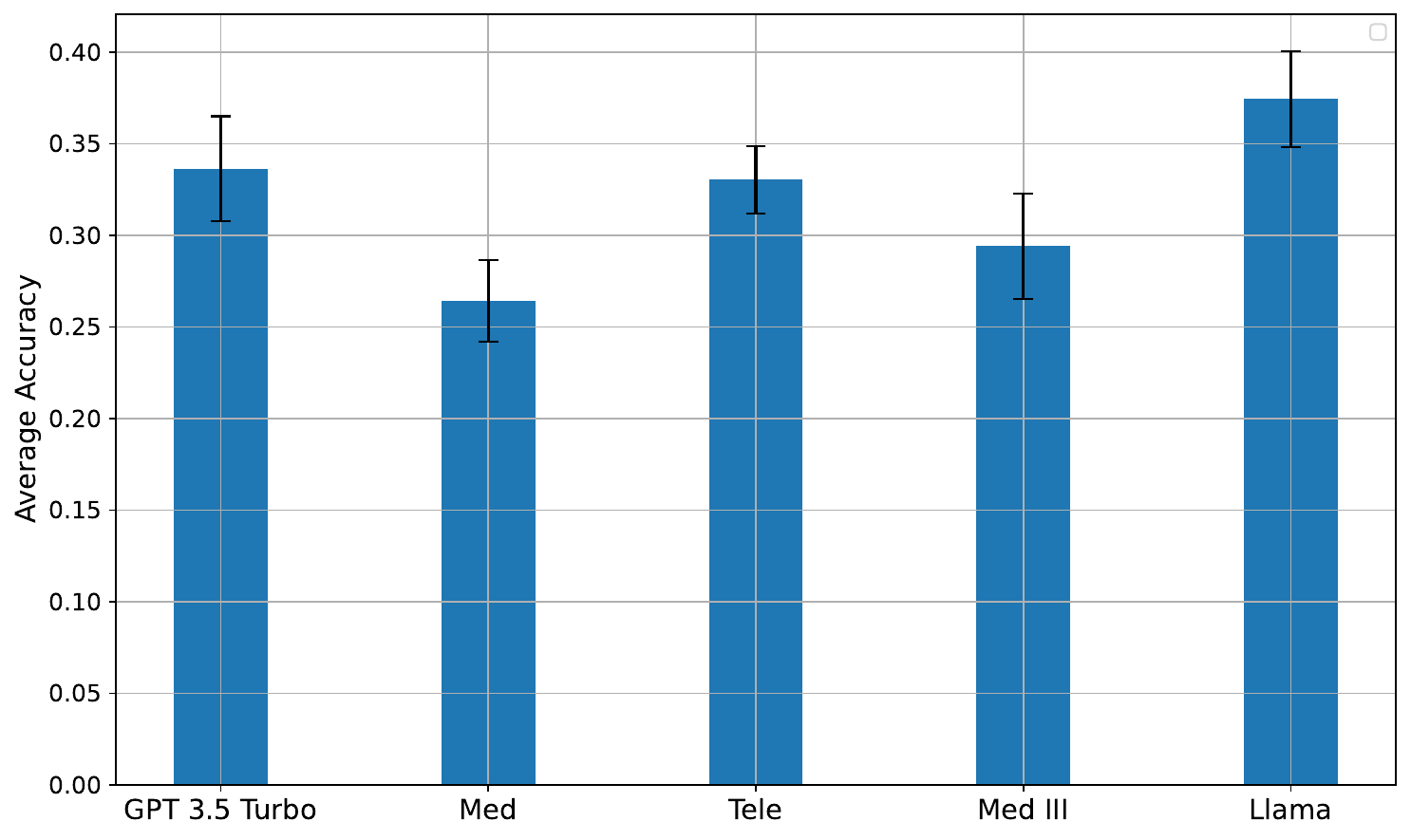}
\caption{Average accuracy of models for the diagnosis subtask in 3 subtask medical setting given the summarizer is selected at random} 
\label{fig:avg_3_med}
\end{figure}

Next, we show the average accuracy of the models for the diagnosis subtask in this 3 subtask pipeline setting as shown in Figure \ref{fig:avg_3_med}. As in Figure \ref{fig:avg_acc}, here accuracy is an indication of how many diagnoses the LLM was able to predict correctly. It can be seen that Llama obtains the highest mean accuracy followed by GPT 3.5 Turbo, Tele, Med III and Med which is a similar ordering to the one we had in Figure \ref{fig:avg_acc}.

\begin{figure*}[ht]
    \centering
    \includegraphics[width=\linewidth]{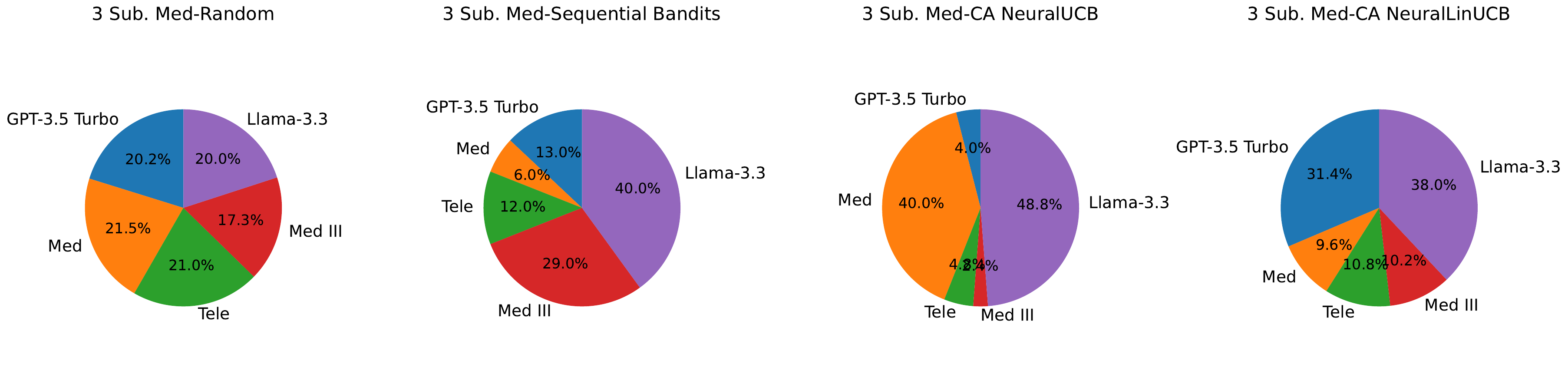}
    \caption{Model selection distribution for Medical (3 subtask) diagnosis subtask across different algorithms. Each pie chart shows the proportion of model usage.}
    \label{fig:pie_tele3_med3_all}
\end{figure*}

For the 3 subtask medical setting, the first row of Figure \ref{fig:pie_tele3_med3_all}, shows the model selections for the diagnosis subtask. Sequential Bandits and CA NeuralUCB are the best performers as is evident from their model choices, as they select Med less frequently compared to the others and select Llama more. 

\subsubsection{3 Subtask Telecom Setting:}
\begin{figure}[ht]
\centering
\includegraphics[width=\linewidth,keepaspectratio]{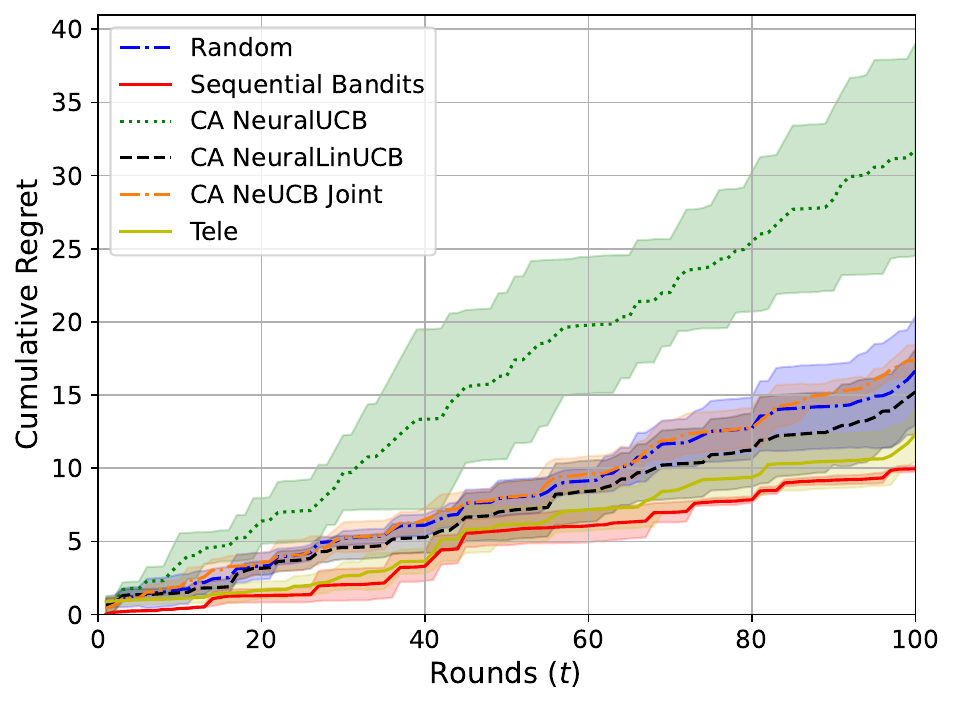}
\caption{Cumulative regret for 3 subtask telecom setting} 
\label{fig:tele_3_reg}
\end{figure}

In this section, we give more results for the cost aware 3 subtask telecom setting, which we introduced in the main paper. First, we present the cumulative regret for this setting as can be seen in Figure \ref{fig:tele_3_reg}. We observe a similar trend to the net reward in Figure \ref{fig:net_utility_cost} where Sequential Bandits outperforms the other baselines also in terms of regret and CA NeuralUCB performs the worst, while CA NeuralLinUCB and Random's performances are quite similar.

\begin{figure}[ht]
\centering
\includegraphics[width=\linewidth,keepaspectratio]{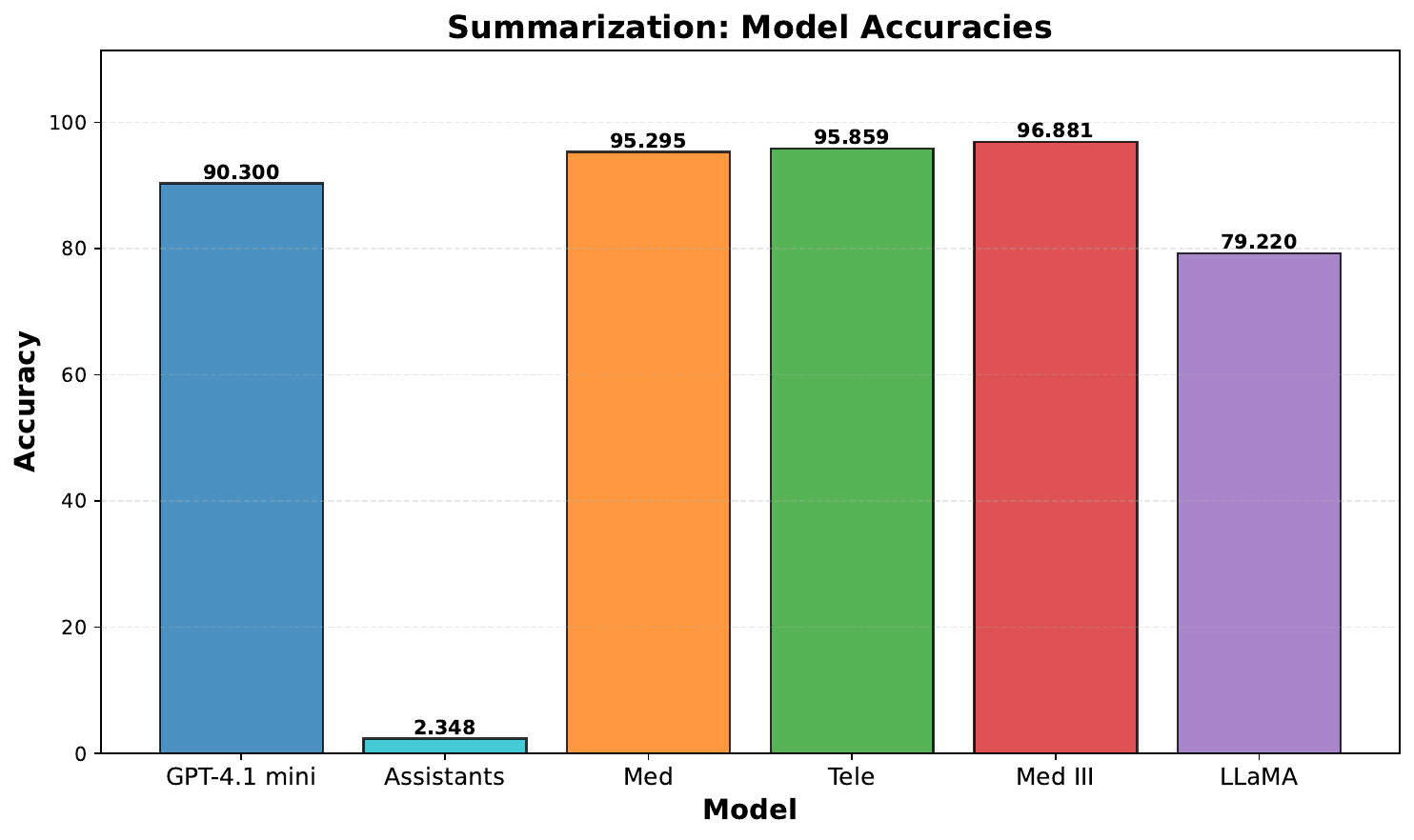}
\caption{Average accuracy of models for the summarization subtask given the summarizer is selected at random} 
\label{fig:tele_3_sum}
\end{figure}

\begin{figure}[ht]
\centering
\includegraphics[width=\linewidth,keepaspectratio]{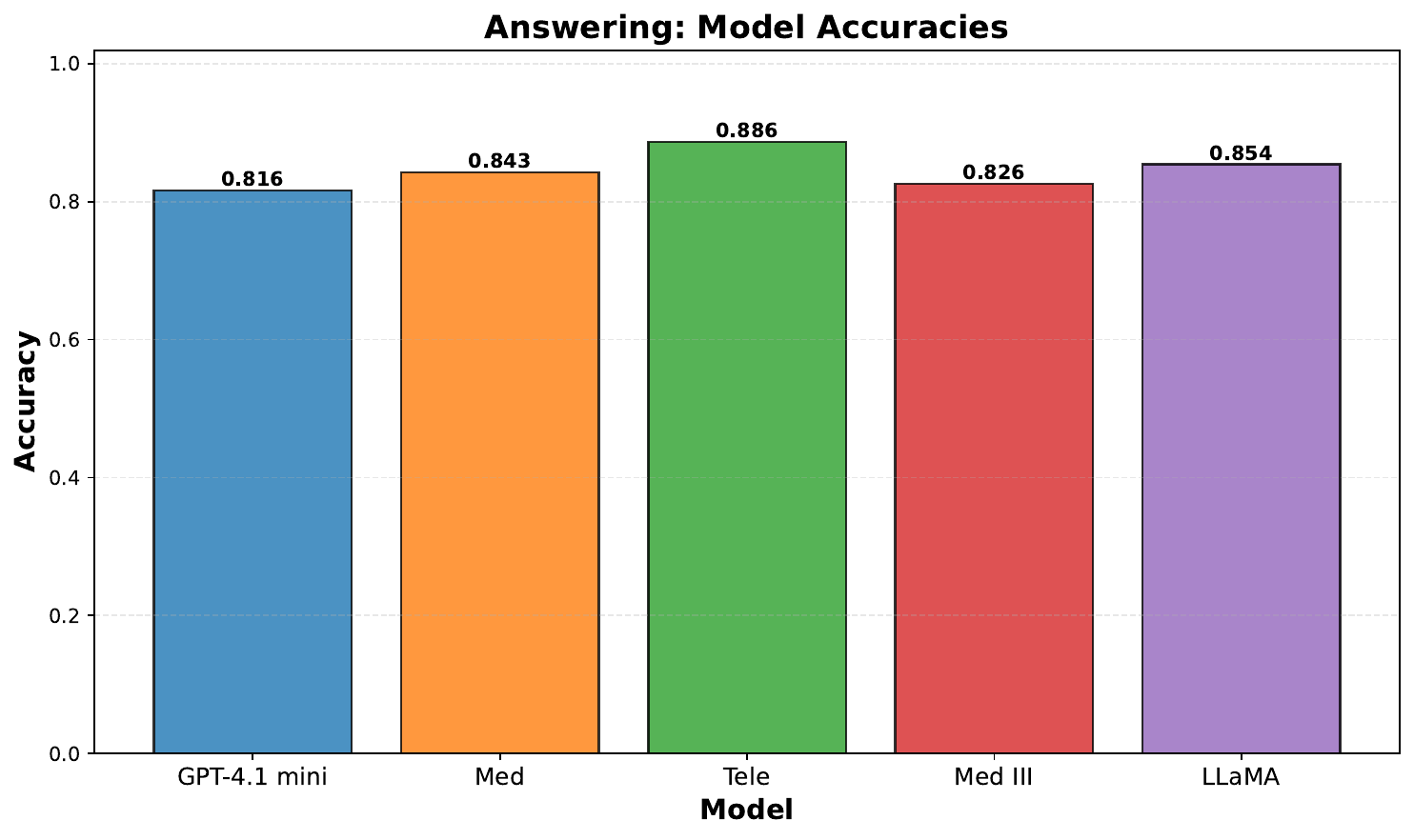}
\caption{Average accuracy of models for the answering subtask given the summarizer and answerer are selected at random} 
\label{fig:tele_3_ans}
\end{figure}

\begin{figure}[ht]
\centering
\includegraphics[width=\linewidth,keepaspectratio]{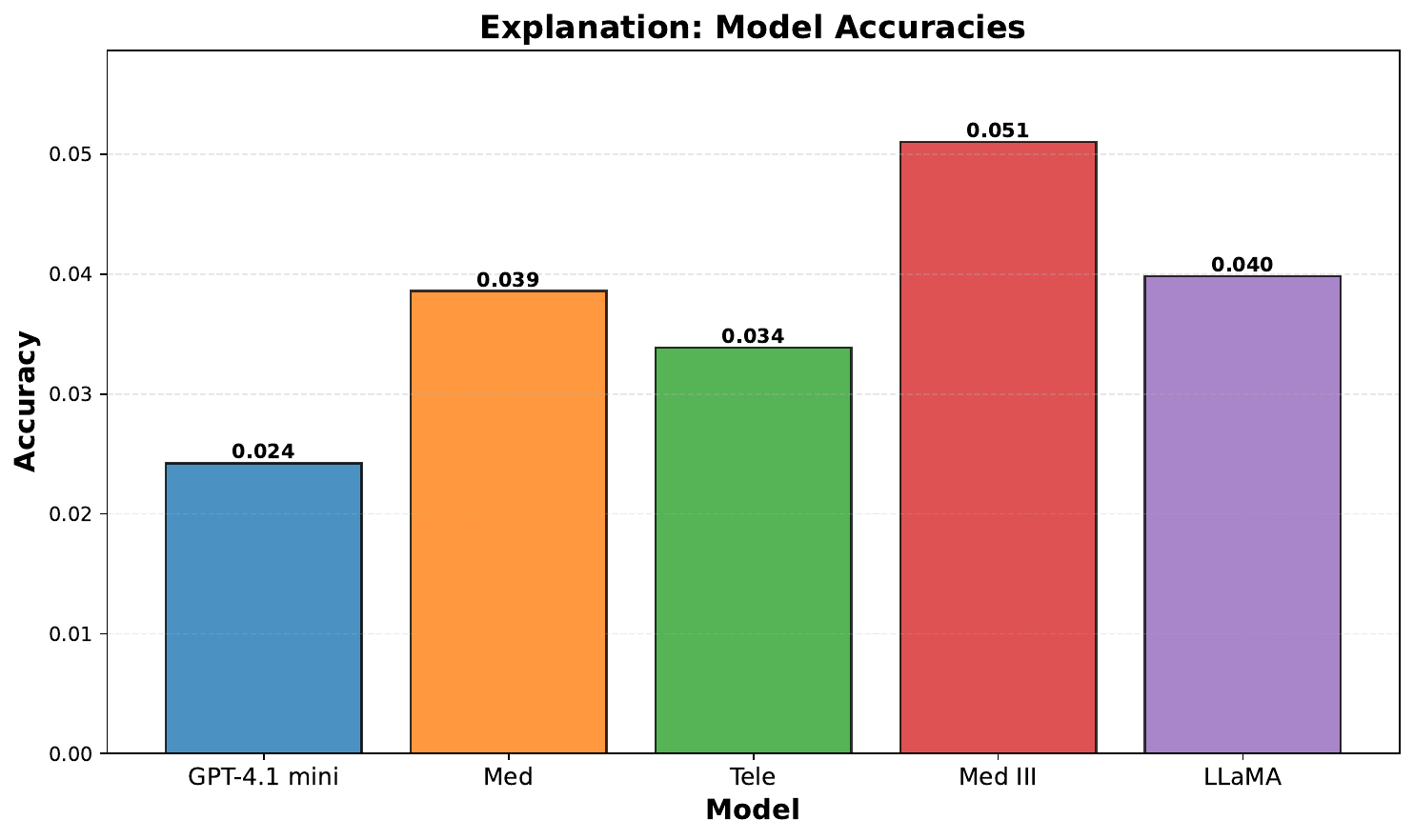}
\caption{Average accuracy of models for the explanation subtask given the summarizer, answerer and explainer are selected at random} 
\label{fig:tele_3_exp}
\end{figure}

Next, we look at the average accuracy of the models across all the subtasks. First we look at the summarization subtask, whose results can be seen in Figure \ref{fig:tele_3_sum} Med, Tele and Med III achieve competitive and similar summarization rewards at $\approx$ \%95, while Llama achieves around \%80 and Assistant does considerably worse at \%2. For the answering subtask, seen in Figure \ref{fig:tele_3_ans}, all models do relatively similarly, with Tele doing the best among them, as expected. For the explanation subtask, seen in Figure \ref{fig:tele_3_exp}, the Med III model performs the best followed by Llama and Med models.

\begin{figure*}[ht]
    \centering
    \includegraphics[width=\linewidth]{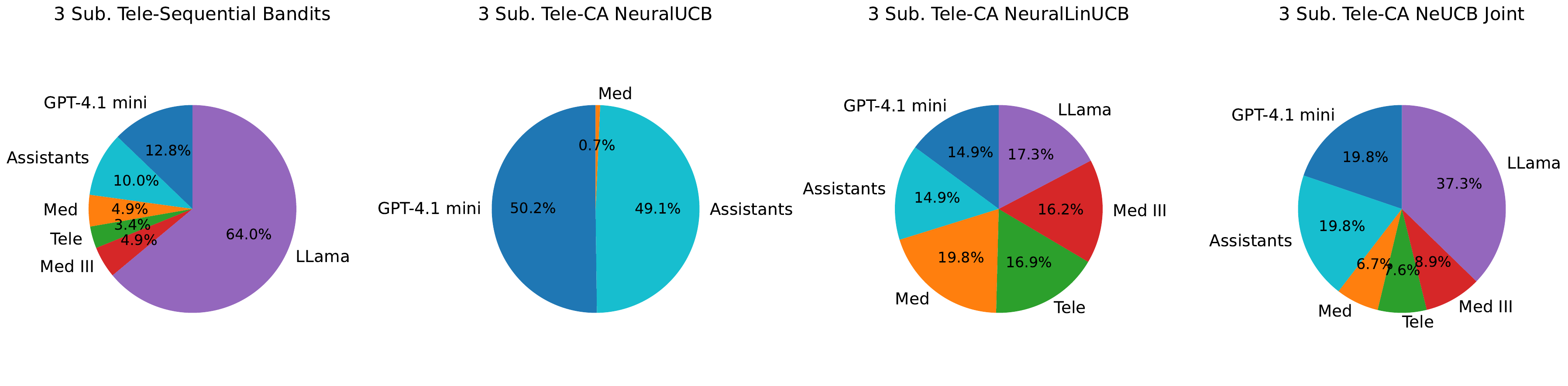}
    \caption{Model selection distribution for Telecom (3 subtask) summarization subtask across different algorithms. Each pie chart shows the proportion of model usage.}
    \label{fig:pie_tele3_sum}
\end{figure*}

\begin{figure*}[ht]
    \centering
    \includegraphics[width=\linewidth]{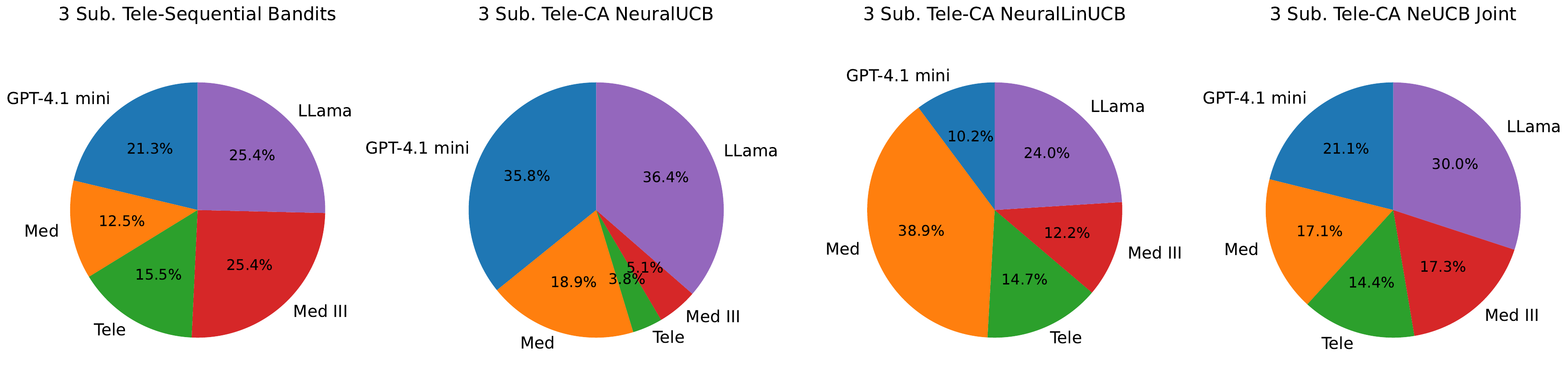}
    \caption{Model selection distribution for Telecom (3 subtask) answering subtask across different algorithms. Each pie chart shows the proportion of model usage.}
    \label{fig:pie_tele3_ans}
\end{figure*}

\begin{figure*}[ht]
    \centering
    \includegraphics[width=\linewidth]{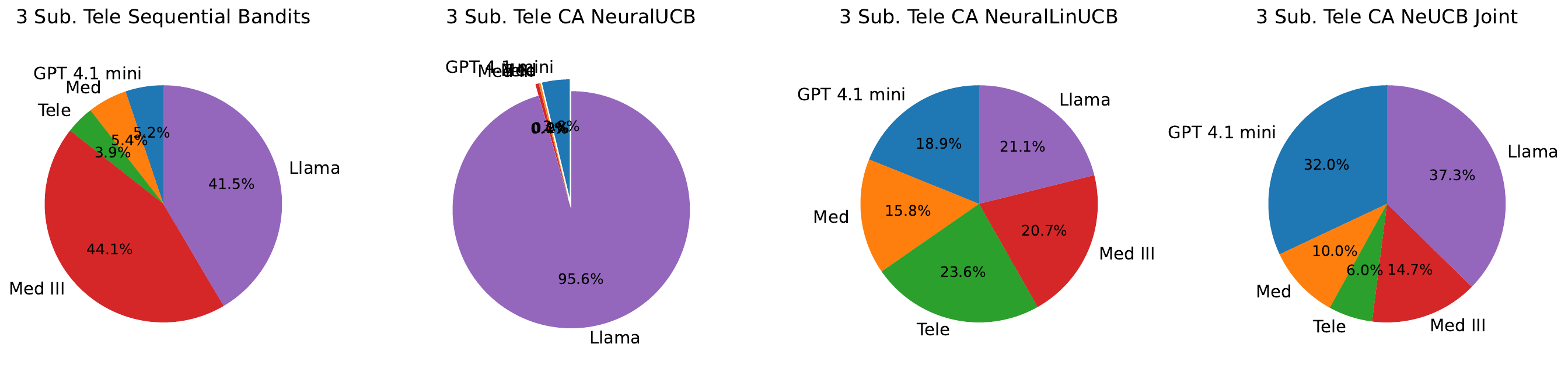}
    \caption{Model selection distribution for Telecom (3 subtask) explanation subtask across different algorithms. Each pie chart shows the proportion of model usage.}
    \label{fig:pie_tele3_exp}
\end{figure*}

Now, we analyze the model selections each of the algorithms made for summarization, answering and explanation subtasks. Figure \ref{fig:pie_tele3_sum}, \ref{fig:pie_tele3_ans} show the model selections for the summarization and answering subtasks. For Sequential Bandits we see that Llama is selected the most often followed by GPT-4.1 mini while for CA NeuralUCB it selects nearly evenly between Assistants and GPT-4.1 mini, showing its suboptimality as Assistants gets the lowest summarization accuracy by far. CA NeuralLinUCB similarly shows near uniform selection as it did for the explanation subtask while CA NeUCB Joint seems to prefer Llama, GPT 4.1 mini and Assistants. For the answering subtasks, Sequential Bandits exhibits a near uniform selection, which makes sense given that the gap in accuracy between the models is quite small for this subtask. Similarly, CA NeUCB Joint also shows near uniform selection while CA NeuralUCB and CA NeuralLinUCB prefer GPT-4.1 mini, Llama and Med, Llama the most respectively. For the telecom explanation subtask seen in Figure \ref{fig:pie_tele3_exp} (i)-(l), Sequential Bandits selects Med III and Llama the most, which achieve the highest and second highest accuracy for this task respectively and selects the remaining models relatively equally. CA NeuralUCB selects Llama the most while CA NeuralLinUCB selects all models nearly equally. CA NeUCB Joint on the other hand, shows a preference towards Llama and GPT 4.1 mini.

\subsubsection{2 Subtask Medical Setting:}
In this section, we present some additional results for the 2 subtask medical setting which we talked in detail in the main paper.

\begin{figure}[ht]
\centering
\includegraphics[width=\linewidth,keepaspectratio]{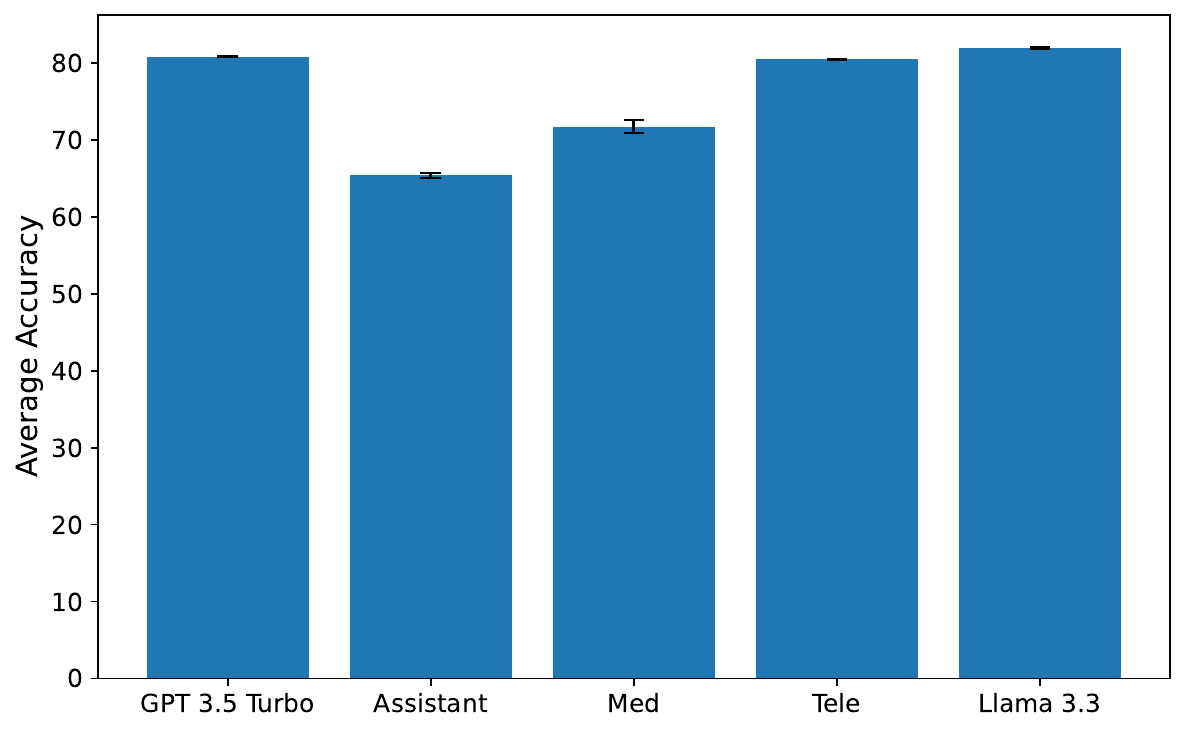}
\caption{Average accuracy of models for the summarization subtask given the summarizer is selected at random} 
\label{fig:sum_acc_med2}
\end{figure}

First, we present a result on the average accuracy of the models for the summarization subtask for 10 runs as shown in Figure \ref{fig:sum_acc_med2}. Assistant is a custom specialized GPT 3.5 Turbo model that we used instead of Med III, as using the Med III model for summarization triggered some errors related to safety from OpenAI's policy. The maximum available accuracy for the summarization task is 100 and it can be seen that GPT 3.5, Tele and Llama perform very similarly and are the best summarizers while Assistant and Med perform worst.

\begin{figure}[ht]
\centering
\includegraphics[width=\linewidth,keepaspectratio]{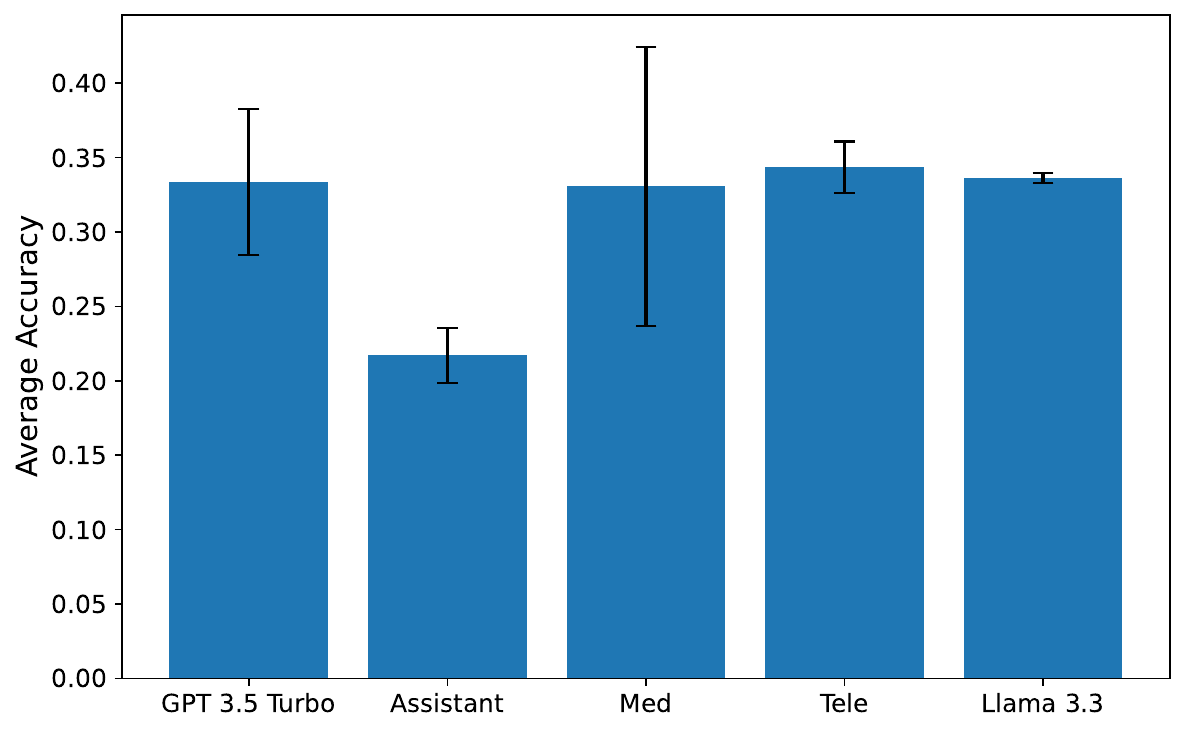}
\caption{Diagnosis accuracies given the summarizer is selected as the model shown for random diagnoser selection} 
\label{fig:sum_diag_eff_rand}
\end{figure}

Next, we investigate whether a higher accuracy summary necessarily leads to higher accuracy for diagnosis. This result, present in Figure \ref{fig:sum_diag_eff_rand} shows that this is not necessarily the case, though there is some correlation between the two. Figure \ref{fig:sum_diag_eff_rand} essentially shows the mean diagnosis accuracy when a particular model is chosen as the summarizer. Hence, the values presented are an average of the diagnoses prediction resulting from the summaries given by the models shown. In this case, both the diagnoser and the summarizer are selected at random to reduce any bias. If a good summary consistently leads to a successful diagnosis, we would expect Figures \ref{fig:sum_acc_med2} and \ref{fig:sum_diag_eff_rand} to look similar in terms of their ordering. While Assistant has the lowest accuracy, the rest of the models are quite similar in terms of accuracy as seen in Figure \ref{fig:sum_diag_eff_rand} while this is not the case in Figure \ref{fig:sum_acc_med2}. Moreover, the standard deviation values shown by the error bars in Figure \ref{fig:sum_diag_eff_rand} are quite large, indicating a high variance which makes it even harder to conclude that there is a direct correlation between summary accuracy and diagnosis accuracy. Therefore, this shows that it is not possible to prioritize accuracy early on in the pipeline by setting a low $\alpha$ for summarization and then prioritizing cost by setting a high $\alpha$ as a high accuracy summary doesn't necessarily guarantee a high accuracy for the diagnoses.

\begin{figure}[ht]
\centering
\includegraphics[width=\linewidth,keepaspectratio]{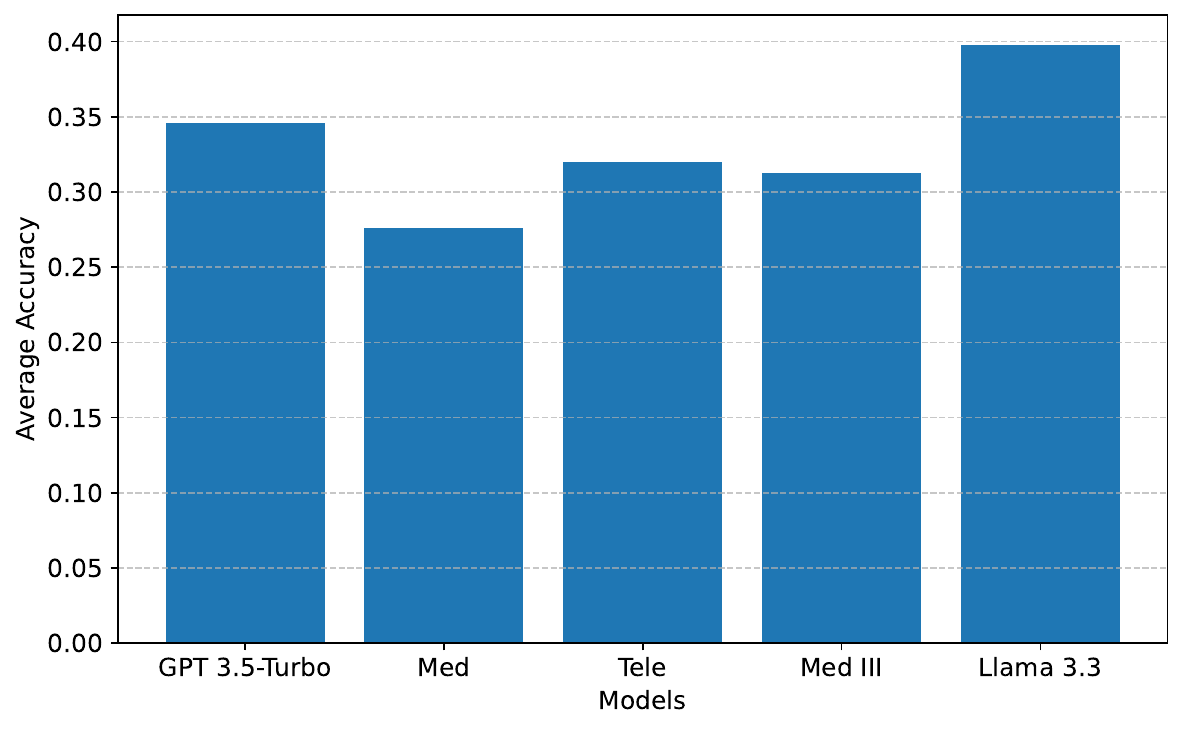}
\caption{Average accuracy of models for the diagnosis subtask for random summarizer and diagnoser selection} 
\label{fig:avg_acc_med2}
\end{figure}

We now look at the average accuracy of the models for the diagnosis subtask in this cost aware setting. Figure \ref{fig:avg_acc_med2} shows the accuracies of the models for this subtask with Llama and GPT 3.5 having the highest accuracies and Med the lowest.

\begin{figure}[ht]
\centering
\includegraphics[width=\linewidth,keepaspectratio]{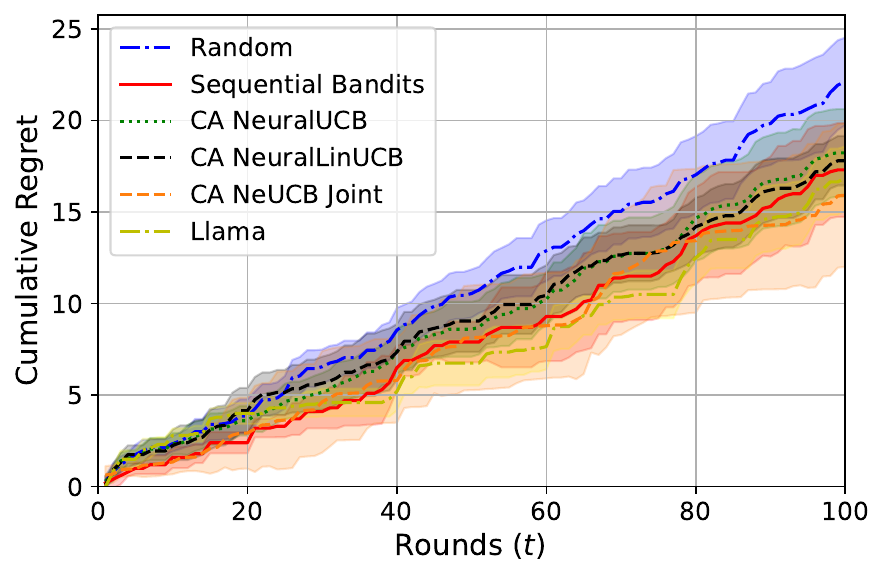}
\caption{Cumulative regret for the diagnosis subtask in 2 subtask medical setting} 
\label{fig:med2_reg_cs_iclr}
\end{figure}

Next, we present the cumulative regret results for this setting as shown in Figure \ref{fig:med2_reg_cs_iclr}. As expected, Random gets the highest regret while the rest of the algorithms perform quite comparably with Sequential Bandits and Llama baselines slightly outperforming CA NeuralUCB and NeuralLinUCB. This result is consistent with the net reward result for 2 subtask medical setting we presented in the main paper.

\begin{figure*}[ht]
    \centering
    \includegraphics[width=\linewidth]{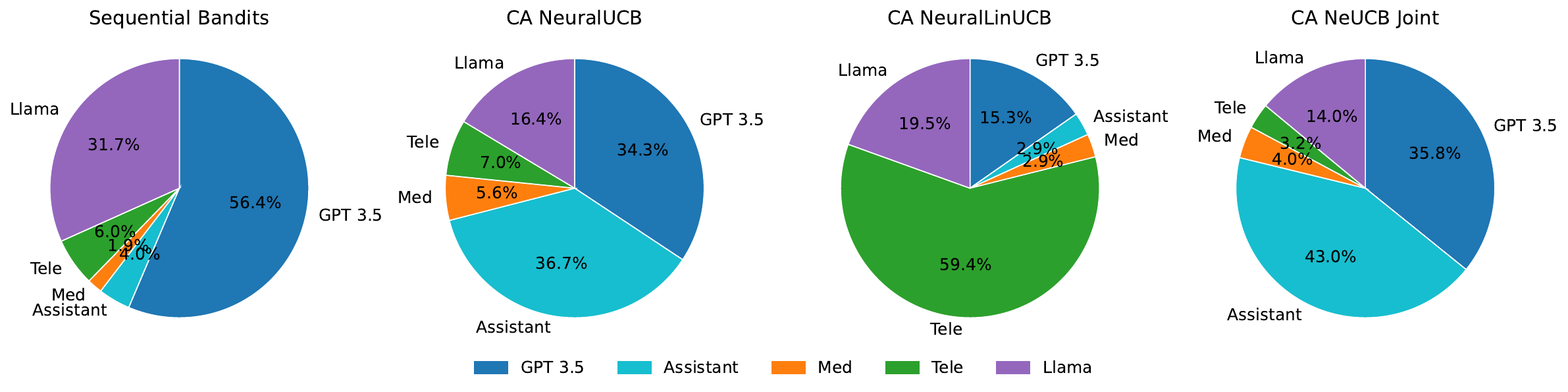}
    \caption{Model selection distribution for Medical (2 subtask) dataset summarization subtask across different algorithms. Each pie chart shows the proportion of model usage.}
    \label{fig:pie_med2_sum_all}
\end{figure*}

Finally, we look at the model selections for the summarization subtask in this setting. Figure \ref{fig:pie_med2_sum_all} shows the model selections of all the algorithms. As mentioned in the main paper, even though Sequential Bandits and CA NeUCB Joint had similar selections for the diagnosis subtask, since their selections for the summarizers differ as shown, this causes the main performance difference. These selections also explain why CA NeuralLinUCB incurs the second highest cost among all algorithms, as it selects Tele very frequently $(\%59.4)$ which is the most expensive model along with the other finetuned models.

\subsection{Results on Response Latency vs Token Length}
\begin{figure}[ht]
\centering
\includegraphics[width=\linewidth,keepaspectratio]{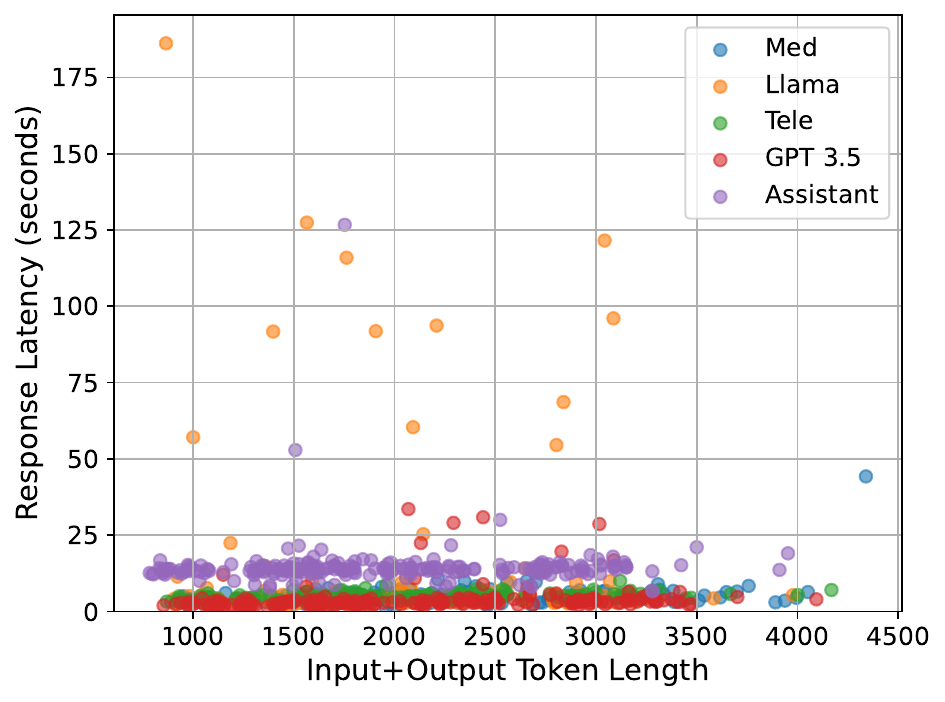}
\caption{Scatter plot showing response latency against the sum of the number of input and output tokens for the summarization subtask} 
\label{fig:scatter_all_sum}
\end{figure}

\begin{figure}[ht]
\centering
\includegraphics[width=\linewidth,keepaspectratio]{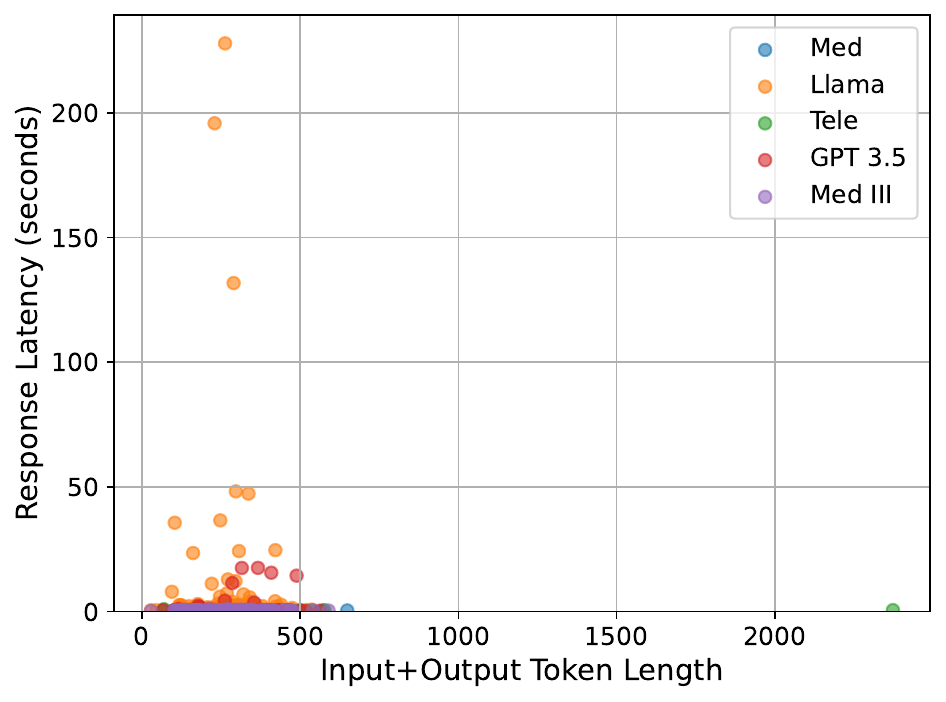}
\caption{Scatter plot showing response latency against the sum of the number of input and output tokens for the diagnosis subtask} 
\label{fig:scatter_all_diag}
\end{figure}

Finally, we present some results on response latency and its variation with the token length. As we mentioned in the main paper, response latency could also be considered as a cost and this is a possible future direction of this work of incorporating response latency as well into the objective. We first start by running Random selection over 1000 instances (10 runs) for the summarization subtask for the medical dataset and do a scatter plot of all points as shown in Figure \ref{fig:scatter_all_sum}. We see that Assistant seems to have a higher latency overall when compared with the other models but there doesn't seem to be a specific trend showing that increasing total number of tokens leads to a higher response latency. We see a similar trend when we look at the response latency variation with the number of tokens for the diagnosis subtask in Figure \ref{fig:scatter_all_diag} with the major difference being that the number of tokens are much lower compared to summarization. We notice some outliers specifically by the Llama model as shown by the yellow scatters in Figure \ref{fig:scatter_all_sum}. We expect this to be because there may be congestion on Llama as it is one of the most widely used LLMs. 

\begin{figure}[ht]
\centering
\includegraphics[width=\linewidth,keepaspectratio]{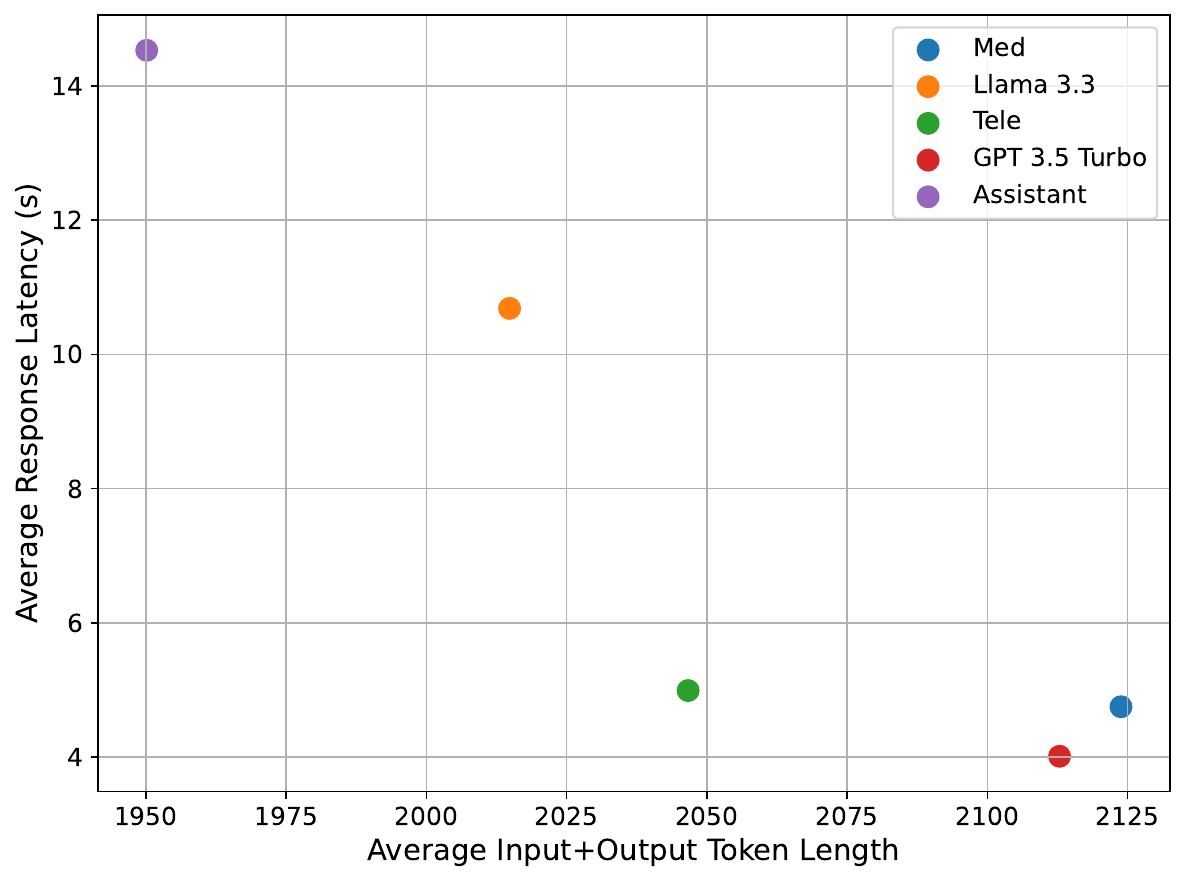}
\caption{Scatter plot showing average response latency against the sum of the number of input and output tokens for the summarization subtask} 
\label{fig:scatter_all_sum_avg}
\end{figure}

\begin{figure}[ht]
\centering
\includegraphics[width=\linewidth,keepaspectratio]{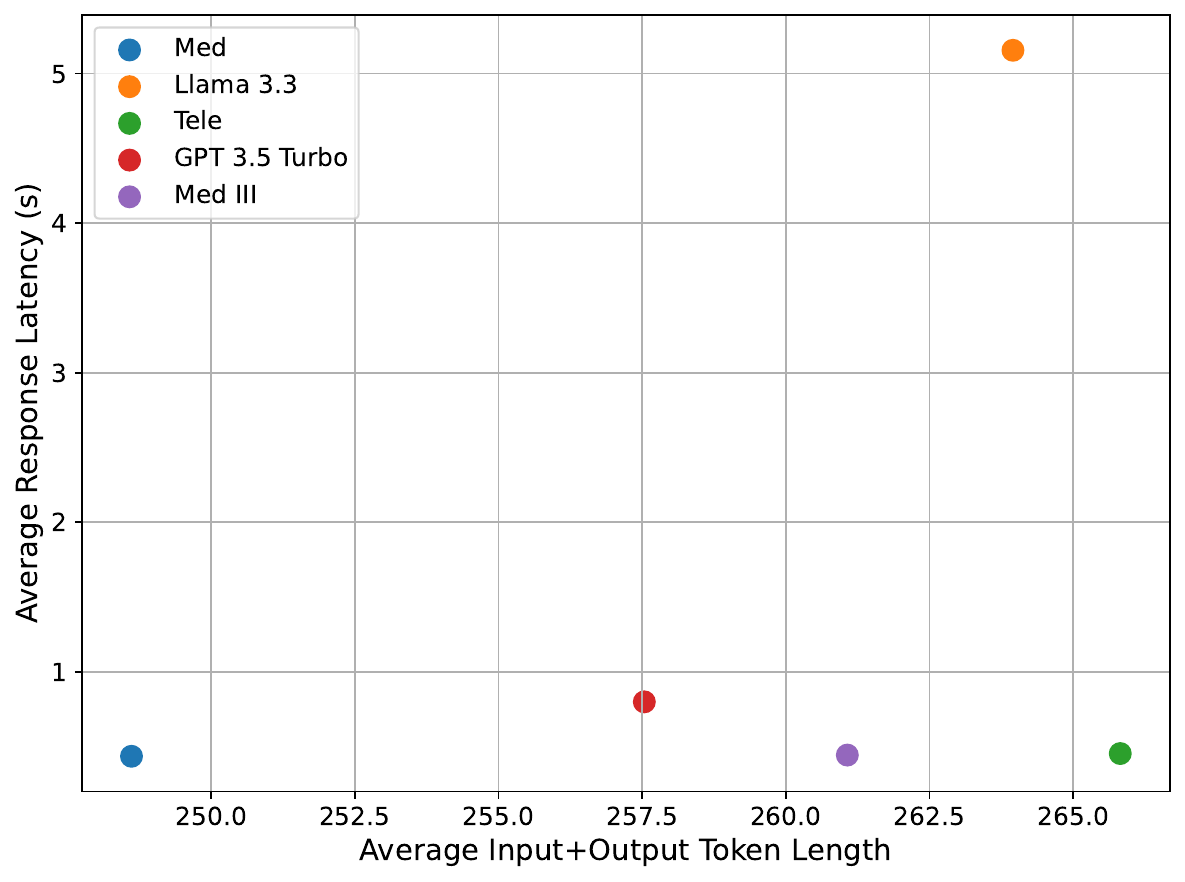}
\caption{Scatter plot showing average response latency against the sum of the number of input and output tokens for the diagnosis subtask} 
\label{fig:scatter_all_diag_avg}
\end{figure}

Next, we show the average response latency with the total number of tokens for the different models. It can be seen from Figures \ref{fig:scatter_all_sum_avg} and \ref{fig:scatter_all_diag_avg} that as we could understand from Figure \ref{fig:scatter_all_sum}, Assistant has the highest average response latency even though it has the lowest total number of tokens. Med and Tele have similar latency values as they are finetuned from the same base model GPT 4o and Llama has the highest latency after Assistant. We see similar trends for the average response latency for the diagnosis subtask in Figure \ref{fig:scatter_all_diag_avg} as Llama has higher latency than the rest of the models, which show similar latencies.

\begin{figure}[ht]
\centering
\includegraphics[width=\linewidth,keepaspectratio]{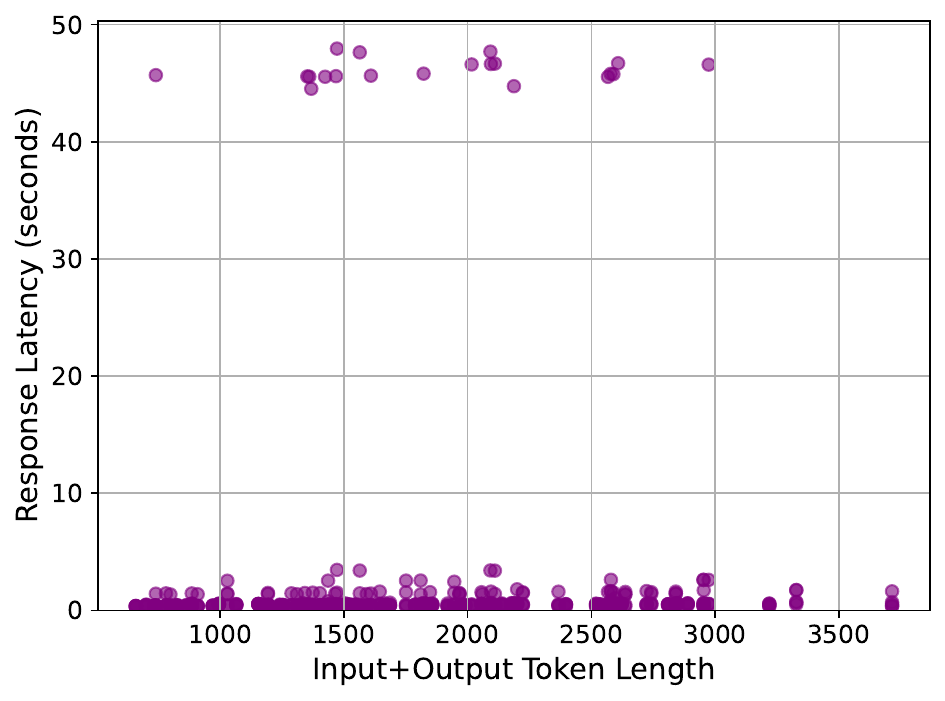}
\caption{Scatter plot showing average response latency against the sum of the number of input and output tokens for the diagnosis subtask for the Med III model} 
\label{fig:scatter_all_diag_medIII}
\end{figure}

\begin{figure}[ht]
\centering
\includegraphics[width=\linewidth,keepaspectratio]{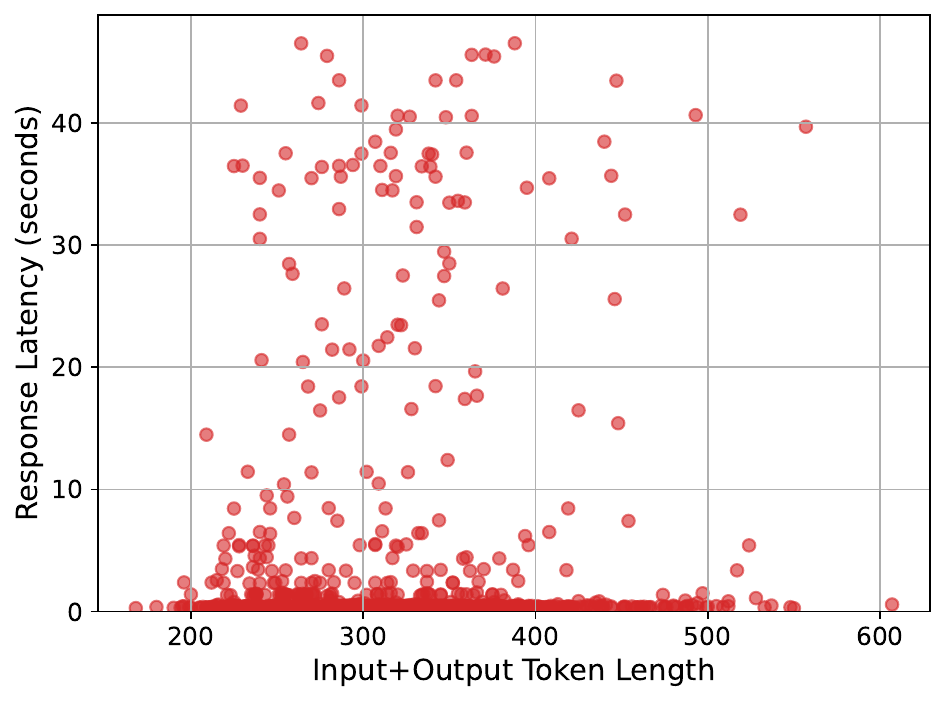}
\caption{Scatter plot showing average response latency against the sum of the number of input and output tokens for the diagnosis subtask for the GPT 3.5 Turbo model} 
\label{fig:scatter_all_diag_gpt35}
\end{figure}

\begin{figure}[ht]
\centering
\includegraphics[width=\linewidth,keepaspectratio]{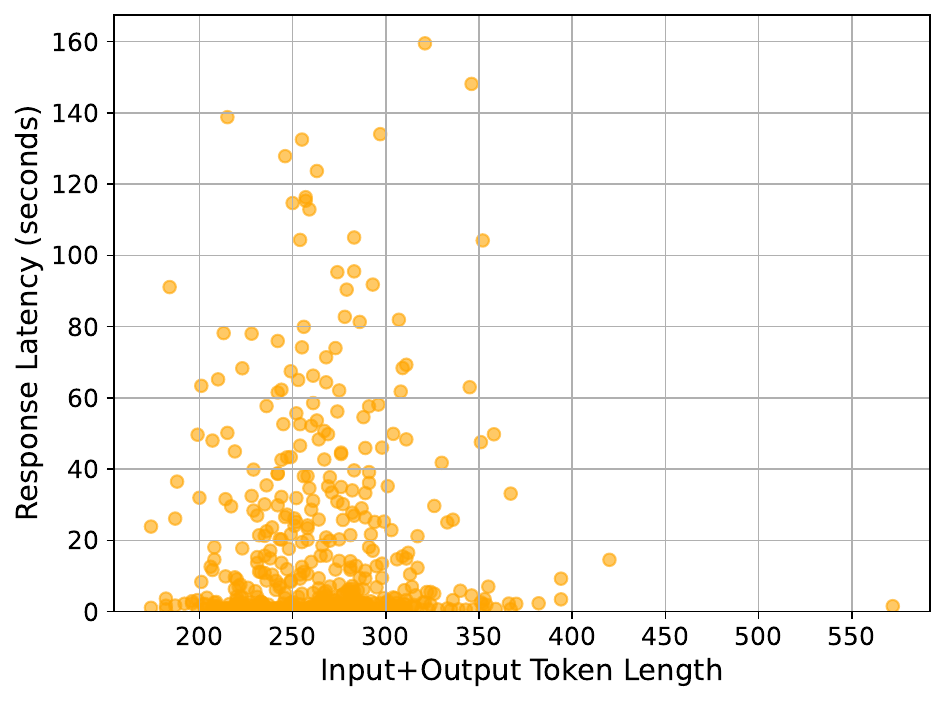}
\caption{Scatter plot showing average response latency against the sum of the number of input and output tokens for the diagnosis subtask for the Llama model} 
\label{fig:scatter_all_diag_llama}
\end{figure}

Finally, we take a look at the scatter plots for response latency against number of tokens for some of the models separately so see if there is some correlation when we consider the models separately. Figures \ref{fig:scatter_all_diag_medIII}, \ref{fig:scatter_all_diag_gpt35} and \ref{fig:scatter_all_diag_llama} show these variations for Med III, GPT 3.5 Turbo and Llama models respectively for the diagnosis subtask. There are 500 points in each of these figures. While there does not seem to be a particular trend between response latency and number of tokens, it can be observed that the more utilized models through APIs such as Llama and GPT 3.5 show a lot more variation and increased latency times for some queries as indicated by the outliers in the plots which are numerous. On the other hand, Figure \ref{fig:scatter_all_diag_medIII} shows that this is a much rarer occurence for the Med III model which is our own finetuned model, which illustrates the benefit of using finetuned models which are not widely used or available to others. This further motivates the need for LLM selection algorithms, as we can avoid choosing the higher latency models if we can identify that certain specialized models for our task at hand can perform better or similarly but with a much improved response latency.

\end{document}